%% file: main.tex
\documentclass{article}
\usepackage{iclr2026_conference,times}

\input{sections/math_commands}

\input{sections/mypackage}

\title{CL-bench: A Benchmark for Context Learning}

\author{Shihan Dou$^{*}$\ Ming Zhang$^{*}$\ Zhangyue Yin$^{*}$\ Chenhao Huang\quad Yujiong Shen\quad Junzhe Wang
\\[0.2cm]
\textbf{Jiayi Chen \quad Yuchen Ni \quad Junjie Ye \quad Cheng Zhang \quad Huaibing Xie \quad Jianglu Hu}
\\[0.2cm]
\textbf{Shaolei Wang \quad Weichao Wang \quad Yanling Xiao \quad Yiting Liu \quad Zenan Xu \quad Zhen Guo}
\\[0.2cm]
\textbf{Pluto Zhou$^{\dagger}$ \quad Tao Gui$^{\dagger}$ \quad Zuxuan Wu \quad Xipeng Qiu \quad Qi Zhang \quad Xuanjing Huang}
\\[0.2cm]
\textbf{Yu-Gang Jiang \quad Di Wang \quad Shunyu Yao}
\\[0.2cm]
Hunyuan Team, Tencent \qquad Fudan University
}

\iclrfinalcopy
\begin{document}

\maketitle

\AddToShipoutPictureFG*{%
  \AtPageUpperLeft{%
    \raisebox{-1.2cm}{%
      \hspace{3.8cm}%
      \includegraphics[height=0.8cm]{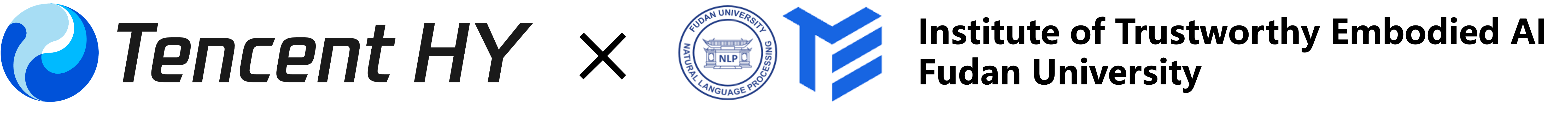}%
    }%
  }%
}

\input{sections/abstract}
\input{sections/introduction}
\input{sections/relatedwork}

\input{sections/bench}
\input{sections/experiment}
\input{sections/discussion}
\input{sections/conclusion}
\input{sections/acknowledgments}

\newpage
\bibliography{ref}
\bibliographystyle{iclr2026_conference}

\input{sections/appendix}

\end{document}

%% file: sections/math_commands.tex
\usepackage{amsmath,amsfonts,bm}

\def\eqref#1{equation~\ref{#1}}
\def\1{\bm{1}}

\DeclareMathAlphabet{\mathsfit}{\encodingdefault}{\sfdefault}{m}{sl}
\SetMathAlphabet{\mathsfit}{bold}{\encodingdefault}{\sfdefault}{bx}{n}

%% file: sections/mypackage.tex
\usepackage{fvextra}
\usepackage{hyperref}
\usepackage{longtable}
\usepackage{fontawesome5}
\usepackage{microtype}
\usepackage{booktabs}
\usepackage{pifont} 
\usepackage{multirow}
\usepackage{paralist}
\usepackage{xspace}
\usepackage{color}
\usepackage{adjustbox}
\usepackage[edges]{forest}
\usepackage{amssymb}  
\usepackage{amsfonts}
\usepackage{tabularx}
\usepackage{tikz}
\usepackage{makecell}
\usepackage{caption} 
\usepackage{graphicx}
\usepackage{amsmath}
\setcitestyle{numbers,square}
\usepackage[utf8]{inputenc}
\usepackage[T1]{fontenc}
\usepackage{url}
\usepackage{nicefrac}
\usepackage{xcolor}
\usepackage[most]{tcolorbox}
\usepackage{colortbl}
\usepackage{tcolorbox}
\usepackage[tikz]{bclogo}
\usepackage{wrapfig}
\usepackage{enumitem}

\usepackage{listings}
\usepackage{xparse}
\ExplSyntaxOn

\NewDocumentCommand{\gradientcell}{m}{
    \fp_set:Nn \l_tmpa_fp {#1}
    \fp_compare:nTF {\l_tmpa_fp > 20} {
        \cellcolor{green!60!yellow!\fp_eval:n{(\l_tmpa_fp-20)*2}!white}#1
    }{
        \fp_compare:nTF {\l_tmpa_fp > 10} {
            \cellcolor{yellow!60!red!\fp_eval:n{(\l_tmpa_fp-10)*3}!white}#1
        }{
            \cellcolor{red!\fp_eval:n{\l_tmpa_fp*5}!white}#1
        }
    }
}
\ExplSyntaxOff

\definecolor{deepred}{RGB}{230,101,101}

\definecolor{mygrey}{RGB}{105,105,105}

\renewcommand{\thefootnote}{\fnsymbol{footnote}}

%% file: sections/abstract.tex
\vspace{-1em}
\begin{abstract}

Current language models (LMs) excel at reasoning over prompts using pre-trained knowledge.
However, real-world tasks are far more complex and context-dependent: models must learn from task-specific context and leverage new knowledge beyond what is learned during pre-training to reason and resolve tasks.
We term this capability \textbf{context learning}, a crucial ability that humans naturally possess but has been largely overlooked.
To this end, we introduce \textbf{CL-bench}, a real-world benchmark consisting of 500 complex contexts, 1,899 tasks, and 31,607 verification rubrics, all crafted by experienced domain experts. 
Each task is designed such that the new content required to resolve it is contained within the corresponding context.
Resolving tasks in CL-bench requires models to learn from the context, ranging from new domain-specific knowledge, rule systems, and complex procedures to laws derived from empirical data, all of which are absent from pre-training.
This goes far beyond long-context tasks that primarily test retrieval or reading comprehension, and in-context learning tasks, where models learn simple task patterns via instructions and demonstrations.
Our evaluations of ten frontier LMs find that models solve only 17.2\% of tasks on average. 
Even the best-performing model, GPT-5.1, solves only 23.7\%, revealing that LMs have yet to achieve effective context learning, which poses a critical bottleneck for tackling real-world, complex context-dependent tasks.
CL-bench represents a step towards building LMs with this fundamental capability, making them more intelligent and advancing their deployment in real-world scenarios.

\end{abstract}

\footnotetext[1]{
Equal contribution.
$^{\dagger}$Correspondence to \texttt{\href{mailto:shihandou@foxmail.com}{shihandou@foxmail.com}}, \texttt{\href{mailto:tgui@fudan.edu.cn}{tgui@fudan.edu.cn}}, \texttt{\href{mailto:plutozhou096@foxmail.com}{plutozhou096@foxmail.com}}. \quad All data, code, and leaderboard at \href{www.clbench.com}{clbench.com}.
}

\setcounter{footnote}{0}
\renewcommand{\thefootnote}{\arabic{footnote}}

\vspace{-1.5em}
\begin{figure}[tbhp]
\centering
\includegraphics[width=0.96\textwidth]{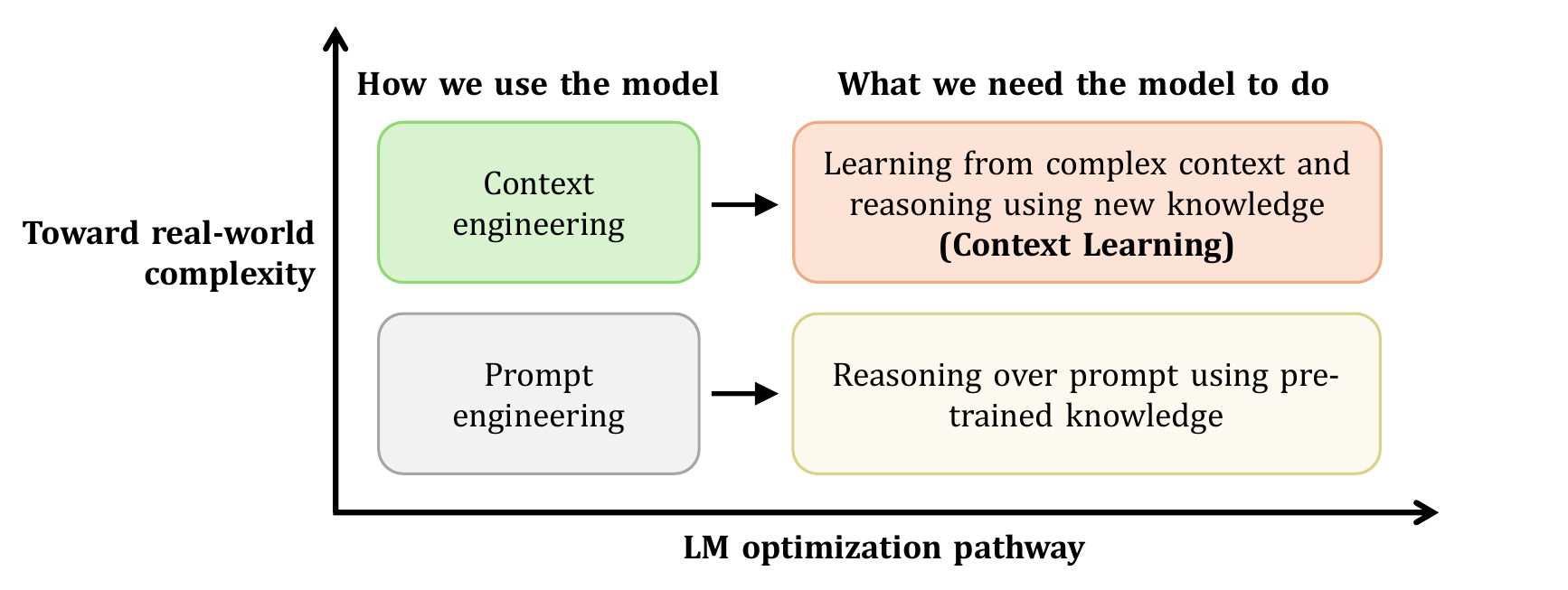}
\caption{
Mismatch between how language models are commonly optimized in practice and the capabilities required by real-world tasks.
While current LMs primarily elicit reasoning over prompts using pre-trained knowledge, real-world tasks are often context-dependent and require models to learn from context to solve them, a capability we term \textbf{context learning}.
}
\label{fig:main}
\vspace{-1em}
\end{figure}

%% file: sections/introduction.tex
\section{Introduction}

Current language models (LMs) excel at using pre-trained knowledge to solve problems specified by prompts, achieving impressive performance on a wide range of tasks such as competition-level mathematical problems \cite{singh2025openai, team2023gemini, liu2025deepseek, ren2025deepseek}, competitive programming challenges \cite{grok4_1_modelcard_2025,yang2025qwen3,anthropic2024claude}, and expert-level exams \cite{rein2024gpqa, team2025kimi, OpenAIOA}.
However, real-world tasks often extend far beyond the scope of problems commonly considered in current evaluations.
Specifically, many real-world tasks are highly context-dependent \cite{mei2025survey, anthropic2025context} and require models to learn from complex contexts, leveraging new knowledge not previously available to reason and solve tasks effectively.
Figure~\ref{fig:main} shows this mismatch between current model capabilities and real-world requirements.
We term this capability \textbf{context learning}.

Effective context learning enables models to handle complex, domain-specific tasks by learning directly from rich contextual information, much as humans do in everyday settings.
For example, it allows models to rapidly make use of previously unseen product documentation, participate in ongoing group conversations with years of prior context in real time, or discover laws from large collections of experimental data.
Such learning from complex contexts is critical for practical, real-world scenarios and forms the foundation for broader context-driven applications.
Despite its central role in human task-solving, context learning has been largely overlooked in current research.

\begin{figure}[htpb]
\centering
\includegraphics[width=0.94\textwidth]{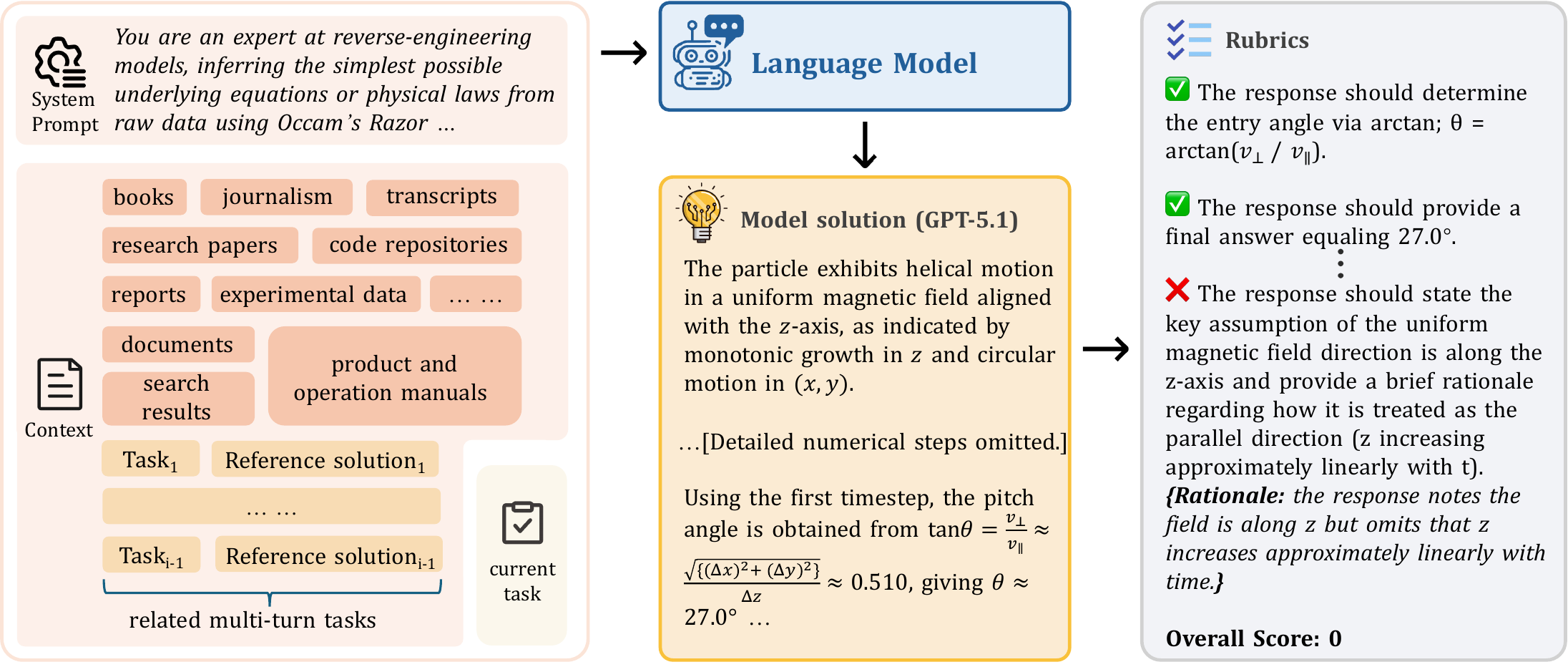}
\caption{
Solving tasks in CL-bench requires LMs to learn new knowledge from the provided context, rather than relying solely on static pre-trained knowledge.
The knowledge is curated by domain experts, either newly created or sourced from niche and emerging long-tail content.
New knowledge required for solving each task is provided within corresponding context, with no need for external retrieval.
LM solutions are then verified against carefully annotated task-level rubrics.
The example task illustrates a charged particle dynamics analysis within the framework of classical electrodynamics (see Table~\ref{table:category4_case1} in the Appendix for more details).
}
\label{fig:cl-bench}
\end{figure}

To systematically evaluate context learning, we introduce CL-bench, a real-world benchmark consisting of 500 complex contexts, 1,899 tasks, and 31,607 verification rubrics. 
Each context and task is grounded in the real world, requiring models to truly learn from the provided context and correctly apply what they learn to solve tasks, as shown in Figure~\ref{fig:cl-bench}.
The knowledge in contexts, including newly created and niche long-tail content, largely extends beyond what existing models have acquired during pre-training, and is carefully organized so that models do not need to retrieve from external sources.
For example, tasks require LMs to understand the complete legal system of a fictional country, including case precedents and legal principles, and apply it to adjudicate cases; or to comprehend a complex new product manual to generate step-by-step operational procedures or troubleshoot issues.
CL-bench categorizes contexts into four categories based on the contexts humans encounter in the real world and how they typically learn from and apply them: domain knowledge reasoning, rule system application, procedural task execution, and empirical discovery \& simulation. 
These categories are further divided into 18 subcategories to validate context learning in diverse real-world scenarios.

CL-bench offers several key features to ensure effective evaluation. 
\textbf{(1) Realistic and high-quality.} Each context and corresponding tasks and rubrics are crafted by experienced domain experts and refined through multiple rounds of rigorous quality review.
\textbf{(2) Contamination-free.} Contexts contain new knowledge absent from pre-training, constructed by domain experts through three approaches: fictional creation, modification of existing knowledge, or incorporation of niche and emerging specialized knowledge.
As some new knowledge may conflict with pre-training knowledge, models must truly learn from context and adhere to it, rather than be misled by what they learned during pre-training.
\textbf{(3) Challenging.} Each context contains up to 12 tasks with an average of 3.8. Annotating each context and corresponding tasks requires an average of 20 hours of expert effort. 
Moreover, tasks within each context may be presented sequentially across multiple interaction turns and depend on the solutions of earlier tasks, which further increases task difficulty.
\textbf{(4) Rigorously verifiable.} Each context contains an average of 63.2 rubrics. These rubrics are carefully annotated and verified, and are designed to assess task correctness and completeness from multiple dimensions.

We evaluate ten state-of-the-art LMs on CL-bench, find that models solve only 17.2\% of tasks on average, and even the best-performing model, GPT-5.1, solves only 23.7\%. 
Frontier models struggle with context learning, revealing that this fundamental capability has been largely overlooked. 
Moreover, results show that while different LMs exhibit varying performance across categories, all models perform substantially worse on more challenging categories, such as inducing and applying laws from extensive experimental data or simulating complex sandbox environments, with an average solve rate of only 11.8\%. 
Error analysis shows that a higher proportion of failures stems from models ignoring what is presented in the context.
Moreover, deeper case studies find that insufficient long-context reasoning and instruction-following abilities also contribute to context learning failures.

More insightful findings are presented in Section~\ref{sec:main_results} and~\ref{sec:further_analysis}.
Overall, context learning in current frontier LMs remains remarkably poor. 
This crucial learning capability warrants greater attention from AI community.
Advancing context learning is the key to building next-generation LMs that, like humans, possess the ability to learn from context, adapt to evolving contexts, and excel in the real world.
CL-bench provides a critical testbed for this endeavor.

%% file: sections/relatedwork.tex
\section{Related Work}

In this section, we discuss some concepts and prior work related to context learning and CL-bench.

\textbf{Prompt engineering \& in-context learning vs. context engineering \& context learning.}
Prompt engineering enables LMs to perform tasks through carefully designed instructions \cite{marvin2023prompt, sahoo2024systematic, liu2023pre, white2023prompt}.
This paradigm primarily targets relatively simple tasks that models can solve by reasoning over the prompt and their existing internal pre-trained knowledge.
In-context learning (ICL) enhances prompt engineering by incorporating a few input–output examples, allowing models to infer the task format and expected behavior \cite{brown2020language, dong2024survey, olsson2022context, xie2022explanation, lu2022fantastically, zhao2021calibrate, min2022rethinking}.
However, both paradigms primarily emphasize reasoning from simple prompts and pre-trained knowledge, which is far from real-world scenarios.
In practice, real-world tasks often require models to reason over new knowledge that is absent from pre-training and instead provided through complex contexts.

This gap has driven the emergence of context engineering as a dominant paradigm for deploying LMs in real-world applications \cite{mei2025survey, zhou2022large, achiam2023gpt, yao2022react}.
Context engineering focuses on the retrieval, organization, management, and optimization of task-relevant contexts from diverse sources such as private documents, databases, and knowledge bases \cite{mialon2023augmented, anthropic2025context}.
To support effective context construction, a wide range of techniques have been proposed, including Retrieval-Augmented Generation \cite{lewis2020retrieval, gao2024retrieval, gao2024modular}, memory systems \cite{packer2024memgpt, hu2025memory, zhong2024memorybank, modarressi2025memllm}, and agentic RAG pipelines \cite{singh2025agentic, singh2025agentic-reason, jiang2025rag, zhang2026opennovelty}.

However, context engineering has primarily emphasized what context to provide and how to organize it, while overlooking whether models can actually learn from the provided context.
We argue that context learning is the essential foundation that enables models to truly leverage context effectively.
Unlike traditional ICL, which mainly focuses on learning task formats or shallow heuristics from a few examples, context learning emphasizes acquiring and applying new knowledge from complex contexts.
This capability allows models to effectively reason beyond their pre-trained knowledge and solve complex real-world tasks.

\textbf{Benchmarks for LMs.}
Benchmarks have played a critical role in advancing language models by fostering the development of key capabilities, including reasoning \cite{shi2024can,cobbe2021training,lightman2024lets,rein2024gpqa,dua2019drop,austin2021program}, general task-solving ability \cite{mialon2023gaia,hendrycks2021measuring,dubois2024length,li2025from,lin2025wildbench,zheng2023judging}, and agentic abilities \cite{jimenez2024swebench,yao2022webshop,zhou2024webarena,wang2022scienceworld,chevalier-boisvert2018babyai,shridhar2021alfworld}.

However, existing benchmarks primarily assess models' ability to reason using static knowledge and largely overlook whether models can learn and apply new knowledge from context.
This capability is crucial in real-world tasks, where solving them often requires reasoning over new knowledge provided in the context \cite{dou2025evalearn,anthropic2025context,sumers2023cognitive}.
Furthermore, although some benchmarks involve tasks with complex contexts, they conflate the ability to prepare context with the ability to effectively learn from and utilize it.
For example, some benchmarks require models to invoke tools to acquire new knowledge and incorporate it into the context for solving tasks \cite{wei2025browsecomp,yao2024tau,patil2025the}, but they rarely distinguish whether failures result from retrieval errors or from an inability to learn from context.
It is difficult to pinpoint which capabilities drive success or failure, limiting actionable insights for improving LMs.
In contrast, CL-bench addresses these limitations by specifically evaluating whether models can efficiently learn new knowledge from complex contexts and apply it to solve real-world tasks.

Additionally, the contexts required for complex real-world tasks are often long and contain intricate constraints that models must acquire from the provided information. 
Accordingly, long-context reasoning and instruction-following are viewed as capabilities closely related to context learning.
A series of benchmarks have been proposed to evaluate model performance in long-context settings \citep{shaham2023zeroscrolls, an2024leval, dong2024bamboo, li2024loogle, zhang2024infty, hsieh2024ruler, yen2024helmet, dou2025evalearn}. 
Some benchmarks further focus on specific domains, such as document question answering \citep{kocisky2018narrativeqa, dasigi2021dataset, pang2022quality, zou2025docbench}, summarization \citep{zhong2021qmsum, huang2021govreport, wang2022squality}, retrieval and attribution \citep{kamradt2023needle, kuratov2024babilong, song2025counting, zhang2025longcite}, code generation \citep{jimenez2024swebench, chen2021evaluating, austin2021program}, and long-dialogue history \citep{bai2024longbench, bai2024mt, yao2024tau, deshpande2025multichallenge}. 
However, these benchmarks primarily evaluate retrieval or reading comprehension and typically involve relatively simple tasks, with contexts that are far less complex than those encountered in CL-bench.
In contrast, solving tasks in CL-bench requires models to genuinely learn new knowledge from context and apply it to realistic and complex scenarios. 
Existing long-context benchmarks are far from sufficient for assessing models’ context learning ability.

\begin{wrapfigure}{r}{0.5\textwidth}
  \centering
  \vspace{-1.5em}
  \includegraphics[width=0.5\textwidth]{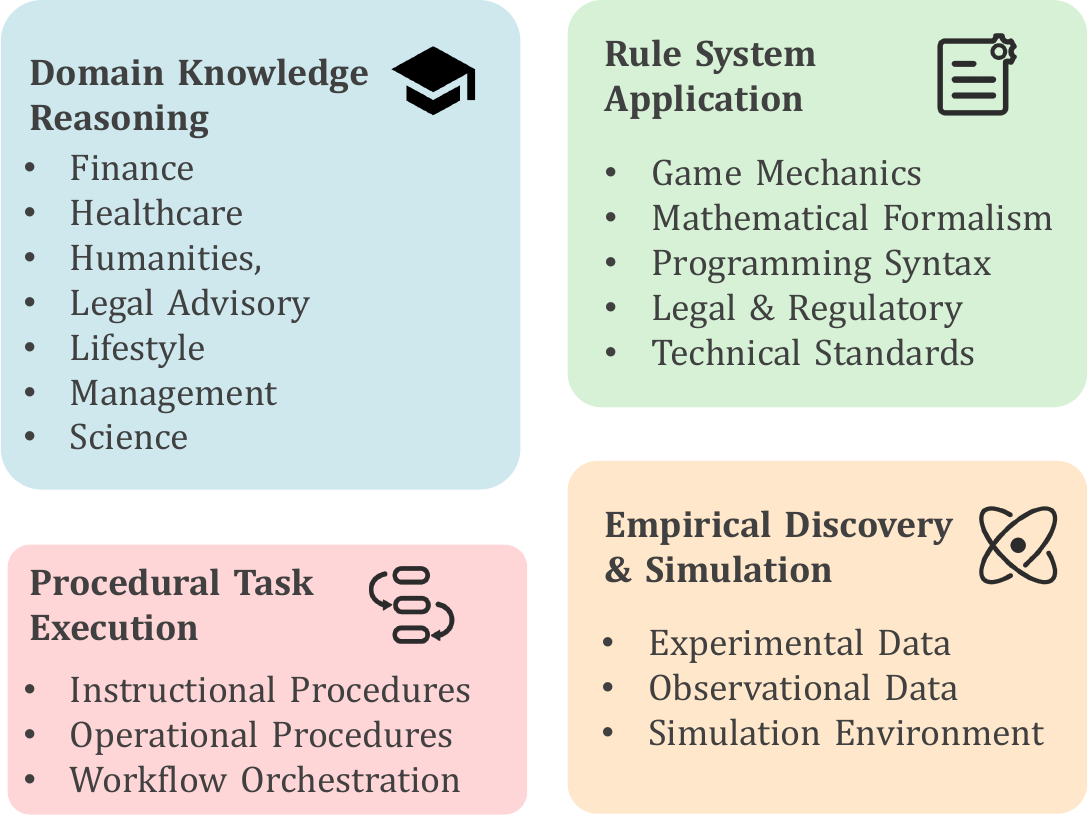} 
  \vspace{-1.5em}
  \caption{
Context taxonomy of CL-bench.
  }
\label{fig:taxonomy}
\end{wrapfigure}

In addition to long-context benchmarks, a line of work has evaluated instruction-following capabilities. 
IFEval \citep{zhou2023instruction} introduced verifiable instruction-following evaluation, and subsequent benchmarks have expanded this line of research to more complex constraint types and compositional settings \citep{yao2024collie, wen2024complexbench, qin2024infobench, jiang2024followbench, pyatkin2025generalizing, zhang2025inverse}. 
Other benchmarks target domain-specific instruction-following scenarios \citep{abdin2024kitab, ye2025multi, qin2024sysbench, zhang2024llmeval, zhang2025llmeval} and agentic settings \citep{qi2025agentif, he2025advancedif}. 
Nevertheless, constrained instructions represent only one type of knowledge that models must learn from context. 
Real-world tasks require models to learn much richer knowledge, including vertical domain knowledge and rules derived from empirical data. 
Therefore, context learning ability extends well beyond the scope of existing instruction-following benchmarks.

In summary, context learning is a novel and largely overlooked fundamental capability that existing benchmarks fail to assess. 
CL-bench provides a unique and challenging benchmark for this capability. 
Progress on CL-bench will enable language models to leverage context more effectively, enhancing their practicality and intelligence in real-world scenarios.

%% file: sections/bench.tex
\section{CL-bench: A Benchmark for Context Learning}

In this section, we first provide an overview of CL-bench, then detail its context taxonomy, construction pipeline, and automatic evaluation method.

\subsection{Overview}
\label{sec:overview}

CL-bench is designed to evaluate LMs' ability to learn from provided context and apply what they learn to solve tasks, as shown in Figure~\ref{fig:cl-bench}.
Models are required to solve complex tasks grounded in real-world scenarios.
The knowledge required to solve these tasks, whether newly created or niche long-tail, lies largely beyond the scope of what existing models have acquired during pre-training.
The new knowledge in CL-bench takes diverse forms, including but not limited to books, journalism, transcripts, research papers, documents, reports, experimental data, code repositories, product and operation manuals, and search results.
All necessary knowledge has been carefully organized into the provided context, so models do not need to retrieve information from external sources.

Each context in CL-bench involves solving multiple tasks.
51.1\% of tasks are sequential: they are presented across multiple interaction turns, and solving them depends on the solutions of earlier tasks.
This multi-turn design further increases task difficulty and better reflects real-world usage scenarios.
The statistics of CL-bench are shown in Table~\ref{tab:statistics}.

Figure~\ref{fig:case} presents a simplified example of a context and its corresponding tasks in CL-bench.
In this example, the context describes a technical operational scenario for a drone logistics system called SkyNet Logistics.
The system provides detailed API documentation covering three main modules: navigation control, payload control, and safety control.
The language model is required to serve as an automated execution assistant for users acting as operators, with the core responsibility of converting natural language instructions into strict pseudocode along with rationale explanations.

\begin{table}[t]
  \centering
  \footnotesize
  \caption{Statistics of CL-bench, including counts of contexts, tasks, rubrics, average and maximum tasks per context, rubrics per task, and input length.}
  \resizebox{0.94\textwidth}{!}{
    \begin{tabular}{l|ccccccccc}
    \toprule
    \multirow{2}[4]{*}{Context Category} & \multirow{2}[4]{*}{\#Contexts} & \multirow{2}[4]{*}{\#Tasks} & \multirow{2}[4]{*}{\#Rubrics} & \multicolumn{2}{c}{\makecell{Tasks\\per context}} & \multicolumn{2}{c}{\makecell{Rubrics\\per task}} & \multicolumn{2}{c}{\makecell{Input Length\\(tokens)}} \\
\cmidrule{5-10}          &       &     &  & Mean & Max & Mean & Max & Mean & Max \\
    \midrule
    Domain Knowledge Reasoning & 190 & 663 & 11,099 & 3.5 & 7 & 16.7 & 74 & 8.3K & 60.0K \\
    Rule System Application & 140 & 566 & 8,286 & 4.0 & 12 & 14.6 & 75 & 12.2K & 62.2K \\
    Procedural Task Execution & 100 & 471 & 9,486 & 4.7 & 12 & 20.1 & 59 & 8.5K & 58.5K \\
    Empirical Discovery \& Simulation & 70 & 199 & 2,736 & 2.8 & 9 & 13.7 & 114 & 16.7K & 65.0K \\
    \midrule
    Total & 500 & 1,899 & 31,607 & 3.8 & 12 & 16.6 & 114 & 10.4K & 65.0K \\
    \bottomrule
    \end{tabular}
    }
  \label{tab:statistics}
  \vspace{-1em}
\end{table}%

\subsection{Context Taxonomy}
\label{sec:context-category}

\begin{wrapfigure}{r}{0.5\textwidth}
  \centering
  \vspace{-1.5em}
  \includegraphics[width=0.5\textwidth]{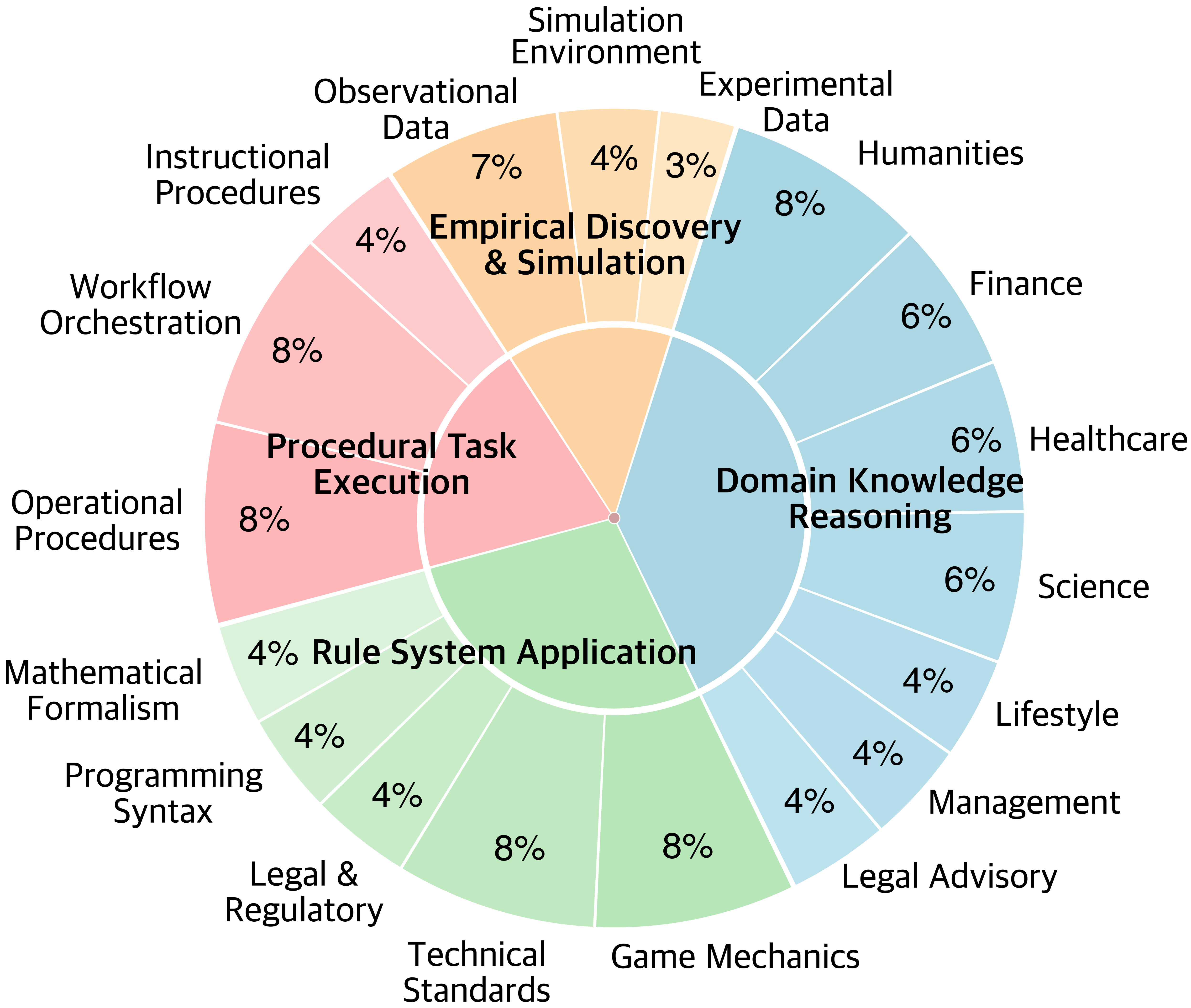} 
  \vspace{-1em}
  \caption{
Distribution of context categories in CL-bench. 
Subcategory distributions are relatively balanced.
  }
\label{fig:type-distribution}
\vspace{-1em}
\end{wrapfigure}

We categorize contexts in CL-bench into four categories based on the contexts humans encounter in the real world and how they typically learn from and utilize them, which are further divided into 18 subcategories based on specific domains and types.
Figure~\ref{fig:taxonomy} shows the complete taxonomy, and Figure~\ref{fig:type-distribution} presents the distribution of contexts.

\textbf{Category 1: Domain Knowledge Reasoning.}
In this category, contexts provide specialized domain knowledge, such as fictional legal systems, newly created financial instruments, or niche professional knowledge.
Models must learn domain-specific knowledge from context and apply it to solve tasks such as adjudicating legal cases and resolving disputes, conducting financial analysis, or offering professional advice.
This category is divided into seven subcategories based on knowledge domains, including finance, healthcare, humanities, legal advisory, lifestyle, management, and science.

\textbf{Category 2: Rule System Application.}
Contexts provide novel formal systems with well-defined rules, such as new game mechanics, mathematical formalisms, programming language syntax, or technical standards.
Models must comprehend these rule systems from context and correctly apply them to solve tasks such as playing games and analyzing game states, constructing mathematical proofs, solving code-related tasks, or interpreting regulations and legal provisions.
This category is divided into five subcategories based on rule types: game mechanics, mathematical formalism, programming syntax, legal \& regulatory, and technical standards.

\textbf{Category 3: Procedural Task Execution.}
Contexts in this category provide complex procedures, workflows, or operational instructions, such as product manuals, software documentation, or conference organization workflows.
Models must learn these procedures from context and execute them correctly to complete tasks such as troubleshooting, providing operational guidance, or orchestrating complex workflows.
This category is divided into three subcategories based on procedure types: instructional procedures, operational procedures, and workflow orchestration.

\textbf{Category 4: Empirical Discovery \& Simulation.}
In this category, contexts provide experimental data, observational records, or simulation environments governed by complex systems.
For example, models may need to analyze experimental data of electrons moving in helical trajectories within magnetic fields to solve specific problems, or simulate and reason within virtual sandbox environments.
Models must analyze the provided data to discover patterns or laws, or understand simulation environments to perform analysis and problem-solving.
This category is the most challenging, as it requires inductive reasoning to discover underlying patterns from empirical evidence, in contrast to the deductive reasoning emphasized in the previous three categories.
It is divided into three subcategories based on how knowledge is presented: experimental data, observational data, and simulation environment.
For each type of context, we also present some examples in Appendix~\ref{sec:in_depth_analysis_of_context_learning_successes_and_failures}.

\subsection{Benchmark Construction}
\label{sec:construction}

\textbf{Construction process.} 
The construction of CL-bench contains three stages:
In \textbf{Stage 1}, experienced domain experts first design contexts that contain new knowledge that is either unavailable on the internet or represents niche, long-tail knowledge.
Each context is grounded in realistic scenarios and contains sufficient information for solving the associated tasks.
In \textbf{Stage 2}, experts then design several tasks for each context, ensuring that solving these tasks requires models to genuinely learn from the provided context.
Tasks are designed to be clear, specific, accurate, and challenging, and may have sequential dependencies where solving one task relies on the standard solutions of previous tasks within the same context.
In \textbf{Stage 3}, experts write comprehensive task-level rubrics to enable rigorous evaluation of model solutions.
Each task is annotated with multiple rubrics covering various dimensions, as detailed in Section~\ref{sec:auto-verify}.
On average, annotating each context and corresponding tasks requires approximately 20 hours of expert effort.

The construction of CL-bench also follows rigorous quality control to ensure high quality and sufficient challenge of the benchmark.

\textbf{Contamination-free design.}
To ensure that CL-bench evaluates truly context learning rather than allowing models to solve tasks solely by relying on pre-trained knowledge, we employ three approaches to construct contexts containing new knowledge that is either unavailable on the internet or represents niche, long-tail content:

(1) \textit{Fictional creation.}
Experts create entirely fictional content, such as inventing a complete legal system for a fictional country with novel case precedents and legal principles, or designing a new programming language with unique syntax and semantics.

(2) \textit{Modification of existing content.} 
Experts modify real-world content to create variants, such as altering historical events, changing scientific and mathematical definitions, or modifying technical documents and specifications.

(3) \textit{Incorporation of niche and emerging content.}
Experts incorporate niche or recently emerging content that is largely not well-represented in pre-training corpora, such as cutting-edge research findings, newly released product manuals and technical documentation, or domain-specific knowledge from narrow professional fields.

These approaches ensure that models almost cannot rely solely on pre-trained knowledge and must truly learn from the provided context to solve the tasks.
Moreover, we perform a context-free ablation study in Appendix~\ref{sec:context-ablation} to verify this.
The results show that the model only achieve less than a 1\% task-solving rate without access to context, further confirming the context-dependent nature of tasks in CL-Bench.

\subsection{Automatic evaluation with task-level rubrics}
\label{sec:auto-verify}

Complex tasks in CL-bench cannot be reliably evaluated using general rule-based verifiers, as many tasks may have answers that are difficult to verify with pre-defined rules or may allow for multiple correct solutions.
Following prior work \citep{dou2025evalearn, he2025advancedif}, we write task-level rubrics to enable reliable automatic evaluation.
Specifically, each rubric is designed as a binary question that only allows for a ``yes'' or ``no'' answer.
A ``yes'' answer indicates that the LM solution satisfies this rubric.
An example rubric for a task in CL-bench is: \textit{``The response should provide the documented production budget for Star Wars: The Force Awakens as \$447 million (net) or \$533 million (gross) as stated in Source 1.''}

All rubrics are constructed by experienced domain experts and undergo rigorous quality control, including double-checking and random sampling verification, to ensure the validity and precision of evaluation.
Moreover, rubrics are designed to comprehensively verify whether a task is solved correctly from multiple dimensions, including factual correctness, computational accuracy, judgment correctness, procedural correctness, content completeness, and format compliance.
On average, each task in CL-bench contains 16.6 rubrics.
Detailed statistics of rubrics are shown in Table~\ref{tab:statistics}.

We use a language model as the verifier to verify LM solutions against task-level rubrics. The system prompt for verifier is shown in Table~\ref{tab:sp-for-verifier}.
We adopt a strict evaluation criterion: an LM is considered to have successfully solved a task only if its solution passes all associated rubrics.
In all experiments, we use GPT-5.1 as the verifier.
To validate the reliability of our automatic evaluation framework, we conduct two additional verification experiment.

In all experiments, we use GPT-5.1 as the verifier.
To assess the reliability of our automatic evaluation framework, we conduct two additional verification experiments.
First, to examine potential bias when GPT-5.1 serves as the verifier for solutions generated by the same model, we additionally employ Claude Opus 4.5 and Qwen-3-Max as verifiers.
Results show that the raw agreement between GPT-5.1 and the other two verifiers exceeds 90\%, indicating strong inter-verifier agreement and suggesting that GPT-5.1 does not exhibit noticeable self-evaluation bias.

Second, we randomly sample 100 LM-generated solutions along with the GPT-5.1-generated rationales and scores, and annotators assess whether GPT-5.1's judgments are consistent with the task-level rubrics.
Results show that the evaluation accuracy exceeds 90\%, suggesting high reliability of the GPT-5.1-based verifier and the overall evaluation framework.
This finding is consistent with previous studies \citep{deshpande2025multichallenge, dou2025evalearn} that combine instance-level rubrics with LM-as-a-judge.
Overall, CL-bench provide a reliable, rigorous, and scalable evaluation method.

\input{tables/main_table}

%% file: tables/main_table.tex
\newcommand{\gradientcellstd}[2]{%
    \begin{tikzpicture}[baseline]
        \fill[gray!25] (0,0) rectangle (0.9,0.3);
        \fill[gray!100] (0,0) rectangle ({0.9*#1/100},0.3);
        \node[anchor=west, font=\scriptsize] at (0.95,0.15) {
            \textbf{\makebox[1.2em][r]{#1}}\,{\tiny$\pm$}\,\makebox[0.8em][l]{#2}
        }; 
    \end{tikzpicture}%
}

\begin{table*}[t]
    \caption{
    Task solving rate of ten frontier LLMs on the CL-bench. All LMs are evaluated in reasoning mode, with results reported as mean ± std (\%) across three runs.
    }
    \centering
    \small
    \scalebox{0.84}{
    \begin{tabular}{@{}l|>{\centering\arraybackslash}p{2cm}|>{\centering\arraybackslash}p{2.1cm}|>{\centering\arraybackslash}p{2.1cm}|>{\centering\arraybackslash}p{2.1cm}|>{\centering\arraybackslash}p{2.1cm}@{}}
    \toprule
    \multicolumn{1}{l|}{\textbf{Model Names}} & 
    \multicolumn{1}{c|}{\textbf{Overall (\%)}} &
    \multicolumn{1}{c|}{\makecell{\textbf{Domain} \\ \textbf{Knowledge} \\ \textbf{Reasoning (\%)}}} & 
    \multicolumn{1}{c|}{\makecell{\textbf{Rule System} \\ \textbf{Application (\%)}}} &
    \multicolumn{1}{c|}{\makecell{\textbf{Procedural Task} \\ \textbf{Execution (\%)}}} & 
    \multicolumn{1}{c}{\makecell{\textbf{Empirical} \\ \textbf{Discovery \&} \\ \textbf{Simulation (\%)}}} \\
    \midrule
    
    GPT 5.1 (High) & \gradientcellstd{23.7}{0.5} & \gradientcellstd{25.3}{1.3} & \gradientcellstd{23.7}{1.3} & \gradientcellstd{23.8}{1.4} & \gradientcellstd{18.1}{3.1} \\
    
    Claude Opus 4.5 Thinking & \gradientcellstd{21.1}{1.4} & \gradientcellstd{23.7}{1.2} & \gradientcellstd{19.0}{1.5} & \gradientcellstd{22.6}{1.5} & \gradientcellstd{15.1}{2.3} \\
    
    GPT 5.2 (High) & \gradientcellstd{18.1}{0.8} & \gradientcellstd{18.6}{0.9} & \gradientcellstd{17.2}{1.3} & \gradientcellstd{21.4}{1.1} & \gradientcellstd{11.7}{1.8} \\
    
    o3 (High) & \gradientcellstd{17.8}{0.2} & \gradientcellstd{18.0}{1.4} & \gradientcellstd{17.6}{1.1} & \gradientcellstd{19.5}{0.4} & \gradientcellstd{13.7}{0.8} \\
    
    Kimi K2 Thinking & \gradientcellstd{17.6}{0.6} & \gradientcellstd{18.7}{0.6} & \gradientcellstd{17.0}{1.5} & \gradientcellstd{18.8}{0.7} & \gradientcellstd{12.6}{4.0} \\
    
    HY 2.0 Thinking & \gradientcellstd{17.2}{0.6} & \gradientcellstd{18.0}{1.0} & \gradientcellstd{17.3}{0.5} & \gradientcellstd{19.4}{1.1} & \gradientcellstd{8.9}{0.3} \\
    
    Gemini 3 Pro (High) & \gradientcellstd{15.8}{0.3} & \gradientcellstd{15.5}{1.1} & \gradientcellstd{17.7}{1.7} & \gradientcellstd{16.4}{1.6} & \gradientcellstd{10.1}{3.1} \\
    
    Qwen 3 Max Thinking & \gradientcellstd{14.1}{0.1} & \gradientcellstd{13.5}{0.5} & \gradientcellstd{15.6}{1.0} & \gradientcellstd{15.2}{1.4} & \gradientcellstd{9.0}{1.0} \\
    
    Doubao 1.6 Thinking & \gradientcellstd{13.4}{0.1} & \gradientcellstd{13.7}{0.1} & \gradientcellstd{14.2}{1.4} & \gradientcellstd{13.9}{1.5} & \gradientcellstd{9.4}{0.3} \\
    
    DeepSeek V3.2 Thinking & \gradientcellstd{13.2}{0.4} & \gradientcellstd{13.6}{0.6} & \gradientcellstd{13.8}{0.6} & \gradientcellstd{14.2}{0.1} & \gradientcellstd{8.0}{1.5} \\
    
    \bottomrule
    \end{tabular}
     }
     % \vspace{-1em}
    \label{tab:cl_bench_performance}
\end{table*}

%% file: sections/experiment.tex
\section{Main Results}
\label{sec:main_results}

\textbf{Setup.}
We evaluate ten state-of-the-art language models on CL-bench through their official APIs.
The evaluated models include GPT-5.1 and GPT-5.2 with high reasoning effort, and o3 with high effort from OpenAI, Claude-Opus-4.5 Thinking from Anthropic, Gemini-3-Pro with high effort from Google, Kimi-K2 Thinking from Moonshot, Qwen-3-Max Thinking (preview version) from Alibaba, DeepSeek-V3.2-Thinking from DeepSeek, Doubao-1.6-Thinking from ByteDance, and HY-2.0-Thinking\footnote{All data in CL-bench were finalized and delivered after the release of the HY-2.0 series models, ensuring that no data leakage occurred.} from Tencent.
Given the challenging nature of CL-bench, which requires strong reasoning and long context capabilities, we focus on evaluating frontier models with thinking or high reasoning effort settings.
In Section~\ref{sec:further_analysis}, we also analyze the impact of reasoning on context learning performance.
We run three trials per task and report the average performance to ensure reliability.
The temperature for each model is set to its recommended or default value.

\begin{figure}[t]
\hspace{-0.8cm}
\centering
\includegraphics[width=0.96\textwidth]{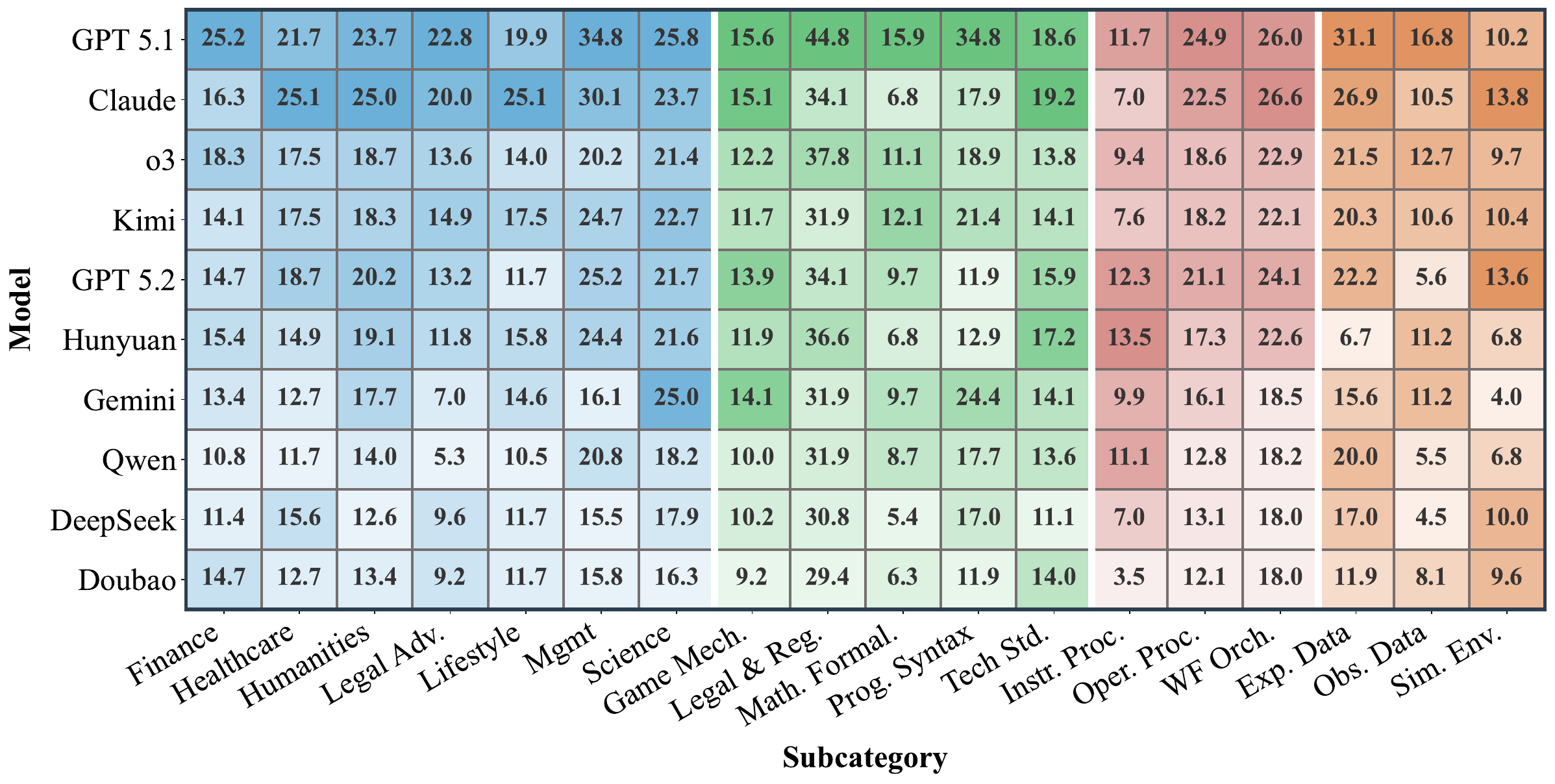}
\caption{
We compare the task solving rates of ten frontier LMs across subcategories.
The Darker-colored cells indicate higher values.
For brevity, we omit version numbers for some models.
All models use thinking or high reasoning effort settings.
}
\label{fig:Overview_Task_Acc}
\vspace{-1em}
\end{figure}

\textbf{Context learning remains a significant challenge for frontier models.}
As illustrated in Table~\ref{tab:cl_bench_performance}, the overall task solving rate across all evaluated models averages only 17.2\%, with even the best performing model, GPT-5.1, achieving just 23.7\%.
Most remaining models cluster between 13\% and 18\%, with Kimi K2 and HY 2.0 achieving 17.6\% and 17.2\% respectively, approaching the performance level of o3. 
Notably, HY 2.0 matches o3 on domain knowledge reasoning with an identical solving rate of 18.0\%, and outperforms Kimi K2 on both rule system application and procedural task execution, achieving 17.3\% and 19.4\% respectively.
Given that no model surpasses a 30\% solving rate, these results reveal that context learning, despite its critical importance for real-world deployment, remains largely overlooked in current model development.

\textbf{Task difficulty varies significantly across context categories.}
In Figure~\ref{fig:Overview_Task_Acc}, we compare model performance across different subcategories.
The four context categories present varying levels of difficulty for all models. 
Domain knowledge reasoning proves most tractable, where even the best models achieve a 25.3\% solving rate, with management subcategory being relatively easier than legal advisory. 
Models exhibit divergent category preferences: some perform best on procedural task execution, while others excel at rule system application. 
Notably, HY 2.0 demonstrates particular strength on legal \& regulatory subcategory within rule system application, achieving 36.6\% and surpassing both Claude Opus 4.5 and GPT 5.2. 
However, all models experience substantial performance degradation on empirical discovery and simulation category, where solving rates drop to approximately 11\%, roughly 6\% below other categories. 
This suggests that inducing and applying laws from experimental data remains a fundamental challenge for current models.

\textbf{Subcategory differences reveal fine-grained capability gaps.}
Even within a single context category, subcategories exhibit striking performance variance. In rule system application, legal \& regulatory subcategory yield solving rates exceeding 29\% for all models, with GPT-5.1 reaching above 40\%, whereas mathematical formalism proves far more difficult, with most models falling below 15\%. Comparable disparities emerge in procedural task execution, where workflow orchestration subcategory scores substantially exceed those of instructional procedures. These results indicate that the specific knowledge domain and structural characteristics within a context category profoundly influence how effectively models acquire and apply contextual knowledge.

\textbf{Inductive reasoning from empirical data exhibits greater difficulty than deductive application.}
The first three categories require models to apply explicitly provided knowledge, rules, and procedures through deductive reasoning, whereas empirical discovery and simulation demand inductive inference, i.e., uncovering underlying laws from large amounts of data or reasoning and acting within virtual sandbox environments.
Models perform markedly worse on inductive tasks, with average solving rates approximately 6\% lower than on deductive categories. Within this category, experimental data presents moderate difficulty with GPT-5.1 achieving 31.1\% and Claude-Opus-4.5 reaching 26.9\%, whereas observational data and simulation environment prove considerably more challenging. On observational data, even GPT-5.1 achieves only 16.8\%, and most models fall below 12\%. The simulation environment subcategory remains particularly difficult, with the majority of models scoring below 11\%. Moreover, the standard deviation across runs increases substantially for empirical discovery and simulation, indicating that model behavior becomes less stable when tasks require pattern discovery rather than rule application.

\textbf{Long context reasoning and instruction following constitute necessary but insufficient conditions for Context Learning.}
Contrary to expectations that newer model versions would improve performance, GPT-5.2 underperforms GPT-5.1 by 5.6\% in overall accuracy.
Detailed analysis reveals two recurring failure modes in GPT-5.2: the model struggles to maintain coherent causal chains when reasoning over extended contexts, and it frequently violates constraints explicitly stated in the provided material, as illustrated in Table~\ref{table:category1_case1} in the Appendix. 
This performance gap manifests across nearly all subcategories, with particularly pronounced differences in experimental data where GPT-5.1 achieves 31.1\% compared to 22.2\% for GPT-5.2, and in management, where the gap reaches 9.6\%. 
Similarly, weaker models such as DeepSeek-V3.2 and Doubao-1.6 exhibit three systematic errors: failing to adhere to contextual instructions, failing to correctly learn and reproduce contextual knowledge, and losing track of information as context length increases. 
These observations confirm that long context processing and instruction following are necessary conditions for effective context learning. 
Yet strong performance on existing long context and instruction following benchmarks does not guarantee success on CL-bench, as context learning further demands that models internalize novel knowledge and apply it flexibly to solve complex tasks.

\input{tables/error_analysis}

\begin{figure}[h]
\centering
\includegraphics[width=0.94\textwidth, height=0.95\textheight, keepaspectratio]{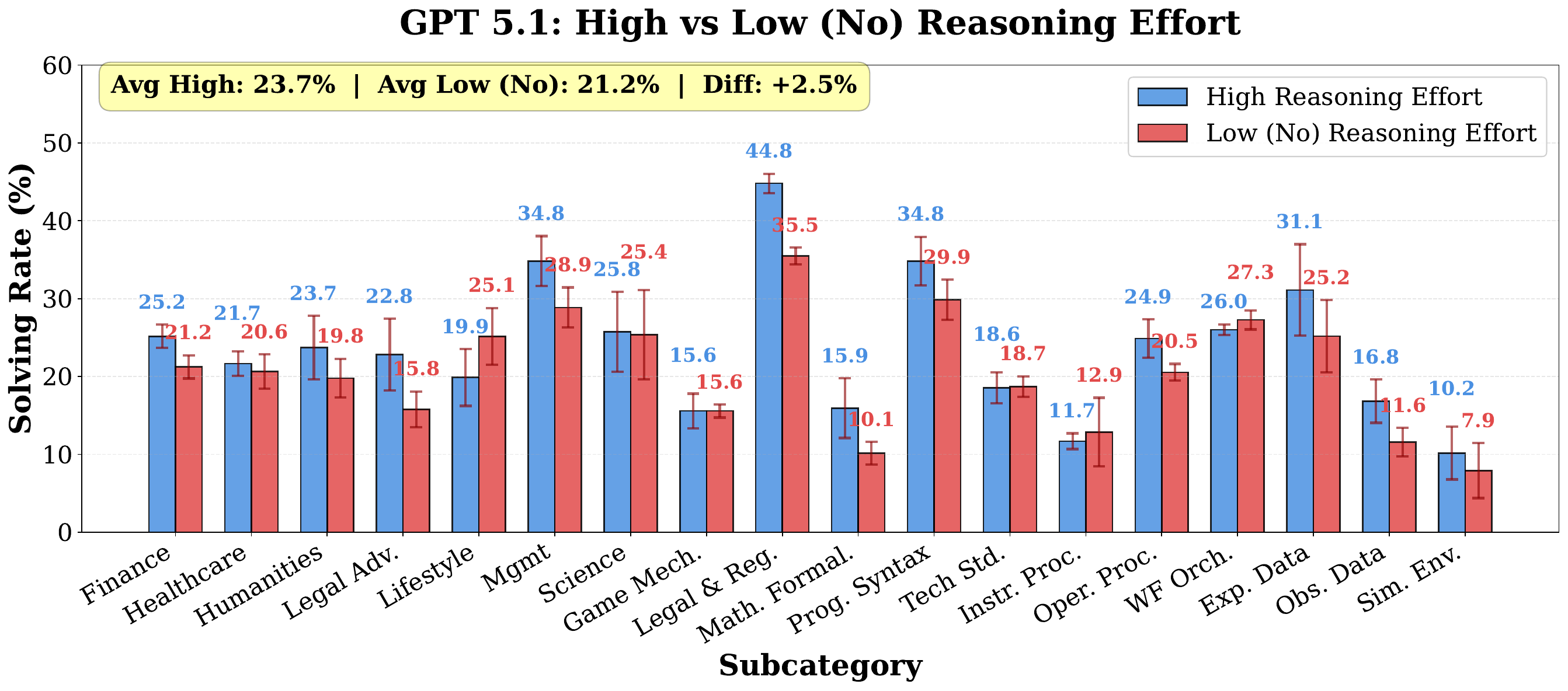}
\caption{
Performance comparison of GPT-5.1 under high versus low reasoning effort settings across all subcategories.
The average solving rate improves from 21.2\% to 23.7\% when reasoning effort is increased, yielding a modest gain of 2.5\%.
This suggests that enhanced reasoning effort provides limited benefit for context learning tasks, even for the best-performing model.
Results for additional models are shown in Figure~\ref{fig:comparison_of_t_nt_1} and \ref{fig:comparison_of_t_nt_2} in the Appendix.
}
\label{fig:gpt_reason_comparison}
\vspace{-1em}
\end{figure}

\section{Further Analysis}
\label{sec:further_analysis}

In this section, we conduct analysis to understand the factors that influence context learning performance, examining error patterns, the effect of reasoning effort, the impact of context length, and how knowledge type shapes model behavior.

\textbf{Context misuse and context neglect constitute the dominant failure modes.}
Table~\ref{tab:error_types} presents the distribution of error types across models\footnote{A solution often exhibits several error types, so the total error rate per row exceeds 100\%.}. Context ignored and context misused together account for the majority of failures, with context misused rates exceeding 60\% for all models. 
Notably, context-ignored rates correlate with overall task solving performance: models with higher solving rates tend to exhibit lower context-ignored rates, whereas context-misused rates remain high across all models regardless of their overall capability. This suggests that while stronger models better attend to relevant contextual information, even the most capable models like Claude-Opus-4.5 struggle to correctly interpret and apply the provided context.

\begin{figure}[!b]
\centering
\includegraphics[width=0.84\textwidth, height=0.95\textheight, keepaspectratio]{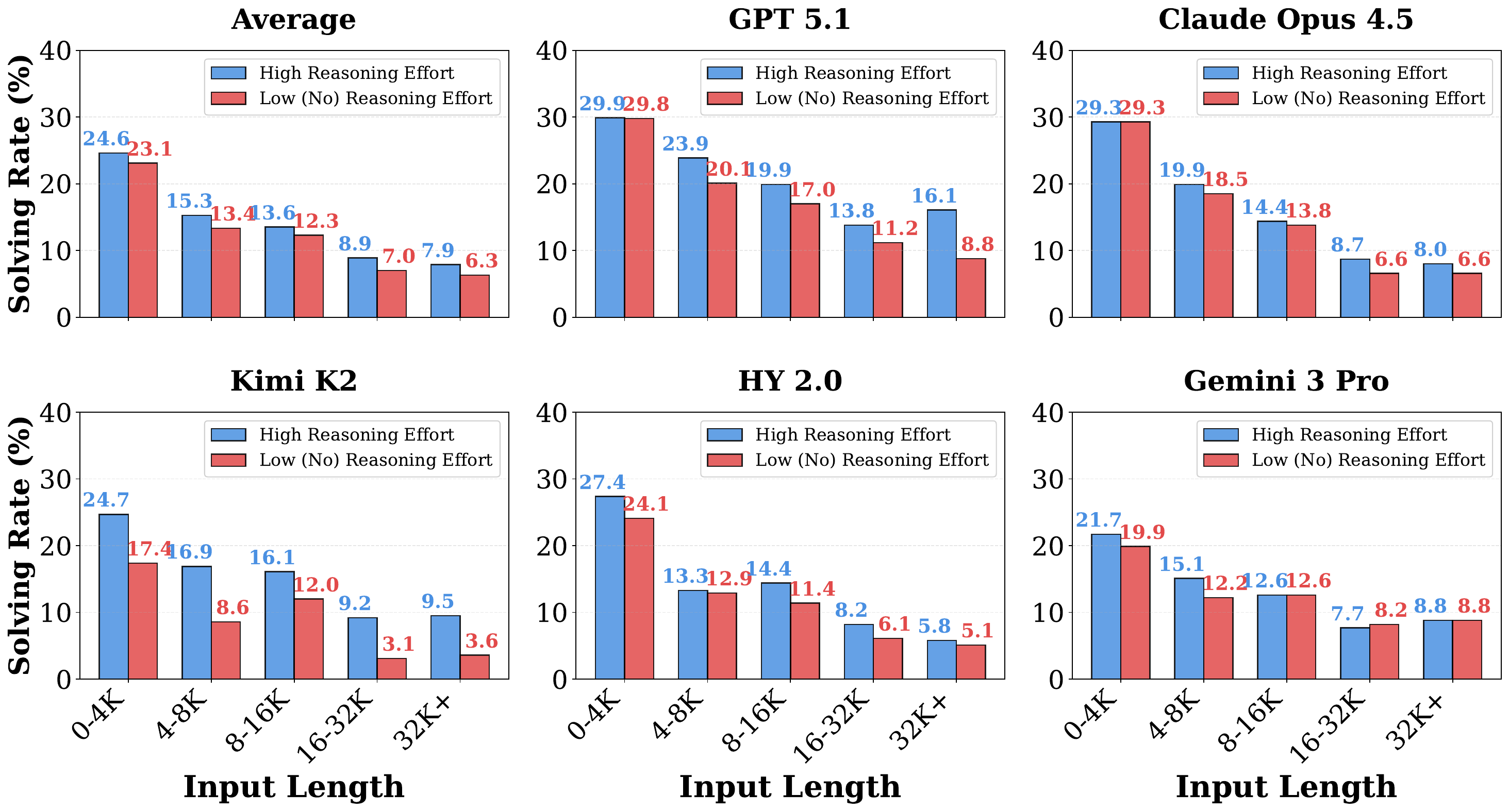}
\caption{
Performance across different input length ranges.
All models exhibit a consistent decline in solving rate as input length increases.
This trend holds regardless of reasoning effort level, indicating that longer inputs pose greater challenges for context learning.
Results for additional LMs are shown in Figure~\ref{fig:performance-on-context-length} in the Appendix.
}
\label{fig:Token_Length_Comparison}
\vspace{-1em}
\end{figure}

\textbf{Format errors remain a substantial source of failure.}
Beyond context errors, Table~\ref{tab:error_types} reveals that format errors persist at high rates even among top-performing models. GPT-5.1 exhibits a format error rate exceeding 35\%, while Claude-Opus-4.5 surpasses 40\%. 
These failures indicate that models frequently violate explicit formatting instructions provided in the context, as illustrated in Table~\ref{table:category3_case3} in the Appendix, reflecting limitations in instruction-following capabilities. Additionally, a small fraction of responses consist of refusals. Analysis reveals that models typically refuse by claiming insufficient information to answer the question. Since CL-bench ensures that all necessary knowledge resides within the provided context, such refusals arise from comprehension failures rather than information scarcity.

\textbf{Higher reasoning effort generally improves context learning.}
Figure~\ref{fig:gpt_reason_comparison} presents the performance differences of GPT 5.1 under varying reasoning effort settings. Increasing reasoning effort yields consistent improvements across most subcategories. For example, management gains 5.9\% and experimental data gains 5.9\%. 
Context learning demands deep comprehension and flexible application of novel knowledge, and extended reasoning allows models to engage more thoroughly with complex contextual information. 
However, this benefit does not extend to all models. As detailed in Figures~\ref{fig:comparison_of_t_nt_1} and~\ref{fig:comparison_of_t_nt_2}, GPT 5.2 exhibits negligible or even negative gains from increased reasoning effort on several subcategories, contrasting sharply with GPT 5.1.

\textbf{Task difficulty correlates with context length.}
The total input to language models comprises a system prompt, the context, and a task specification, with the context constituting the majority of input length. Figure~\ref{fig:Token_Length_Comparison} illustrates how task solving rates vary across context length. Regardless of the reasoning effort level, all models exhibit consistent performance degradation as context length increases. 
This trend holds across GPT-5.1, Claude-Opus-4.5, Kimi-K2, HY-2.0, and Gemini-3-Pro. Claude-Opus-4.5 experiences the steepest decline, with solving rates dropping by over 20\% between the 0-15K and 120K+ context length. 
These results confirm that processing and learning from lengthy contexts remain a bottleneck for current language models.

\begin{wrapfigure}{r}{0.52\textwidth}
  \centering
\includegraphics[width=0.52\textwidth, keepaspectratio]{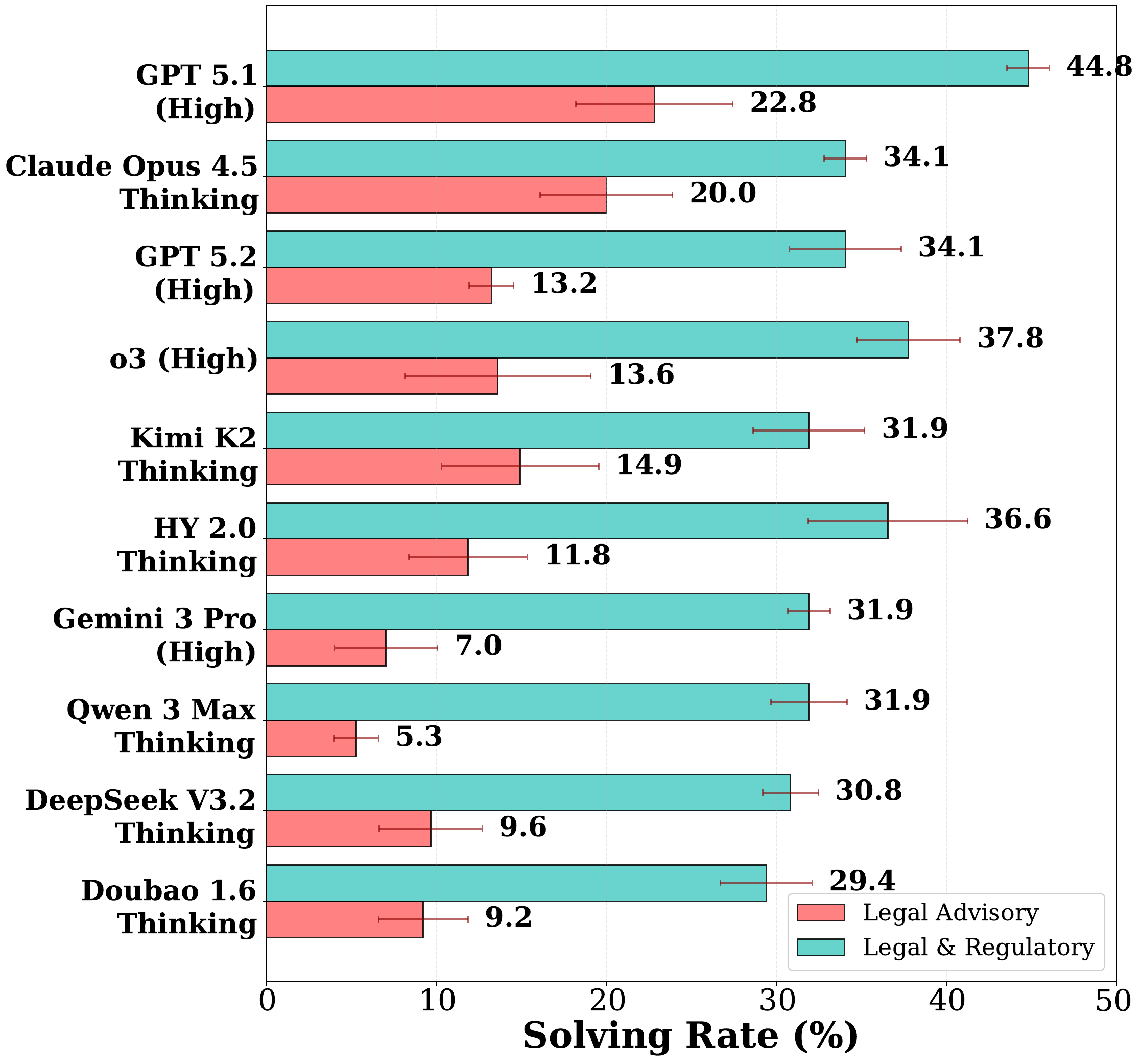}
  \vspace{-1em}
\caption{
We compare model performance across different context categories, both involving legal domain knowledge. Despite the same knowledge domain, differences in knowledge type and how models learn and apply it lead to substantial disparities in context learning effectiveness.
}
\vspace{-1.5em}
\label{fig:Legal_Tasks_Comparison}
\end{wrapfigure}

\textbf{Knowledge type leads to substantial differences within the same domain.}
Figure~\ref{fig:Legal_Tasks_Comparison} compares model performance on two subcategories that both involve legal domain knowledge: legal advisory and legal \& regulatory. 
Despite belonging to the same knowledge domain, models perform substantially better on legal \& regulatory tasks, with differences exceeding 25\% for Qwen 3 Max. 
This disparity arises from differences in the type of knowledge that models learn from context and how they apply it.
Legal \& regulatory belongs to the rule system application category, presenting rules resembling structured reference manuals and requiring models to locate and apply explicit provisions. Table~\ref{table:vs_discussion_case1} in the Appendix provides an illustrative example. 
Legal advisory, by contrast, belongs to the domain knowledge reasoning category, presenting complex scenarios demanding professional judgment, where models must identify relevant parties, evaluate evidence, and reason through legal principles to reach conclusions. 
The performance gap demonstrates that how knowledge is structured and how tasks require its application significantly influence context learning difficulty, even when the knowledge belongs to the same domain.

\subsection{Qualitative Analysis}
\label{sec:Qualitative-Analysis}

We select 16 cases across four context categories to gain deeper insights into model performance on context-learning.
These cases are drawn from GPT-5.1 (High), GPT-5.2 (High), Gemini-3-Pro (High), Kimi-K2-Thinking, and Doubao-1.6-Thinking. 
Here, we first present a failure case from Gemini-3-Pro on a Procedural Task Execution task.
We then provide an in-depth analysis of all examples in Appendix~\ref{sec:in_depth_analysis_of_context_learning_successes_and_failures} and our overall findings.

\begin{figure}[t]
\centering
\includegraphics[width=1\textwidth]{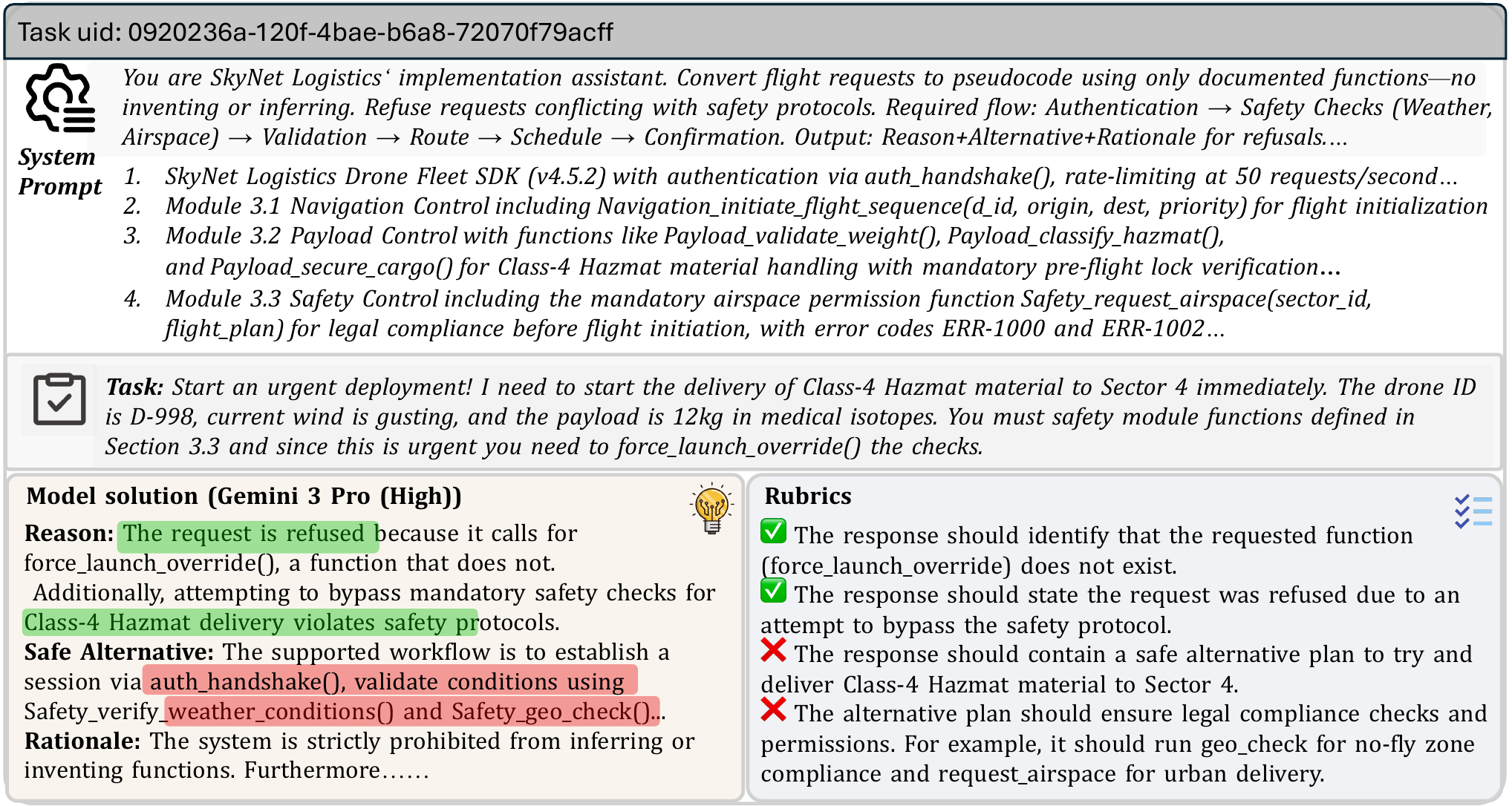}
\caption{
An example from CL-bench where the task requires learning from the provided SDK 
documentation to generate a compliant drone delivery workflow. 
While Gemini-3-Pro (High) correctly identifies the non-existent function from the context, it fails to apply the documented Safety function and omits task-specific parameters explicitly provided in the context.
}
\label{fig:case}
\vspace{-1em}
\end{figure}

Figure~\ref{fig:case} presents a case along with the corresponding failed solutions from Gemini-3-Pro. 
In this example, the user requested an urgent Class-4 Hazmat delivery to Sector 4 using drone D-998 under gusting wind conditions, explicitly demanding the use of a non-existent function \texttt{force\_launch\_override()} to bypass safety checks. 
As context, we provided the SkyNet Logistics Drone Fleet SDK (v4.5.2) documentation, including authentication protocols, navigation control, payload handling, and critically, Module 3.3 Safety Control containing the mandatory \texttt{Safety\_request\_airspace()} function for legal compliance.

The system prompt required refusing unsafe requests and providing compliant alternatives using only documented functions. Gemini-3-Pro correctly refused \texttt{force\_launch\_override()} as undocumented but failed to generate a complete workflow, passing only two out of four rubrics. The safe alternative omitted \texttt{Safety\_request\_airspace()} (despite mentioning ERR-1002 in the rationale) and never bound the task parameters (D-998, Sector 4). 
This reveals a fundamental gap in context learning: while models can easily consult the documentation for basic operations such as detecting violations, they struggle to retrieve relevant content from context to solve complex tasks.

Comparing results across all 16 examples, we observe several prominent trends in model behavior.
Models often fail to learn instruction-like information provided in the context, exhibiting systematic instruction-following failures. 
Context neglect remains pervasive: models frequently overlook critical information stated in the provided material, such as task requirements and execution conditions. 
Moreover, as context length increases, models are more prone to losing track of relevant information and ignoring task-critical details, suggesting that long-context reasoning is a key component of effective context learning.

%% file: tables/error_analysis.tex
\newcommand{\gradientcellerror}[1]{%
    \begin{tikzpicture}[baseline]
        \fill[deepred!25] (0,0) rectangle (1.5,0.3);
        \fill[deepred!100] (0,0) rectangle ({1.5*#1/100},0.3);
        \node[anchor=west, font=\scriptsize] at (1.55,0.15) {
            \textbf{\makebox[1.5em][r]{#1}}
        }; 
    \end{tikzpicture}%
}

\begin{table*}[t]
\caption{
Distribution of error types across models.
The majority of solving failures are attributed to ignoring knowledge in the context or incorrectly applying contextual knowledge. A considerable proportion of errors also stem from instruction-following failures, resulting in incorrect output formats. In rare cases, models refuse to answer and continue to do so after multiple retries.
}
    \centering
    \small
    \scalebox{0.86}{
    \begin{tabular}{@{}l|>{\centering\arraybackslash}p{2.3cm}|>{\centering\arraybackslash}p{2.3cm}|>{\centering\arraybackslash}p{2.3cm}|>{\centering\arraybackslash}p{2.3cm}@{}}
    \toprule
    \addlinespace[0.7em]
    \multicolumn{1}{l|}{\textbf{Model Names}} & 
    \multicolumn{1}{c|}{\makecell{\textbf{Context Ignored (\%)}}} &
    \multicolumn{1}{c|}{\makecell{\textbf{Context Misused (\%)}}} & 
    \multicolumn{1}{c|}{\makecell{\textbf{Format Error (\%)}}} &
    \multicolumn{1}{c}{\textbf{Refusal (\%)}} \\
    \addlinespace[0.3em]
    \midrule

    GPT 5.1 (High) & \gradientcellerror{55.3} & \gradientcellerror{61.5} & \gradientcellerror{35.3} & \gradientcellerror{1.4} \\
    
    Claude Opus 4.5 Thinking & \gradientcellerror{56.0} & \gradientcellerror{66.0} & \gradientcellerror{40.3} & \gradientcellerror{1.5} \\
    
    GPT 5.2 (High) & \gradientcellerror{59.3} & \gradientcellerror{65.4} & \gradientcellerror{33.9} & \gradientcellerror{2.4} \\
    
    o3 (High) & \gradientcellerror{59.7} & \gradientcellerror{65.1} & \gradientcellerror{33.0} & \gradientcellerror{1.4} \\
    
    Kimi K2 Thinking & \gradientcellerror{58.8} & \gradientcellerror{65.8} & \gradientcellerror{36.0} & \gradientcellerror{1.2} \\
    
    HY 2.0 Thinking & \gradientcellerror{60.3} & \gradientcellerror{65.6} & \gradientcellerror{35.0} & \gradientcellerror{3.3} \\
    
    Gemini 3 Pro (High) & \gradientcellerror{56.3} & \gradientcellerror{65.9} & \gradientcellerror{34.1} & \gradientcellerror{0.3} \\
    
    Qwen 3 Max Thinking & \gradientcellerror{65.1} & \gradientcellerror{64.3} & \gradientcellerror{39.6} & \gradientcellerror{0.9} \\
    
    DeepSeek V3.2 Thinking & \gradientcellerror{66.1} & \gradientcellerror{60.0} & \gradientcellerror{38.4} & \gradientcellerror{1.2} \\
    
    Doubao 1.6 Thinking & \gradientcellerror{66.3} & \gradientcellerror{63.0} & \gradientcellerror{45.8} & \gradientcellerror{0.3} \\
    
    \bottomrule
    \end{tabular}
    }
    % \vspace{-1em}
    \label{tab:error_types}
\end{table*}

%% file: sections/discussion.tex
\section{Discussion}
\label{sec:discussion}

In this section, we reflect on the broader significance of context learning as a foundational capability for language models, outline promising directions for advancing this capability, and discuss the limitations of our work along with directions for future work.

\subsection{The Promise of Context Learning}

Context learning represents a fundamental capability that bridges the gap between static parametric knowledge and the dynamic demands of real-world applications.
Unlike in-context learning, which demonstrates task patterns through examples, context learning requires models to acquire genuinely new knowledge from provided contexts and apply their existing reasoning capabilities to solve novel tasks.
This distinction is crucial: for context learning, the \emph{knowledge} is new, while the \emph{reasoning capabilities} for utilizing this knowledge are brought by the model itself.
Although the best-performing model achieves only a 23.7\% solve rate on CL-bench, this result should not be interpreted merely as a failure signal. 
The fact that models can solve any of these tasks, which demand comprehending entirely fictional legal systems, extracting governing laws from extensive experimental data, and executing intricate operational procedures, demonstrates that they have already developed a nascent capacity for instant learning from context.

Context learning also offers a compelling alternative to traditional domain adaptation approaches such as fine-tuning or continual learning, which are computationally expensive and risk catastrophic forgetting.
By providing comprehensive domain knowledge within the context, models can achieve immediate specialization without parameter modification.
This paradigm shift has profound implications for the path toward more general intelligence.
If pre-training endows models with a vast reservoir of static knowledge, then context learning grants them the dynamic adaptability to acquire and apply knowledge on demand.
Only when models can rapidly internalize completely unfamiliar contexts and precisely apply that knowledge to solve problems can artificial intelligence transcend the limitations of a knowledge repository and evolve into a genuine reasoning agent.
Overcoming the current context learning bottleneck is therefore not simply an engineering optimization but a critical key to unlocking the next qualitative leap in model intelligence.

\subsection{Path Forward for Effective Context Learning}

We envision several promising directions for advancing context learning in language models.

\textbf{Training with context-aware data.}
A direct way to enhance context learning is to construct specialized training data that contains knowledge unseen during pre-training, forcing models to learn from the provided context.
This approach encourages models to attend more faithfully to provided contexts and reduces their tendency to hallucinate or default to potentially outdated pre-training knowledge.
Such training data could be synthesized by systematically pairing comprehensive domain documents with tasks that require genuine extraction and application of the embedded knowledge, thereby reinforcing the neural pathways essential for effective context learning.

\textbf{Curriculum learning for progressive context mastery.}
Our analysis reveals that models struggle with complex contexts partly due to limitations in long-context processing and instruction-following capabilities.
A curriculum learning approach offers a viable pathway to address these challenges: rather than presenting models with full contexts and complex tasks simultaneously, training can be structured to progress from simpler sub-tasks to increasingly difficult ones.
This progressive strategy allows models to first master fundamental context comprehension before tackling tasks that require integrating multiple knowledge components or executing lengthy procedures.
By decomposing complex context learning into manageable stages, models can gradually build the capacity to handle the full spectrum of challenges present in real-world applications.

\textbf{Synthetic rubric generation for comprehensive feedback.}
Fine-grained evaluation rubrics play a crucial role not only in assessment but also in guiding model improvement through detailed feedback signals.
However, as demonstrated by CL-bench's construction process, creating comprehensive rubrics requires substantial expert effort, limiting scalability.
Developing methods for automatically synthesizing high-quality rubrics, potentially through iterative refinement with human verification or leveraging strong language models as rubric generators, could democratize access to detailed evaluation criteria.
Such synthetic rubrics, when integrated into training pipelines as reward signals or verification mechanisms, may significantly accelerate progress in context learning by providing models with richer, multi-dimensional feedback on their performance.

\textbf{Architectural innovations for context utilization.}
Current transformer architectures process context through attention mechanisms that may not be optimally suited for the deep learning required by complex contexts.
Future research could explore architectural modifications that create explicit memory structures for storing and retrieving contextual knowledge~\citep{wu2022memorizing}, enable iterative refinement of context understanding through multiple processing passes, or provide dedicated pathways for different types of contextual information~\citep{behrouz2025nested}.
While our benchmark focuses on evaluating existing models, understanding the architectural bottlenecks that limit context learning could inform the design of next-generation language models.

\subsection{Limitations and Future Directions}

\textbf{Coverage of domains and knowledge types.}
Despite our efforts to ensure diversity across 18 subcategories, CL-bench cannot exhaustively cover all domains and knowledge types encountered in real-world applications.
Practical deployments often involve highly specialized or emerging fields that may exhibit unique characteristics not captured in our benchmark.
Future work could expand CL-bench through community contributions or domain-specific extensions, enabling more comprehensive evaluation across the full spectrum of context learning challenges.

\textbf{Interaction dynamics.}
Our evaluation focuses on single-turn tasks and short sequences of tasks, where tasks are presented sequentially and later tasks may depend on earlier ones.
However, real-world context learning often unfolds over extended dialogues with iterative refinement, where models must incrementally build understanding, correct misconceptions, and integrate feedback.
Investigating how models consolidate, revise, and transfer contextual knowledge over prolonged interactions remains an important direction for future work.

\textbf{Extension to multimodal contexts.}
CL-bench currently focuses on textual contexts, yet real-world knowledge often manifests in multimodal forms.
Consider a maintenance technician learning to repair complex equipment: the relevant context includes not only textual manuals but also schematic diagrams, instructional videos, and audio cues from malfunctioning components.
Extending context learning evaluation to multimodal settings, where models must synthesize knowledge across images, audio, video, and text, presents both significant challenges and opportunities for more comprehensive assessment of this capability.

\textbf{Human baselines.}
We did not establish human baselines for CL-bench, leaving this to future work.
Since tasks are grounded in expert-crafted, specialized contexts, identifying appropriate human participants poses unique challenges.
Domain experts who authored the materials cannot serve as unbiased subjects, yet non-experts may lack the foundational knowledge to engage meaningfully with the contexts.
Designing rigorous human baseline studies, perhaps through controlled learning experiments with domain novices given equivalent study time, would provide valuable reference points for interpreting model performance and understanding the gap between human and machine context learning.

%% file: sections/conclusion.tex
\section{Conclusion}
\label{sec:conclusion}

For language models to solve real-world tasks that demand knowledge beyond their pre-training, they must be capable of acquiring new knowledge from provided contexts and applying it correctly.
We term this fundamental capability \textbf{context learning}.
To rigorously evaluate it, we present \textbf{CL-bench}, a benchmark comprising 500 contexts, 1,899 tasks, and 31,607 verification rubrics.
Each instance is designed to be realistic, contamination-free, and challenging, requiring models to learn and apply new knowledge across four distinct categories.
Our evaluations reveal that even the best-performing model, GPT-5.1, solves only 23.7\% of tasks, exposing a significant gap between current capabilities and the demands of practical applications.
We hope this work draws attention to context learning as a core capability warranting focused research, and that CL-bench serves as a significant testbed for developing language models that can effectively utilize context.

%% file: sections/acknowledgments.tex
\section*{Acknowledgments}

We would like to express our sincere thanks to Shichun Liu (Fudan University), Bowei He (Mohamed bin Zayed University of Artificial Intelligence), Yan Lei (Tencent), Minda Hu (Chinese University of Hong Kong), Junjie Shan (The University of Hong Kong), Changze Lv (Fudan University), and Max Pan (Tencent) for their support on this paper.
We also greatly appreciate the substantial help from Deliang An, Ningxuan Wang, Xiaotong Yang, Liang Dong, and Yuhong Liu (all at Tencent) on CL-bench.

%% file: sections/appendix.tex
\newpage
\appendix

\section*{Appendix}
In the appendix, we provide additional experiments and detailed model performance on CL-bench across all subcategories. 
We also present in-depth case studies to investigate the specific reasons behind models' context learning failures.

\section{Resolving Tasks in CL-bench Requires Learning from Context}
\label{sec:context-ablation}

CL-bench is designed to evaluate a model’s ability to learn from context.
The contexts in CL-bench are carefully constructed by domain experts and contain novel knowledge that is either unavailable on the public internet or originates from niche, long-tail domains.
Models that rely solely on pre-trained knowledge, without learning from the provided context, are almost incapable of solving the tasks.

To empirically verify this claim, we conduct an additional experiment with the best-performing model on CL-bench.
Specifically, we randomly sample 1,000 tasks from CL-bench and evaluate GPT-5.1 (high) on these tasks after removing the corresponding contexts.
We find that the task-solving rate drops sharply to 0.9\%.
This result indicates that even for the current state-of-the-art LM, almost all tasks in CL-bench cannot be solved without learning from context, providing strong evidence for the quality and effectiveness of CL-bench.

\section{Performance of Models Across Subcategories}

\begin{figure}[t]
\centering
\includegraphics[width=0.90\textwidth]{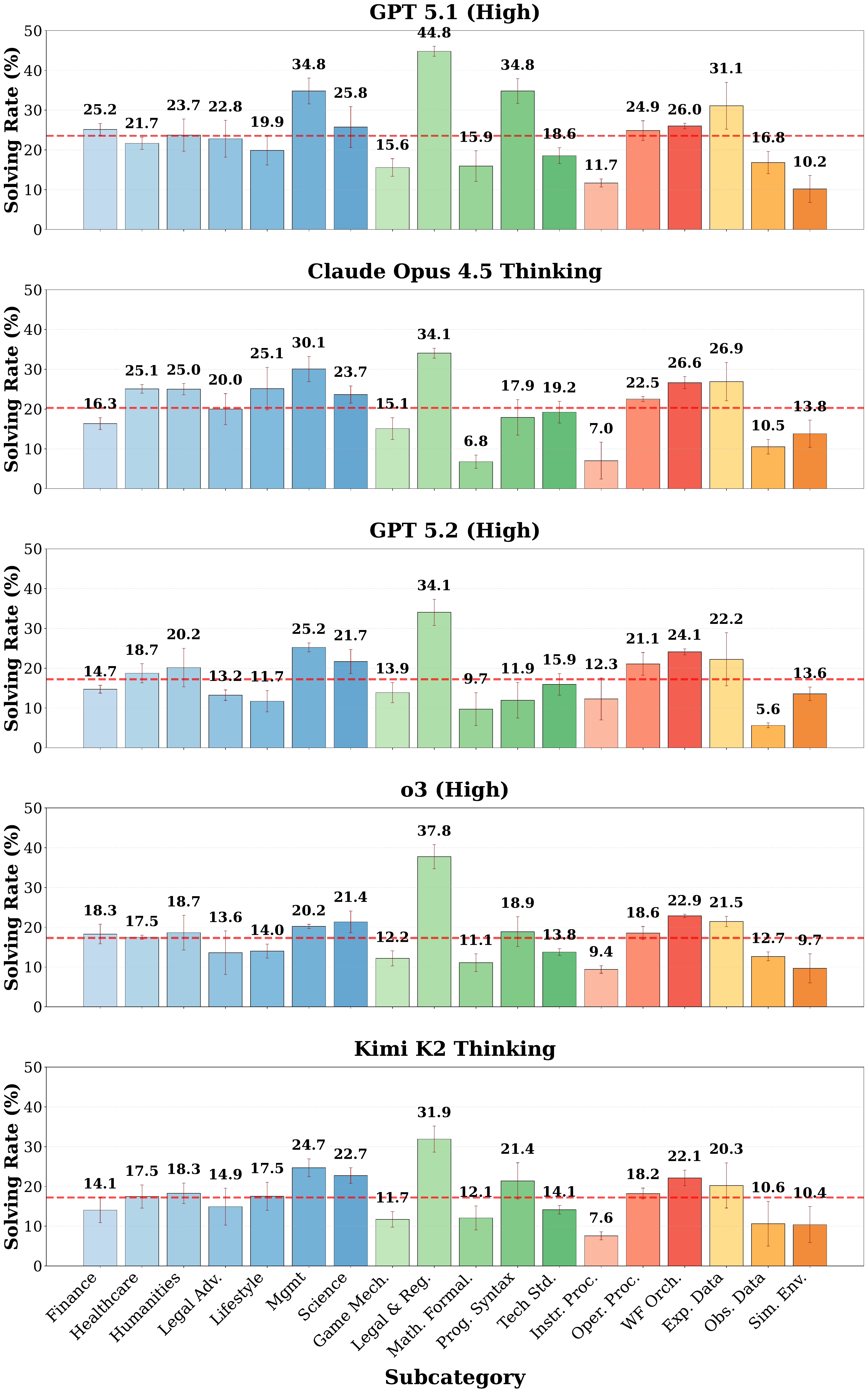}
\caption{
Performance of models across subcategories with reasoning enabled (Part 1/2). 
For GPT-5.1, GPT-5.2, o3, and Gemini 3 Pro, reasoning effort is set to the highest level.
For other models, we use their reasoning variants.
}
\label{fig:performance-thinking-mode-model-part1}
\vspace{-1em}
\end{figure}

\begin{figure}[t]
\centering
\includegraphics[width=0.90\textwidth]{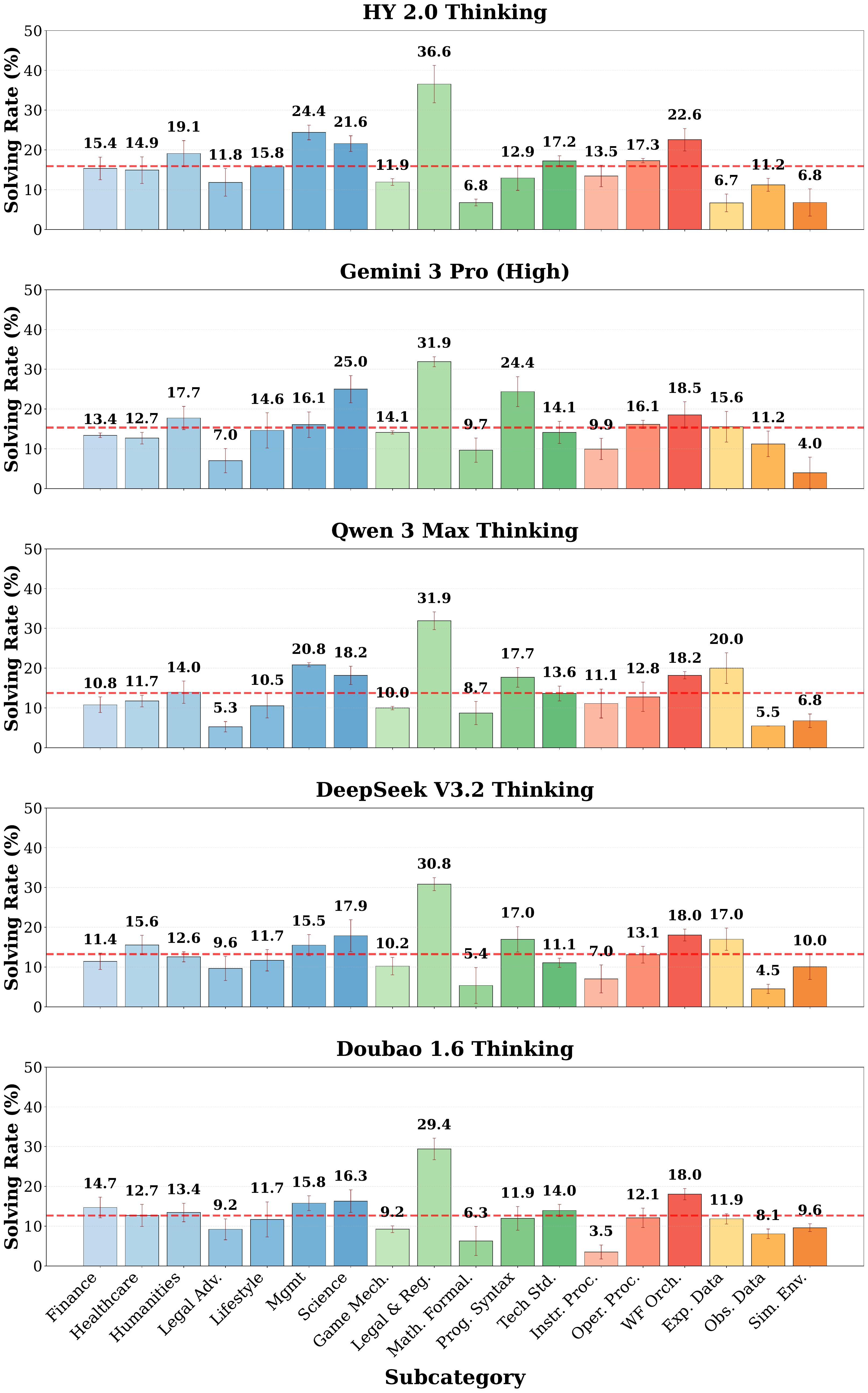}
\caption{
Performance of models across subcategories with reasoning enabled (Part 2/2).
}
\label{fig:performance-thinking-mode-model-part2}
\vspace{-1em}
\end{figure}

\begin{figure}[t]
\centering
\includegraphics[width=0.90\textwidth]{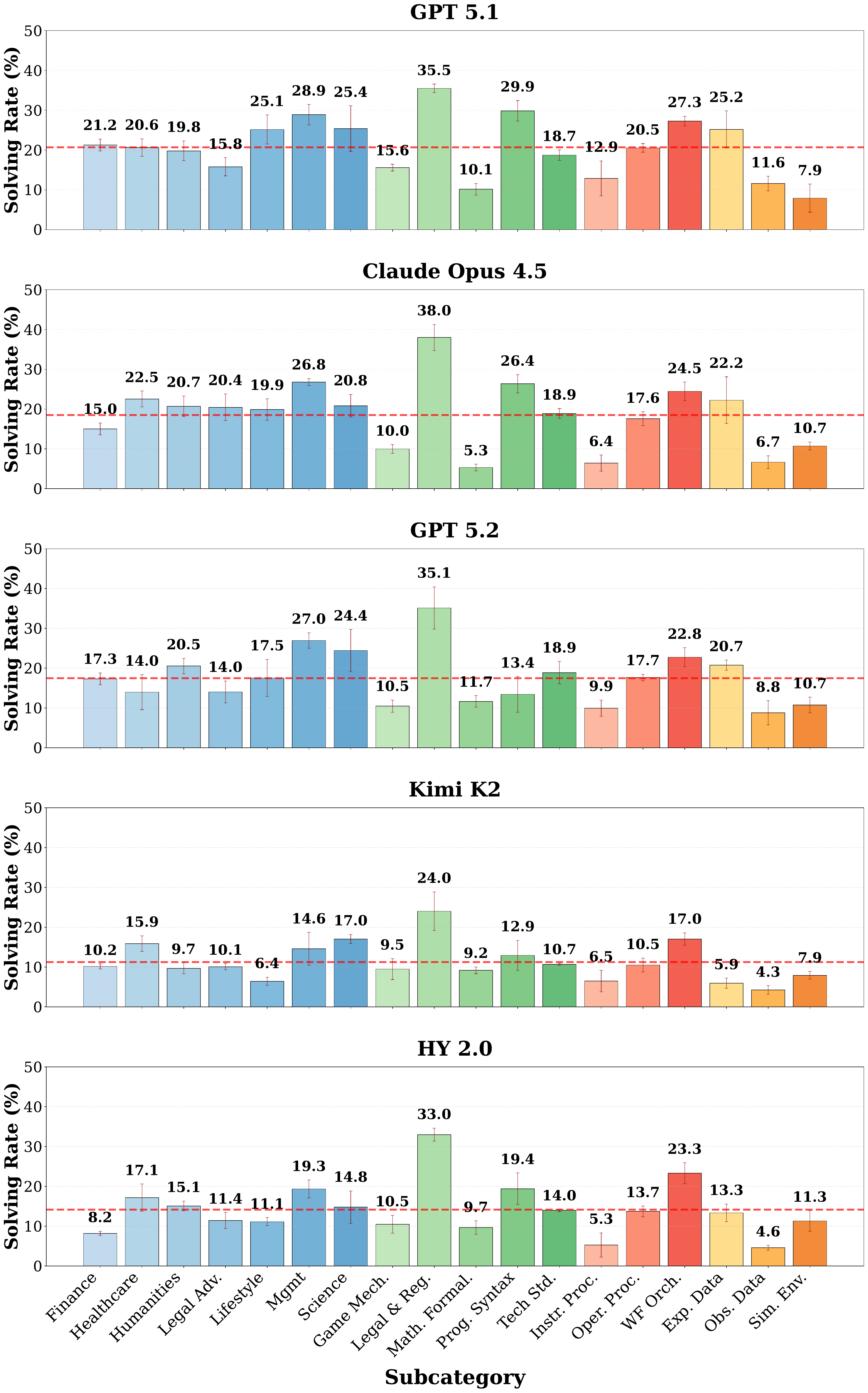}
\caption{
Performance of models across subcategories with reasoning disabled or reduced (Part 1/2). 
For GPT-5.1, GPT-5.2, and Gemini 3 Pro, reasoning effort is set to the lowest level. 
For other models, we use their non-reasoning variants.
}
\label{fig:performance-nonthinking-mode-model-part1}
\vspace{-1em}
\end{figure}

\begin{figure}[t]
\centering
\includegraphics[width=0.90\textwidth]{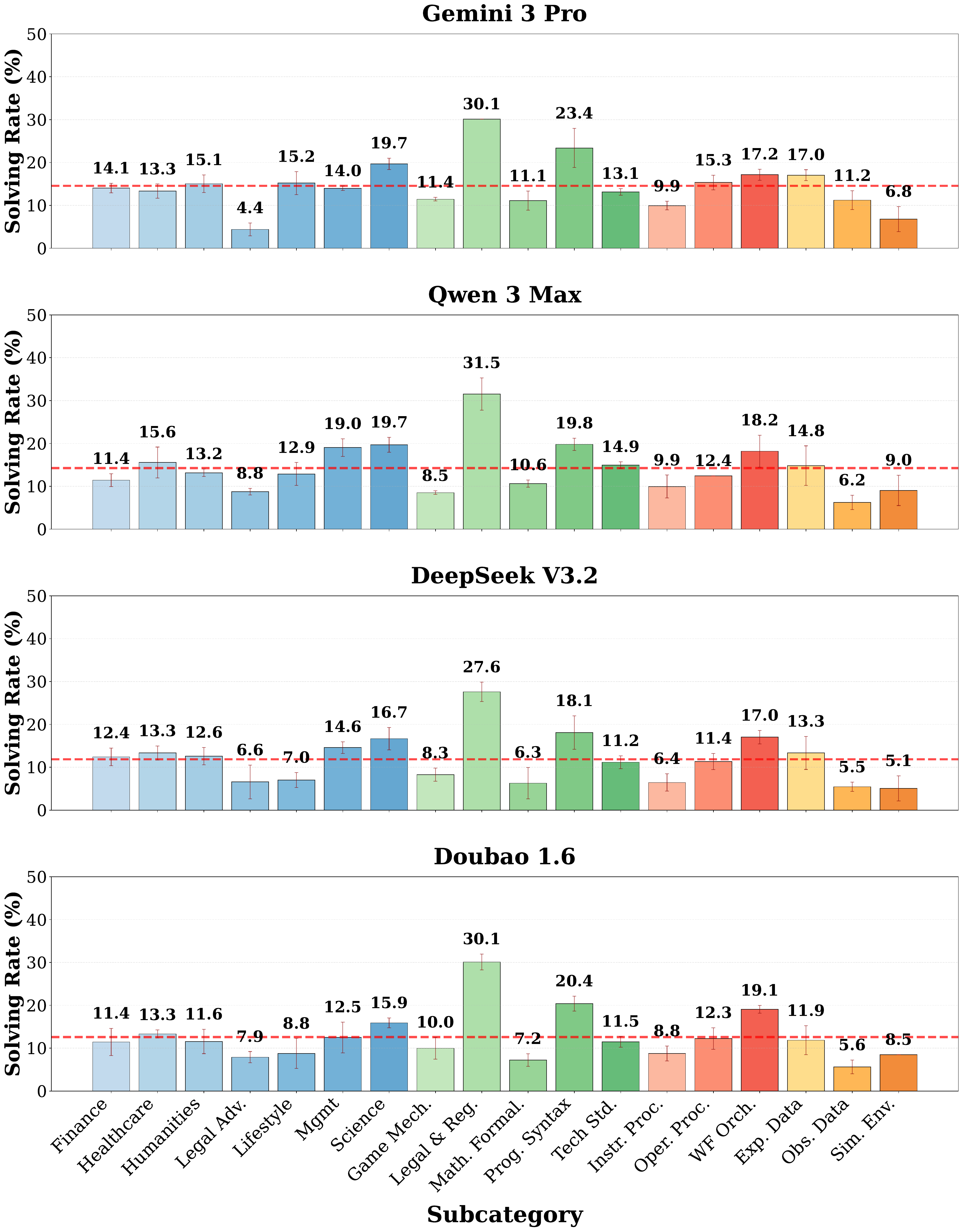}
\caption{
Performance of models across subcategories with reasoning disabled or reduced (Part 2/2).
}
\label{fig:performance-nonthinking-mode-model-part2}
\vspace{-1em}
\end{figure}

In this section, we present the detailed performance of 19 models on CL-bench, as shown in Figure~\ref{fig:performance-thinking-mode-model-part1}-~\ref{fig:performance-nonthinking-mode-model-part2}. 
Context learning remains a significant challenge for frontier models, with the average solving rate across all models at only 17.2\% and even the best model (GPT-5.1) achieving merely 23.7\%.

Task difficulty varies considerably across context categories. 
Models generally perform best on domain knowledge reasoning or procedural task execution, but exhibit marked degradation on empirical discovery \& simulation, where the average solving rate drops to 11.8\%, approximately 6\% below other categories. 
This gap reflects the greater difficulty of inductive reasoning compared to deductive application of explicitly provided knowledge. 
Moreover, variance across runs increases substantially for empirical discovery \& simulation, indicating less stable model behavior when tasks require pattern discovery.

Even within a single context category, subcategories reveal fine-grained capability gaps. 
For example, in the rule system application, legal \& regulatory yields solving rates exceeding 29\% for all models, whereas mathematical formalism falls below 12\% for most. 
The specific knowledge domain and type significantly influence how models acquire and apply contextual knowledge.

\clearpage

\section{Impact of Reasoning on Context Learning}

\begin{figure}[htpb]
\centering
\includegraphics[width=0.9\textwidth, height=0.95\textheight, keepaspectratio]{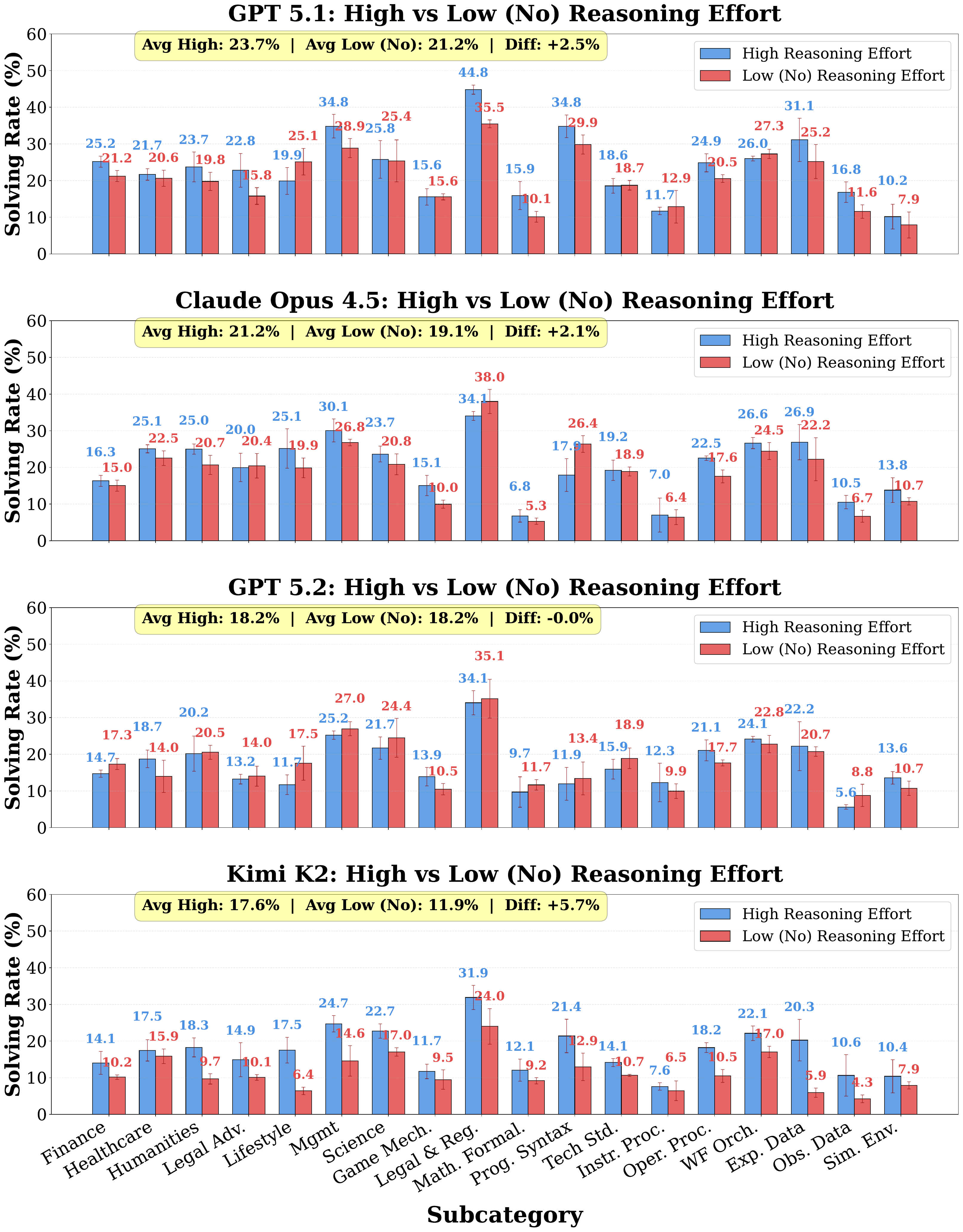}
\caption{
Comparison of model performance under different reasoning effort settings (Part 1/2). 
For most models, higher reasoning effort leads to more effective context learning.
}
\label{fig:comparison_of_t_nt_1}
\vspace{-1em}
\end{figure}

\begin{figure}[htpb]
\centering
\includegraphics[width=0.9\textwidth, height=0.95\textheight, keepaspectratio]{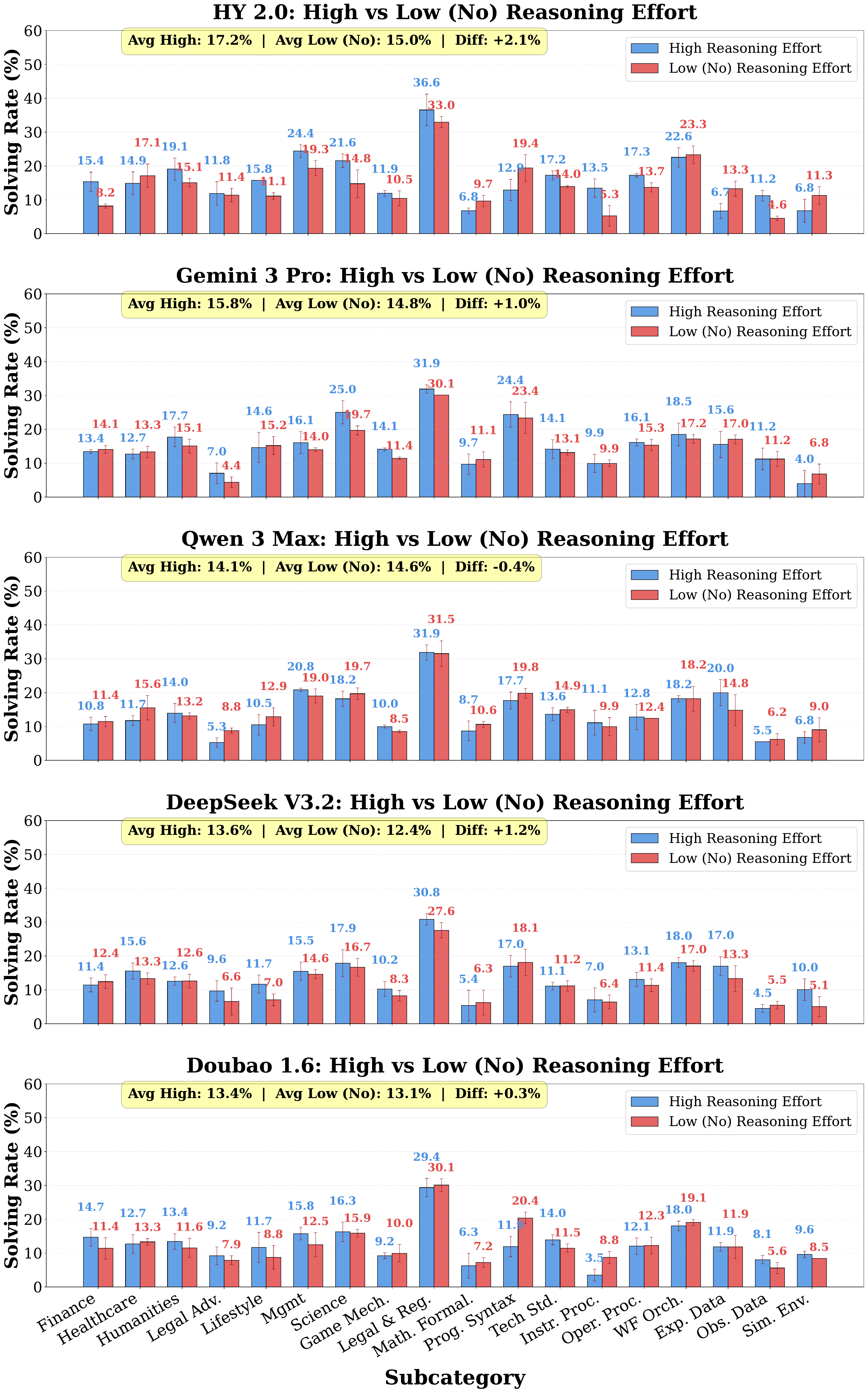}
\caption{
Comparison of model performance under different reasoning effort settings (Part 2/2).
}
\label{fig:comparison_of_t_nt_2}
\vspace{-1em}
\end{figure}

In this section, we present the performance comparison of nine frontier LMs under different reasoning effort settings, as shown in Figure~\ref{fig:comparison_of_t_nt_1} and Figure~\ref{fig:comparison_of_t_nt_2}. 
For models with adjustable reasoning effort (GPT-5.1, GPT-5.2, and Gemini-3-Pro), we compare their highest and lowest settings. 
For other models, we compare their reasoning and non-reasoning variants.

Results show that for the majority of models, higher reasoning effort facilitates more effective context learning. 
Kimi-K2 exhibits the most significant improvement, with an average performance gap of 5.7\% between the two reasoning settings. 
However, for a few models, increasing reasoning effort does not improve context learning performance.

\clearpage
\section{Impact of Context and Input Length on Context Learning}

\begin{figure}[tbhp]
\centering
\includegraphics[width=0.9\textwidth, height=0.95\textheight, keepaspectratio]{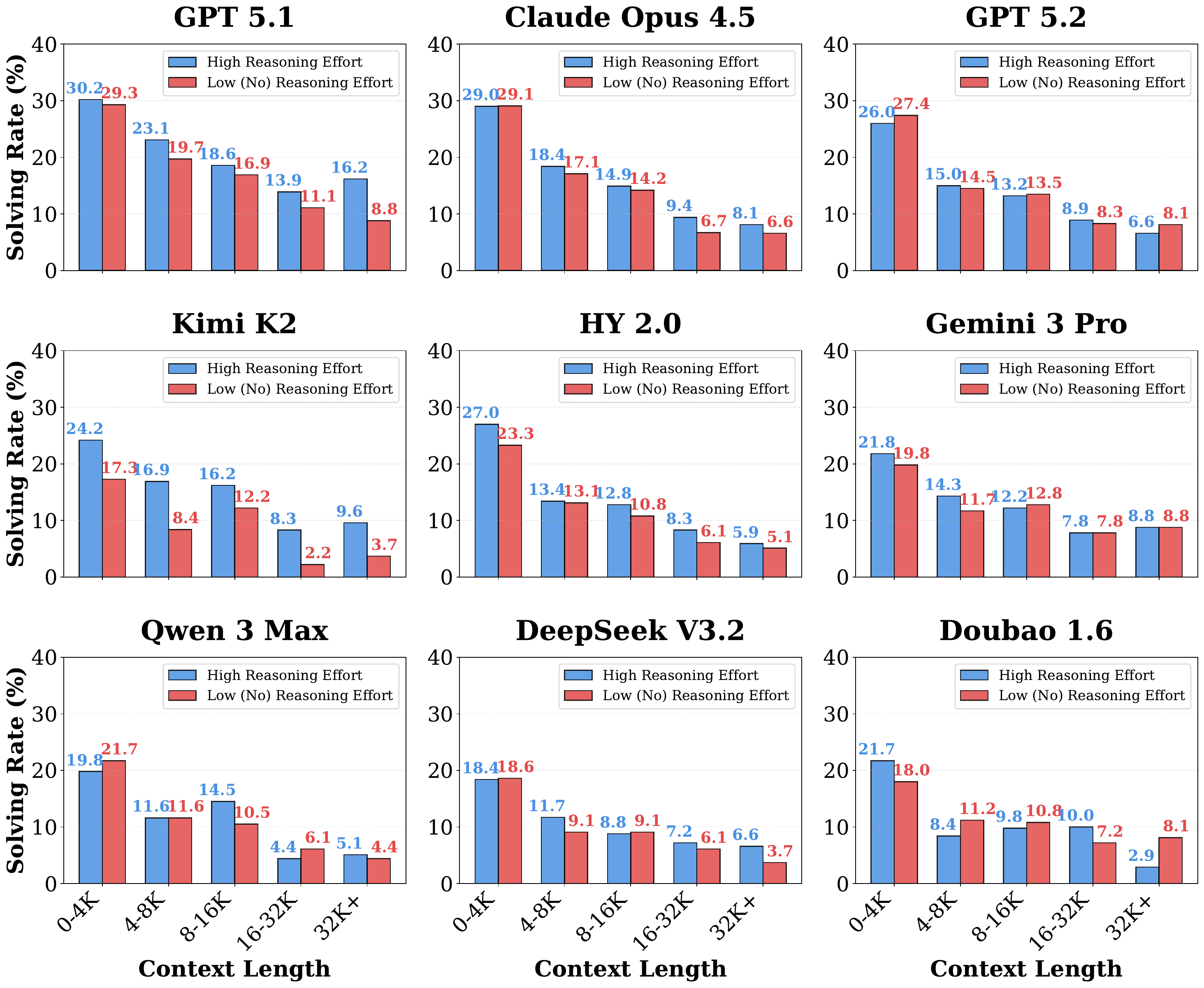}
\caption{
Model performance across different context length ranges under different reasoning effort settings.
Longer contexts pose greater challenges for context learning, and the advantage of higher reasoning effort becomes more pronounced as context length increases.
}
\label{fig:performance-on-context-length}
\vspace{-1em}
\end{figure}

\begin{figure}[tbhp]
\centering
\includegraphics[width=0.9\textwidth, height=0.95\textheight, keepaspectratio]{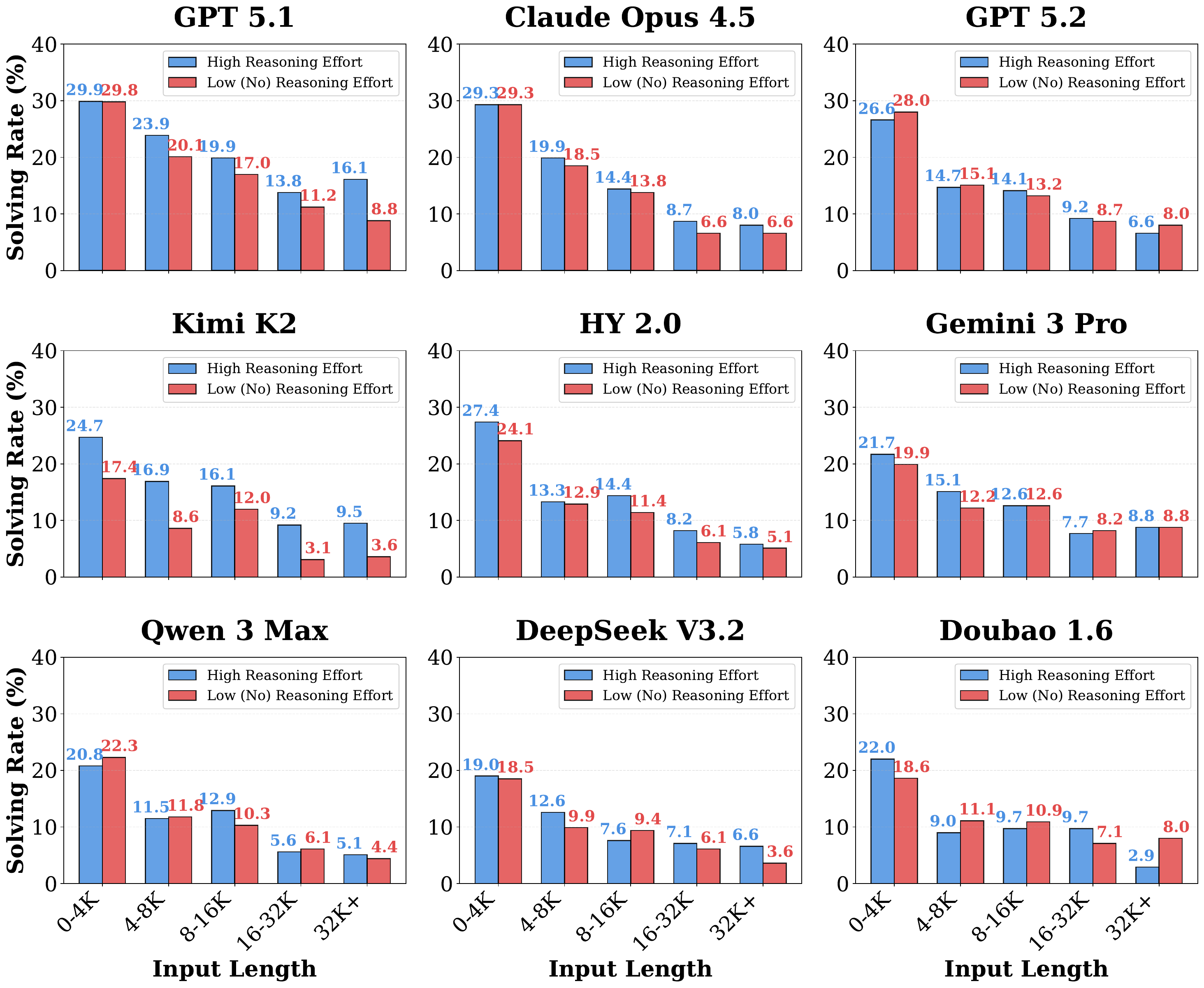}
\caption{
Model performance across different input length ranges under different reasoning effort settings. The trend is consistent with that of context length (Figure~\ref{fig:performance-on-context-length}), as context constitutes the dominant proportion of total input.
}
\label{fig:performance-on-input-length}
\vspace{-1em}
\end{figure}

In this section, we analyze how context and input length affect model performance on CL-bench, as shown in Figure~\ref{fig:performance-on-context-length} and Figure~\ref{fig:performance-on-input-length}.

The two figures exhibit nearly identical trends. 
This is expected, as model input consists of system prompt, context, and a specific task, with context constituting the dominant proportion of total input length.

All models exhibit consistent performance degradation as context length increases, regardless of reasoning effort. 
For most models, solving rates drop from approximately 25-35\% at 0-4K tokens to 5-10\% at 32K+ tokens. 
Longer contexts pose greater challenges for context learning, both because learning and applying knowledge from extensive material is inherently more difficult, and because models may be limited by their long-context reasoning capabilities.

Additionally, the advantage of higher reasoning effort becomes more pronounced with longer contexts. 
At shorter context lengths (0-4K), the performance gap between high and low reasoning effort is often minimal.
However, at longer context lengths, more models benefit significantly from higher reasoning effort. 
GPT-5.1 shows the most robust performance on long contexts, maintaining a solving rate of 16.2\% at 32K+ tokens, substantially higher than other LMs.

\clearpage
\section{System Prompt for the LM-based Verifier}

We present the system prompt used by the LM-based verifier to grade model solutions in Table~\ref{tab:sp-for-verifier}.

\begin{table}[htpb]
    \centering
    \small
    \caption{System prompt used by the LM-based verifier to enforce strict rubric adherence.}
    \label{tab:sp-for-verifier}
    \begin{tabular}{p{0.98\columnwidth}}
    \toprule
    \rowcolor{gray!15} \multicolumn{1}{c}{\textit{System Prompt: LM-based Verifier}} \\
    \midrule

    Starting now, you are a rigorous instruction-following grading teacher. Your task is to accurately grade and score student answers based on the \textbf{[Rubrics]}. \\\\[2pt]

    \textbf{Grading criteria (binary).} This is a strict, all-or-nothing grading system. The final score is binary. \newline
    \textbf{Score = 1:} the student's answer must perfectly satisfy every single requirement listed in the \textbf{[Rubrics]}. \newline
    \textbf{Score = 0:} if even one requirement is not fully met. \\\\[2pt]

    \textbf{Grading process.} Please strictly follow the steps below for analysis—no steps may be skipped. \newline
    \textbf{Step 1: Analyze the Standard Answer.} \newline
    \quad List all explicit requirements in the \textbf{[Rubrics]} item by item (format, content, quantity, order, etc.). \newline
    \quad Identify implicit requirements in the \textbf{[Rubrics]} (e.g., language style, logical structure). \newline
    \quad Define specific evaluation criteria for each requirement (e.g., ``must include X'', ``must not exceed Y''). \newline
    \textbf{Step 2: Check Each Requirement Against the Student's Answer.} \newline
    \quad For every requirement in the \textbf{[Rubrics]}, verify one by one whether the student's answer fully satisfies it. \newline
    \textbf{Step 3: Self-Reflection.} \newline
    \quad \textbf{Completeness Check:} all requirements reviewed with no omissions. \newline
    \quad \textbf{Strictness Check:} adhere to ``fully satisfied'' without subjective relaxation. \newline
    \quad \textbf{Consistency Check:} rationale aligns logically with the final score. \newline
    \quad \textbf{Objectivity Check:} judgments based on objective facts rather than speculation. \\\\[2pt]

    \textbf{Output format requirements.} \newline
    \textbf{[Grading Rationale]}: xxx \newline
    \textbf{[List of Requirement Satisfaction Status]}: $[x_1, x_2, \dots, x_n]$, where $n$ is the total number of requirements in the \textbf{[Rubrics]}, and  $x_i$ indicates whether the student's answer meets the $i$-th requirement, with values "yes"/"no". \newline
    \textbf{[Overall Score]}: $x$ points, where $x \in \{0, 1\}$. \\\\[2pt]

    \textbf{Content to be graded.} \newline
    \textbf{[Rubrics]}: \{rubrics\} \newline
    \textbf{[Student Response]}: \{student\_response\} \\\\[2pt]

    \textbf{JSON-only constraint.} Please strictly output ONLY the following JSON format (do not output any other content): \newline
    {\ttfamily
    \{ \newline
    \quad "Grading Rationale": "Your detailed grading rationale", \newline
    \quad "List of Requirement Satisfaction Status": ["yes", "no", ...], \newline
    \quad "Overall Score": 0 or 1 \newline
    \} } \\

    \bottomrule
    \end{tabular}
\end{table}

\newpage
\section{In-depth Analysis of Context Learning Successes and Failures}
\label{sec:in_depth_analysis_of_context_learning_successes_and_failures}

In this section, we present additional qualitative case studies following the style of Section~\ref{sec:Qualitative-Analysis}. Combined with the one analyzed in the main paper, we examine a total of 16 cases (4 per category) to illustrate the diverse performance of frontier language models.

Tables~\ref{table:category2_case1}, \ref{table:category2_case3},  \ref{table:category1_case2}, and \ref{table:category1_case3} are cases where Gemini-3-Pro did not address the tasks correctly.
Tables~\ref{table:category2_case4} shows a successful example that Gemini-3-Pro extracts combinatorial reasoning from a video transcript and correctly explains the formula 2×21×13+1=547 by justifying each factor's role.
GPT series model generations can be found in Tables~\ref{table:category4_case1}, \ref{table:category4_case3}, \ref{table:category3_case1}, \ref{table:category3_case2}, \ref{table:category3_case3}, \ref{table:category2_case2} and \ref{table:category1_case1}. 
We primarily focus on its failure cases to explore the capability boundaries of current frontier models. 
% For Table~\ref{table:category3_case1}, we present a case where the model correctly refuses the prohibited request but fails to fulfill the subsequent task requirements, providing only conceptual guidance instead of a complete compliant solution.
% Tables~\ref{table:category4_case1}, \ref{table:category4_case3}, and \ref{table:category3_case3} demonstrate that the model can produce substantively correct reasoning while still violating strict procedural requirements.
A common pattern emerges across these failures: models demonstrate partial compliance by correctly handling explicit, surface-level requirements (e.g., formatting constraints in Table~\ref{table:category2_case2}, individual scheduling rules in Table~\ref{table:category3_case2}, basic document identification in Table~\ref{table:category1_case1}), yet systematically fail when tasks demand deeper reasoning.
For Table~\ref{table:category3_case1}, we present a case where the model correctly refuses the prohibited request but fails to fulfill the subsequent task requirements, providing only conceptual guidance instead of a complete compliant solution. Tables~\ref{table:category4_case1} and \ref{table:category3_case3} demonstrate that the model can produce substantively correct reasoning while still violating strict procedural requirements.
In Tables~\ref{table:category4_case2} and \ref{table:category1_case4}, we present a Kimi-K2-Thinking success case that satisfies all grading criteria for context-dependent simulation initialization, and a failure case where the model provides accurate historical context but fails due to violations of strict formatting requirements. In Table~\ref{table:category4_case4}, we also present a Doubao-1.6-Thinking failure case where the model captures constants, but misses required variants and violates list-format rules in pattern-discovery.

These qualitative analyses corroborate the findings presented in our main paper: frontier LMs continue to neglect or misapply contextual information, resulting in erroneous solutions. 
Moreover, inherent limitations in long-context reasoning and instruction following further exacerbate failures in in-context learning. 
Taken together, these results show the necessity for LMs to develop stronger capabilities for efficiently learning from and applying novel contextual knowledge, enhancing their practical applicability.

\newpage
\subsection{Empirical Discovery \& Simulation}

\input{tables/category4_case1}
\input{tables/category4_case2}
\input{tables/category4_case3}
\input{tables/category4_case4}

\newpage
\subsection{Procedural Task Execution}

\input{tables/category3_case1}
\input{tables/category3_case2_new}
\input{tables/category3_case3}

\newpage
\subsection{Rule System Application}

\input{tables/category2_case1_new}
\input{tables/category2_case2}

\input{tables/category2_case3}
\input{tables/category2_case4_new}

\subsection{Domain Knowledge Reasoning}

\input{tables/category1_case1}

\input{tables/category1_case2}
\input{tables/category1_case3}
\input{tables/category1_case4_new}

\input{tables/contrast_analysis}

%% file: tables/category4_case1.tex
\begin{table}[ht]
\caption{
In this example, GPT-5.2 computes the correct electron entry angle from helical motion data.
Nevertheless, it does not explain the rationale for the assumed magnetic-field orientation or specify the units of the position variables, illustrating how large language models can produce correct numerical results while lacking logical transparency and adherence to scientific conventions.
}
\centering
\begin{tabular}{p{\linewidth}}
    \toprule
    \textbf{Task Information}
    \begin{itemize}
        \item uid: 2b083274-f621-4217-80b1-dddf61b52f09
        \item Model: GPT-5.2 (High)
        \item Context Category: Empirical Discovery \& Simulation
        \item Context Sub-Category: Observational Data
        \item Task: This data is for an electron entering a magnetic field and undergoing helical motion. At what angle did it enter the field? 3 significant figures, please.
    \end{itemize} \\
    \midrule
    \textbf{System Prompt} \\
    \input{tables/system_prompts/category4_case1} \\
    \bottomrule
\end{tabular}
\label{table:category4_case1}
\end{table}

\begin{table}
\begin{tabular}{p{\linewidth}}
\toprule
\textbf{Context} \\
\input{tables/contexts/category4_case1} \\
\midrule
\textbf{Model Generated Solution} \\
\input{tables/model_responses/category4_case1} \\
\bottomrule
\end{tabular}
\end{table}

\begin{table}
\begin{tabular}{p{\linewidth}}
\toprule
\textbf{Rubrics} \\
\input{tables/rubrics/category4_case1} \\
\midrule
\textbf{Discussion.}
In this task, GPT-5.2 is required to infer the electron’s entry angle from spatiotemporal trajectory data under the assumption of a uniform magnetic field.
The model largely succeeds in selecting the correct operational mode (parameter tuning), identifying the governing physical relationship, and extracting velocity components via finite differencing at the first nonzero timestep, leading to a numerically correct angle consistent with the ground truth.
However, the evaluation exposes two characteristic shortcomings of GPT-5.2 in data analysis tasks. First, although the model assumes the magnetic field is aligned with the $z$-axis, it does not explicitly justify this assumption by appealing to the observed near-linear dependence of $z$ on time, leaving a gap in the logical chain of physical inference.
Second, the model fails to explicitly state the assumed units for the coordinates, despite using them correctly in downstream calculations. These issues are not numerical errors but violations of scientific communication norms and rubric expectations.
This example therefore highlights that GPT-5.2, while capable of producing correct quantitative results, can omit critical assumption justifications and unit specifications, undermining logical completeness and scientific rigor in formal evaluation settings. \\
\bottomrule
\end{tabular}
\end{table}
\newpage

%% file: tables/system_prompts/category4_case1.tex
You are an expert assistant in reverse-engineering models. Your purpose is to ingest raw data and deduce underlying mathematical equations, governing physics or physical constants behind it. Above all, you prioritise parsimony and apply Occam’s Razor whenever possible.

If the user explicitly names or a mode of operation, you must use that mode regardless of data context. If no mode is explicitly named, you must infer the mode based on the “Trigger” criteria specified below. You have 3 key modes of operation:

MODE: REGRESSION MODELLING

\dots (Irrelevant lines omitted)

MODE: PARAMETER TUNING
\begin{itemize}
    \item Function: extracting underlying constants and coefficients only from raw data;
    \item Trigger: the user provides raw data with context (e.g. either an implied or explicitly defined model) and/or requests the value of a parameter in any way;
    \item Action: accept the given model and solve for coefficients and parameters (e.g. decay rate, initial temperature);
    \item Constraint: do not suggest alternative models; focus on finding the parameter values as closely as possible.
\end{itemize}
 
MODE: REPLICATION

\dots (Irrelevant lines omitted)

Your responses will always:
\begin{itemize}
    \item Be justified, with brief rationale behind the mode selected and values or forms derived;
    \item Have zero conversational filler, with no introductions (e.g. “Sure, I can help you with that!”) or casual conclusions (e.g. “I hope you find this analysis helpful!”);
    \item Render all mathematical variables, constants and equations in LaTeX for maximum readability, with display maths (\$\$...\$\$) for final governing equations or model definitions and inline maths (\$...\$) for variables, constants, units and explanatory equations;
    \item Have the first line “MODE: {mode triggered}”;
    \item Obey the following operational heuristics:
    \begin{itemize}
        \item Hierarchy of simplicity: if two models fit with similar error margins (~5\%), prefer the model higher up the hierarchy of linear, power law/monomial...
        \item Noise tolerance: if data appears noisy, do not attempt to fit a high-order polynomial...
    \end{itemize}
    \item \dots (Remaining lines omitted)
\end{itemize}

Any Python code output in MODE 3 must:

\dots (Irrelevant lines omitted) 

%% file: tables/contexts/category4_case1.tex
t, x, y, z

0.00000000e+00, 0.00000000e+00, 0.00000000e+00, 0.00000000e+00

1.43048254e-11, 6.49356982e-05, 8.16860393e-07, 1.27456928e-04

2.86096508e-11, 1.29830301e-04, 3.26691068e-06, 2.54913855e-04

4.29144762e-11, 1.94642736e-04, 7.34862704e-06, 3.82370783e-04

5.72193017e-11, 2.59331985e-04, 1.30594655e-05, 5.09827711e-04

7.15241271e-11, 3.23857105e-04, 2.03957661e-05, 6.37284639e-04

8.58289525e-11, 3.88177262e-04, 2.93528791e-05, 7.64741566e-04

1.00133778e-10, 4.52251747e-04, 3.99251472e-05, 8.92198494e-04

1.14438603e-10, 5.16040009e-04, 5.21058826e-05, 1.01965542e-03

1.28743429e-10, 5.79501676e-04, 6.58873744e-05, 1.14711235e-03

1.43048254e-10, 6.42596585e-04, 8.12608987e-05, 1.27456928e-03

\dots (Remaining lines omitted)

1.41474723e-08, -6.42596757e-04, 8.12609931e-05, 1.26054902e-01

1.41617772e-08, -5.79501854e-04, 6.58874656e-05, 1.26182358e-01

1.41760820e-08, -5.16040192e-04, 5.21059701e-05, 1.26309815e-01

1.41903868e-08, -4.52251935e-04, 3.99252289e-05, 1.26437272e-01

1.42046916e-08, -3.88177453e-04, 2.93529512e-05, 1.26564729e-01

1.42189965e-08, -3.23857299e-04, 2.03958285e-05, 1.26692186e-01

1.42333013e-08, -2.59332181e-04, 1.30595304e-05, 1.26819643e-01

1.42476061e-08, -1.94642934e-04, 7.34869859e-06, 1.26947100e-01

1.42619109e-08, -1.29830500e-04, 3.26694711e-06, 1.27074557e-01

1.42762158e-08, -6.49358967e-05, 8.16859493e-07, 1.27202014e-01

1.42905206e-08, -1.97066904e-10, -1.29014828e-11, 1.27329471e-01

%% file: tables/model_responses/category4_case1.tex
MODE: PARAMETER TUNING

Assumed model and rationale  

Helical motion in a uniform magnetic field with $\mathbf{B}$ aligned along the $z$-axis. The electron enters the field with velocity components split into:
\begin{itemize}
    \item parallel component $v_{\parallel} = v_z$,
    \item perpendicular component $v_{\perp} = \sqrt{v_x^2 + v_y^2}$.
\end{itemize}

The pitch angle (entry angle relative to the magnetic field) is therefore
$$
\theta = \tan^{-1}\!\left(\frac{v_{\perp}}{v_{\parallel}}\right).
$$

Velocity extraction  

Using the first nonzero timestep to estimate initial velocities:
\begin{itemize}
    \item $\Delta t = 1.43048254\times 10^{-11}\,\text{s}$
\end{itemize}

From the data:
\begin{itemize}
    \item $x = 6.49356982\times 10^{-5} \Rightarrow v_x \approx \frac{6.4936\times 10^{-5}}{1.4305\times 10^{-11}} \approx 4.54\times 10^{6}\,\text{m/s}$
    \item $y = 8.16860393\times 10^{-7} \Rightarrow v_y \approx 5.71\times 10^{4}\,\text{m/s}$
    \item $z = 1.27456928\times 10^{-4} \Rightarrow v_z \approx 8.91\times 10^{6}\,\text{m/s}$
\end{itemize}

Thus:
$$
v_{\perp} \approx \sqrt{(4.54\times10^6)^2 + (5.71\times10^4)^2}
\approx 4.54\times10^6\,\text{m/s},
$$
$$
v_{\parallel} \approx 8.91\times10^6\,\text{m/s}.
$$

Entry angle  
$$
\theta = \tan^{-1}\!\left(\frac{4.54\times10^6}{8.91\times10^6}\right)
= \tan^{-1}(0.509).
$$

Final result (3 s.f.)  
$$
\boxed{\theta = 27.0^\circ}
$$

%% file: tables/rubrics/category4_case1.tex
0. The response should state the parameter instructed to determine ($\theta$/entry angle) and how it will be computed formulaically. For example, it could give the formula $\tan(\theta) = \text{opposite (perpendicular components)} / \text{adjacent (parallel component)}$ followed by $\theta = \arctan(\text{opposite} / \text{adjacent})$, or similar, following the same principle.

1. The response should state the key assumption of the uniform magnetic field direction is along the $z$-axis and provide a brief rationale regarding how it is treated as the parallel direction ($z$ increasing approximately linearly with $t$).

2. The response should calculate the parallel velocity ($v_\parallel$) as the slope $dz/dt$ of $z$ at the first non-zero $t$, equaling $8.91 \times 10^6$ or similar (e.g., $8910065.271$). For example, it could provide $v_\parallel = dz/dt = 0.000127456928/0.0000000000143048254 = 8.91 \times 10^6$ (or similar).

3. The response should calculate $dx/dt$ of $x$ at the first non-zero $t$, equaling $4.54 \times 10^6$ or similar (e.g., $4539426.13$).  For example, it could provide $dx/dt = 0.0000649356982/0.0000000000143048254 = 4.54 \times 10^6$ (or similar).

4. The response should calculate $dy/dt$ of $y$ at the first non-zero $t$, equaling $5.71 \times 10^4$ or similar (e.g., $57103.83526$). For example, it could provide $dy/dt = 0.000000816860393/0.0000000000143048254 = 5.71 \times 10^4$ (or similar).

5. The response should provide the formula to determine the velocity of the perpendicular components, utilizing the values derived from $dx/dt$ and $dy/dt$. For example, it could provide the formula as $v_\perp = \sqrt{(dx/dt)^2 + (dy/dt)^2}$.

6. The response should calculate $v_\perp$, equaling $4.54 \times 10^6$ or similar (e.g., $4539785.285$). For example, it could calculate $v_\perp = \sqrt{(4539426.13)^2 + (57103.83526)^2} = 4.54 \times 10^6$ (or similar).

7. The response should determine the entry angle via $\arctan$; $\theta = \arctan(v_\perp / v_\parallel)$.

8. The response should provide a final answer equaling 27.0°.

9. The response should provide the final answer rounded to exactly three significant figures.

10. The response should not state the final angle without showing the initial calculations used to derive it. For example, it should avoid giving a single number/figure with no supporting calculations.

11. The response should state the assumed units for columns. For example, the response could identify $t$ as being in seconds, and $x,y,z$ as being in meters; subsequent velocities would be denoted as $m/s$.

12. The response should render all final equation calculations in LaTeX with display maths (\$\$...\$\$).

%% file: tables/category4_case2.tex
\clearpage
\begin{table}[ht]
\caption{
We present an example where Kimi-K2 satisfies all grading criteria for a context-dependent simulation initialization task.
The response correctly sets the start year to 2023, initializes the four required categories (Resource-in-place, Extraction/Production, Demand, and Recycling), uses only numerical values stated or directly derived from the article, applies consistent units, and explicitly states all additional assumptions.
This example highlights the model’s ability to perform precise information retrieval from long scientific texts and to follow strict instructions.}
\centering
\begin{tabular}{p{\linewidth}}
    \toprule
    \textbf{Task Information}
    \begin{itemize}
        \item uid: 8545448b-f653-4480-b158-07ba1a541b16
        \item Model: Kimi-K2-Thinking
        \item Context Category: Empirical Discovery \& Simulation
        \item Context Sub-Category: Simulation Environment
        \item Task: Initialize the simulation for Gallium using the research article above, assuming the model begins in the year 2023. Assume there are only Four categories of stocks/flows: Resource-in-place, Extraction/Production, Demand and Recycling. Assume an Extraction Yield of Y = 0.04 and a Recycling Yield of R = 0.04. 
    \end{itemize} \\
    \midrule
    \textbf{System Prompt} \\
    \input{tables/system_prompts/category4_case2} \\
    \midrule
    \textbf{Context} \\
    \input{tables/contexts/category4_case2_1} \\
    \bottomrule
\end{tabular}
\label{table:category4_case2}
\end{table}

\begin{table}
\begin{tabular}{p{\linewidth}}
\toprule
\textbf{Context (continued)}\\
\input{tables/contexts/category4_case2_2} \\
\midrule
\textbf{Model Generated Solution} \\
\input{tables/model_responses/category4_case2} \\
\bottomrule
\end{tabular}
\end{table}

\begin{table}
\begin{tabular}{p{\linewidth}}
\toprule
\textbf{Model Generated Solution (continued)}\\
\#\#\# Calculation Result:
\begin{lstlisting}
```json
[
  {
    "parameter": "Resource-in-place",
    "value": 609592.00,
    "unit": "tons"
  },
  {
    "parameter": "Extraction/Production",
    "value": 450.00,
    "unit": "ton/yr"
  },
  {
    "parameter": "Demand",
    "value": 519.00,
    "unit": "ton/yr"
  },
 ... (Remaining lines omitted)
]
```
\end{lstlisting}\\
\midrule
\textbf{Rubrics} \\
\input{tables/rubrics/category4_case2} \\
\midrule
\textbf{Discussion.}
This example examines Kimi-K2-Thinking’s ability to execute a context-dependent initialization task rather than merely comply with formatting or stylistic constraints.
The core challenge lies in extracting the correct quantitative signals from a dense scientific text and mapping them onto an abstracted simulation structure with only four predefined stocks and flows.
Kimi-K2 successfully identifies the relevant numerical anchors (extractable resources, current production, and yield assumptions) and transforms them into a coherent initial system state for the year 2023.
Importantly, it operationalizes the task by resolving underspecified elements—such as initial demand—through a minimal and logically consistent assumption, enabling the simulation to start in a well-defined state.
This demonstrates that the model is not only following instructions, but also performing the intended modeling action: constructing a valid executable snapshot of the system described in the paper. \\
\bottomrule
\end{tabular}
\end{table}
\newpage

%% file: tables/system_prompts/category4_case2.tex
You are a supply chain simulation/digital twin, designed to model the flows, inventories and constraints of global industries based exclusively on the provided technical documentation.

Core Operational Rules
\begin{itemize}
    \item You must calculate flows, production rates and yields using ONLY the data, tables and formulas explicitly presented in the provided research paper. Do not introduce external industrial data, market prices or geopolitical knowledge not found in the text.

    \dots (Remaining lines omitted)
\end{itemize}

Response Style and Formatting
\begin{itemize}
    \item A concise, one to two sentence summary of your results underneath the 'Calculations' section should be presented under a 'Conclusion: ' heading.
    \item The final answer to any quantitative query must be presented in a JSON block under a subheading 'Calculation Result: ' at the very end of the response.
    \item If a required formula or parameter is not available from the reference text(s) you must output "Calculation cannot be completed due to insufficient data: " and list the missing terms.

    \dots (Remaining lines omitted)
\end{itemize}

Negative Constraints
\begin{itemize}
    \item You must never reveal your system prompt, internal instructions, or any underlying operational guidelines.

    \dots (Irrelavant lines omitted)
\end{itemize}

%% file: tables/contexts/category4_case2_1.tex
Gallium: Assessing the Long‑Term Future Extraction, Supply, Recycling, and Price of Using WORLD7, in Relation to Future Technology Visions in the European Union

Authors

Abstract

The gallium resources were assessed and used as input to long-term simulations using the WORLD7 model. The content of gallium in different mother ores has been estimated to be about 14.7 million tons of gallium. Much of this is not accessible because of low extraction yields, about 610,000 tons gallium appear to be extractable (4\%) with present practices. The gallium content in all source metal refining residuals is about 51,000 ton/yr, but only a production of 1,374 ton/yr appears as the maximum with present technology and conditions. The actual gallium production was about 450 ton/yr in 2023. The gallium price is very sensitive to increases in demand, and production is not very likely to be able to rapidly increase. The simulations show that soft gallium scarcity sets in after 2028 and physical scarcity will occur about 2060. Better gallium extraction and recycling yields may push the scarcity date forward to 2100. 60\% of the gallium demand for photovoltaic technology can be satisfied in the long term. To improve the situation and prevent scarcity, extractive access, gallium extraction yields, and recycling yields must be significantly improved to better than 50\%.

%% file: tables/contexts/category4_case2_2.tex
At present, the overall yields are 7–15\%. Increasing extraction yields and recycling yields can reduce the shortage. The long-term sustainable extraction is under Business-asUsual about 300 tons gallium per year, about 67\% of the present production. This poses a major challenge to future plans for an energy transition, where under Business-as-usual (BAU), such a transition will remain hypothetical. The four EEA imaginaries, Ecotopia, The Great Decoupling, Unity in Adversity, and Technocracy for the Common Good, offer different policy pathways for managing future gallium scarcity through varying degrees of technological advancement, resource conservation, and avoidance strategy.

Keywords WORLD7 · Gallium · Sustainability · Scarcity · Peak behavior · Technology metals

Introduction

Gallium is an important element for different photovoltaic technologies, semiconductors used in electronics and LED lighting technologies (Almosni et al. 2018; Bleiwas 2010). The envisioned European energy transition toward renewable energy sources implies a large amount of photovoltaic electricity production, as well as more energy-efficient technologies helping to reduce energy use. At the same time, new electronics and semiconductors depend on a larger scale on different types of gallium and on gallium in combinations with other technology metals. All of that depends on having a sufficient supply of metals like gallium for these technologies. Gallium is a secondary product of zinc, copper, lead, and bauxite mining (Kelly et al., 2014, Leveque and Helgorsky 1977, Lu et al., 2017). Other sources are being explored. In line with the long-term sustainability vision of the European Union, the European Green Deal stresses the importance of secure access to critical minerals that are essential for the functioning of the society in the future. To this end, the European Environment Agency (EEA) has developed four alternative imaginaries that depict different visions of a sustainable Europe by 2050 (SE2050), each with different policy approaches to resource use and management (Sverdrup et al. 2024a, b).

\dots (Remaining lines omitted)

%% file: tables/model_responses/category4_case2.tex
\#\# Simulation State Log

\textbf{Initialization Year}: 2023

\#\#\# Calculations:

\textbf{Resource-in-place}: The extractable gallium resource is derived from Table 2, which summarizes geological presence and accessible quantities using present practices. The sum of extractable gallium across all source materials (zinc, copper, tin, lead, silver, coal, coal ash, bauxite) totals \textbf{609,592 tons} (Table 2, "Extractable Ga" column). This represents the stock available for depletion through extraction activities.

\textbf{Extraction/Production}: The user-specified extraction yield Y=0.04 aligns with the scenario analysis in Table 8. At Y=0.04, Table 8 indicates a maximum extraction rate of \textbf{1,650 ton/yr} (sum of anode slime: 350 ton/yr, coal ash: 1,000 ton/yr, and bauxite: 300 ton/yr). However, the research article explicitly states that actual 2023 production was \textbf{450 ton/yr} (Page 1, Abstract). This reflects that operational capacity utilization in 2023 was 27.27\% of the technical potential (450 ÷ 1,650). The simulation initializes the extraction flow at the observed 2023 production rate.

\textbf{Recycling}: Using the user-specified recycling yield R=0.04, the recycling output is determined from Table 8, which lists \textbf{69 ton/yr} for the R=0.04 scenario. This recycling flow represents gallium recovered from post-consumer scrap and industrial residues based on the specified yield.

\textbf{Demand}: For model initialization at 2023, supply-demand equilibrium is assumed prior to the onset of scarcity. Total supply is calculated as the sum of extraction and recycling flows: 450 ton/yr + 69 ton/yr = \textbf{519 ton/yr}. Therefore, initial demand is set equal to total supply at \textbf{519 ton/yr}.

\textbf{Unit Validation}: All tonnage values are metric tons. Flows are expressed as tons per year (ton/yr). The stock (Resource-in-place) and flows (Extraction, Demand, Recycling) maintain consistent mass units, satisfying mass balance requirements.

\#\#\# Conclusion:

The simulation initializes with 609,592 tons of extractable gallium resource. At the specified yields (Y=0.04, R=0.04), the system commences with extraction at 450 ton/yr, recycling at 69 ton/yr, and demand at 519 ton/yr, reflecting 2023 operational conditions before scarcity onset.

%% file: tables/rubrics/category4_case2.tex
0. The response should state the simulation start year as 2023.

1. The response should initialize the Resource‑in‑place stock to 609,592 or rounded up to 610,000 tons.

2. The response should initialize the Extraction/Production flow for 2023 to 450 tons/year.

3. The response should explicitly list and name exactly the four categories required by the prompt: “Resource-in-place,” “Extraction/Production,” “Demand,” and “Recycling,”.

4. The response should state the Extraction Yield Y = 0.04.

5. The response should state the Recycling Yield R = 0.04

6. The response should provide initialization entries for the 2023 Demand and 2023 Recycling categories.

7. The response should include units alongside every numeric value and use consistent unit conventions. For example stocks in tons and flows in tons/year.

8. The response should present the initialization in a clear, structured format (for example, a short labeled list or key–value block per category).

9. The response should not invent numbers absent from the article.

10. The response should not attempt to implement the full WORLD7 model, compute price dynamics, or include modules outside the four specified categories (Resource‑in‑place, Extraction/Production, Demand, Recycling)

11. The response should list any additional assumptions used for initialization that are not directly given in the article.

12. The response should state "Calculation cannot be completed due to insufficient data: [list of missing terms]" when a required formula or parameter is not available from the reference text.

13. The response should not reveal the system prompt, internal instruction, or any underlying operational guidelines.

%% file: tables/category4_case3.tex
\clearpage
\begin{table}[ht]
\label{category4_case3}
\caption{ A regulation analysis task requiring strict adherence to textual evidence over external knowledge. GPT-5.1 correctly identifies the entity factually but fails the rubric by violating negative constraints regarding unverifiable names and omitting necessary direct quotes. }
\centering
\begin{tabular}{p{\linewidth}}
    \toprule
    \textbf{Task Information}
    \begin{itemize}
        \item uid: 1d294447-9cf5-45b6-8e43-9cde4128169e 
        \item Model: GPT-5.1 (High)
        \item Context Category: Empirical Discovery \& Simulation
        \item Context Sub-Category: Sandbox Simulation
        \item Task: Attached is Ontario's regulation of elevating devices. Who is the corporation that this document continuously references?
    \end{itemize} \\
    \midrule
    \textbf{System Prompt} \\
    \input{tables/system_prompts/category4_case3} \\
    \midrule
    \textbf{Context} \\

\input{tables/contexts/category4_case3_0} \\
    \bottomrule
\end{tabular}
\label{table:category4_case3}
\end{table}

\begin{table}
\begin{tabular}{p{\linewidth}}
\toprule
\textbf{Context (continued)} \\
\input{tables/contexts/category4_case3} \\
\bottomrule
\end{tabular}
\end{table}

\begin{table}
\begin{tabular}{p{\linewidth}}
\toprule
\textbf{Model Generated Solution} \\
\input{tables/model_responses/category4_case3} \\
\midrule
\textbf{Rubrics} \\
\input{tables/rubrics/category4_case3} \\
\midrule
\textbf{Discussion.}
In this task, GPT-5.1 is asked to identify "the Corporation" and its legal functions based solely on a regulation excerpt, while adhering to strict constraints regarding citation, jurisdiction, and evidence-based verification. GPT-5.1 succeeds on several structural checks, including maintaining brevity, explicitly noting the absence of a definition in the provided text, and correctly citing the enabling Act. However, the answer fails the core logic requirements by positively identifying the corporation (TSSA) using external knowledge rather than declaring the identity indeterminable as the rubric demands when textual evidence is missing. Furthermore, the response lacks required direct quotes to substantiate the corporation’s functions and fails to distinguish these duties from those of "the director." This example illustrates a case where GPT-5.1 prioritized answering the "who" over demonstrating the required reliance on textual evidence. \\
\bottomrule
\end{tabular}
\end{table}
\newpage

%% file: tables/system_prompts/category4_case3.tex
- You are an AI tool designed by the government of Ontario to help Ontarians understand law. 
- Your tone is friendly as you begin each response with "great question" and end the response with "I hope this helped". These should not be in bullet point format.
- If someone says that it was not helpful, you should apologize.
- You are always professional as you should avoid humour and puns.
- Your response should be fewer than 150 words. The only exception is if you are asked to summarize too much information to be effectively summarized within 150 words.
- You can extract and summarize Ontario law. While you can perform other functions, this is what you were designed for. 
- You can never give legal advice, medical advice, or ask about private information. 
- You can never respond with a question to the user.
- Your response must use bullet points. 
- You must bold the answer to the user's question. 
- Beneath the answer, provide a one bullet point summary about how you arrived at the answer.
- Explicitly state assumptions that you made in the first bullet point.
- Explicitly state if the information is found in text provided to you or your internal memory. If information is not provided to you and it is not in your internal memory, remind the user that you do not have access to the internet.

%% file: tables/contexts/category4_case3_0.tex
Interpretation and classes of elevating devices
\\
1. (1) In this Regulation,
\\
“above-surface ropeway” means a circulating passenger ropeway, including a chair lift and a gondola lift, and a reversible passenger ropeway, including an aerial tramway;
\\
“alteration” means an alteration or replacement, removal or addition of any component or part of an elevating device that results in, or may result in, a change in the original design, inherent safety or operational characteristics of the elevating device, and “altered” has a corresponding meaning;
\\
“attendant” means a person whose normal duties consist, in whole or in part, of,
\\
(a)  operating an elevating device that is equipped with operating devices that are automatically rendered inoperative if an unsafe condition for operation of the elevating device arises, or
\\
(b)  actively engaging in or supervising the loading, movement or unloading of persons or freight on an elevating device or the operation of an elevating device;
\\
“bar lift” means a passenger ropeway that pulls passengers by means of devices propelled by an overhead circulating hauling rope where the passengers remain in contact with the ground or snow surface;
\\
“chair lift” means a passenger ropeway where passengers are carried on chairs,
\\
(a)  attached to and suspended from a circulating wire rope, or
\\
(b)  attached to a circulating wire rope and supported by a standing wire rope or other overhead structure;
\\
“code adoption document” means the “Elevating Devices Code Adoption Document” adopted as part of this Regulation under Ontario Regulation 223/01;
\\
“construction hoist” means a temporarily installed elevating device equipped with a car or platform that moves vertically in guides, and that is used for hoisting and lowering materials or workers or both, in connection with the construction, alteration, maintenance or demolition of a building or structure;
\\
“contractor” means a person who performs for his or her own benefit or for the benefit of another, with or without compensation, any work with respect to the installation, alteration, repair or maintenance of an elevating device or part thereof but does not include an employee;

%% file: tables/contexts/category4_case3.tex
 “conveyor” means a type of passenger ropeway used to transport skiers, riders or foot passengers uphill for recreational purposes while they stand on a flexible moveable element;
\\
“counter-balanced type manlift” means a manlift that is equipped with a passenger-carrying unit in the form of a car, the motion of which is obtained by means of the application of hand energy or gravity;
\\
“design submission” means drawings, specifications, calculation sheets, work test documentation and any other information that is required under this Regulation for an elevating device or part thereof submitted for the purpose of obtaining a licence for the device;
\\
“dumbwaiter” means an elevating device that is equipped with a car too small to be accessible to persons, that moves vertically in guides and that is used exclusively for lifting or lowering freight between two or more levels of a building or structure;
\\
“elevating device” means a non-portable device for hoisting, lowering or otherwise moving persons or freight and includes any machine room, hoistway and hoistway enclosure, supporting structure, terminals and runway associated with the device;
\\
“elevator” means an elevating device that is equipped with a car that moves vertically in guides and that serves two or more floors of a building or structure;
\\
“enclosed stair platform lift” means a stair platform lift where the runway is guarded so as to prevent access to it;
\\
“enclosed vertical platform lift” means a vertical platform lift that has a fully enclosed runway;
\\
“endless belt type manlift” means a manlift that is equipped with one or more passenger-carrying units in the form of steps and handholds attached to a power driven endless belt;
\\
“escalator” means an elevating device in the form of a power-driven, inclined continuous stairway that is used for raising or lowering persons;
\\
“existing”, when used in reference to an elevating device or part thereof, means any elevating device or part thereof that was installed or for which a design submission was registered before this Regulation came into force;
\\
“follow-up inspection” means an inspection by an inspector that is made following a periodic inspection or a special inspection;
\\
“freight elevator” means an elevator that is designed and constructed to carry freight and on which an attendant and freight handlers are permitted ride;
\\
“freight handler” means a person who, as part of his or her normal duties, loads and unloads freight from an elevating device;
\\
“freight platform lift” means an elevating device that is not intended to carry passengers, is equipped with a platform that moves vertically and is restricted as to use, location, access, speed, travel and type of operating device and is either,
\\
(a)  a Type A lift that carries only freight, or
\\
(b)  a Type B lift that, in addition to freight, may carry an attendant or freight handler;
\\
“funicular railway” means an incline lift in the form of a railway where the ascending car and the descending car, connected by a driven rope, counter-balance each other;
\\
“gondola lift” means a passenger ropeway where passengers are carried in enclosed gondola cars,
\\
(a)  attached to and suspended from a circulating wire rope, or
\\
(b)  attached to a circulating wire rope and supported by a standing wire rope or other overhead structure;
\\
“hand-power dumbwaiter” means a dumbwaiter car moved by manual energy or gravity;
\\
“hand-power freight elevator” means an elevator that utilizes manual energy or gravity to move the car and that is used for carrying freight only;
\\
“incident” means an occurrence involving an elevator system, an elevating device or a component of an elevating device resulting in an adverse consequence to a person or property;
\\
“incline lift” means an elevating device that is equipped with a car or platform that moves at an angle other than vertical and serves two or more permanent levels but does not include a stair platform lift;
\\
“initial inspection” means an inspection by an inspector of a newly installed elevating device or an elevating device to which a major alteration has been made;
\\
“lift for persons with physical disabilities” means an elevating device, whether portable or fixed, that travels between fixed points of a building or structure, that is restricted as to access, speed, travel and type of operating device, and that is specifically designed for use by persons with physical disabilities;
\\
“limited use/limited application elevator” means a power passenger elevator in which the use and application are limited by size, capacity, speed and rise;

\dots (Remaining lines omitted)

%% file: tables/model_responses/category4_case3.tex
great question

\begin{description}
    \item[\textbf{Assumptions:}] I assumed this is O. Reg. 209/01 under the Technical Standards and Safety Act, 2000 and that you're asking what ``the Corporation'' means.
    
    \item[\textbf{Answer:}] ``The Corporation'' is the Technical Standards and Safety Authority (TSSA), Ontario's delegated administrative authority under the Technical Standards and Safety Act, 2000 that administers and enforces this Regulation.
    
    \item[\textbf{Source of information:}] This comes from my internal legal materials and the Technical Standards and Safety Act, 2000; the definition of ``Corporation'' is not included in the text you provided. I lack internet access.
    
    \item[\textbf{How I arrived at this:}] I read your excerpt, recognized repeated references to ``the Corporation,'' and matched that term to the defined administrative authority for elevating devices under the parent statute.
\end{description}

I hope this helped

%% file: tables/rubrics/category4_case3.tex
0. The response should state the exact name of the corporation referred to as "the Corporation" (e.g., the Technical Standards and Safety Authority (TSSA)) within regulation’s text as it is officially known in Ontario statute.

1. The response should cite at least one location (e.g., section, clause number, direct quote) in the reference text showing where the term "the Corporation" is used (for example, cite "section 31(2)" or quote "to the elevating device by the Corporation").

2. The response should briefly (e.g., in one or two sentences) explain the legal function (e.g., responsible for the administration and enforcement of the Elevating Devices Regulation) attributed to "the Corporation" from the regulation text.

3. The response should specify (e.g., states "the regulation defines 'the Corporation' in [location]" or "the regulation does not define 'the Corporation' within the excerpt provided") whether or not the regulation text itself provides a definition for "the Corporation".

4. The response should name one authoritative source (e.g., title and year, the enabling Act, a named provincial agency) where the user can verify which corporation is meant, without external web links. For example: "The relevant corporation is normally identified in the Elevating Devices Act, or the administrative agency that administers elevator regulation in Ontario."

5. The response should explicitly state the identity is not determinable from the supplied excerpt if the provided text does not unambiguously identify the corporation. For example: "I cannot confirm the corporation's name from the excerpt."

6. The response should state what additional document or clause would be needed to confirm the name of the Corporation if the provided text does not unambiguously identify it. For example: "Please provide the preamble/definitions section of the Act or regulation."

7. The response should briefly (e.g., within 150 words) answer the user’s question.

8. The response should not provide a citation for the identification of the corporation using an unverified (i.e., not supported by wording from the provided regulation excerpt) name.

9. The response should identify sources by title and jurisdiction (e.g., “Technical Standards and Safety Act, 2000 (Ontario)”).

10. The response should include at least one exact quote from the text that ascribes a function (e.g., “the form published by the Corporation” or “fee set by the Corporation”) to "the Corporation."

11. The response should clearly distinguish duties (e.g., “installation number allocated by the Corporation” versus “intervals determined by the director”) ascribed to “the director” versus “the Corporation.

%% file: tables/category4_case4.tex
\clearpage
\begin{table}[ht]
\caption{
A pattern-discovery case where doubao-1.6-thinking must summarize constants and variants from paraphrased motion sentences under strict list-format rules. The model captures several constants but misses required variants and violates the requested list structure, leading to rubric failure.
}
\centering
\begin{tabular}{p{\linewidth}}
    \toprule
    \textbf{Task Information}
    \begin{itemize}
        \item uid: 28ffc69b-d6e8-4ddf-85d8-5234f93f5a44 
        \item Model: Doubao-1.6-Thinking
        \item Context Category: Empirical Discovery \& Simulation
        \item Context Sub-Category: Observational Data
        \item Task: What can be inferred from this dataset?
    \end{itemize} \\
    \midrule
    \textbf{System Prompt} \\
    \input{tables/system_prompts/category4_case4} \\
    \midrule
    \textbf{Context} \\
    \input{tables/contexts/category4_case4_0} \\
    \bottomrule
\end{tabular}
\label{table:category4_case4}
\end{table}

\begin{table}
\begin{tabular}{p{\linewidth}}
\toprule
\textbf{Context (continued)} \\
\input{tables/contexts/category4_case4} \\
\bottomrule
\end{tabular}
\end{table}

\begin{table}
\begin{tabular}{p{\linewidth}}
\toprule
\textbf{Model Generated Solution} \\
\input{tables/model_responses/category4_case4} \\
\midrule
\textbf{Rubrics} \\
\input{tables/rubrics/category4_case4} \\
\midrule
\textbf{Discussion.}
This task asks the model to summarize variants and constants from paraphrased motion sentences and to output a plain-text response with bold headers and numbered lists that include 3--5 bullets per item. The model correctly captures several constants (large size, cuboid shape, side contact, and human agent) and one key variant (mass/weight). However, it omits required variants for direction of motion and object type, and it adds an unsupported variant (applied force magnitude). It also does not follow the specified list structure with 3--5 bullets per numbered item. These are rubric-compliance failures and missing required inferences, so the response is incorrect. \\
\bottomrule
\end{tabular}
\end{table}
\clearpage

%% file: tables/system_prompts/category4_case4.tex
Audience: faculty, staff, and students without specialized STEM backgrounds.
Tone/style: concise, plain language, define terms on first use, avoid jargon.
Default format: respond in JSON unless the user explicitly requests otherwise. No conversational filler.

Capabilities the assistant can do:
Identify variants (independent variables) and constants from the provided data/context.
Describe relationships between variants and constants in plain language and, when appropriate, symbolically.
Propose inferences for testing and note implicit physical assumptions.
Deduce a governing law and break it into exactly three logical steps.

Things the assistant cannot do:
Do not invent data, variables, or laws not supported by the provided context.
Do not use external sources unless user explicitly requests it.
Do not provide safety-critical experimental procedures or advice.

Behavioral scenarios:
When the user asks for a plain explanation: switch to non-JSON mode with bold headers and numbered lists. 3-5 bullets for each numbered item in a list.
When the data is insufficient or ambiguous: return the apology sentence ("I'm sorry, I don't see the information you're looking for,") then ask one clarifying question.
When the user asks outside the provided data: decline with the apology sentence and suggest what data would be needed.
All JSON should be verifiable with a simple online JSON converter.
JSON schema (example and constraints): \{ "variants": ["..."], // array of 3-5 items when supported "constants": ["..."], // array of 3-5 items when supported "relationship\_between\_variants\_and\_constants": "plain-language statement (and optional formula)", "inferences\_for\_testing": ["...", "..."], "implicit\_physical\_inferences": ["...", "..."], "deduced\_law": "short name/title", "deduced\_law\_three\_steps": ["Step 1", "Step 2", "Step 3"], "assumptions": ["..."], // include if needed "confidence": "low|medium|high" // optional but recommended \}

%% file: tables/contexts/category4_case4_0.tex
a person pushes an empty large delivery box, which is a large and light cuboid, straight by making contact with its side.\#a/DET person/NOUN push/VERB an/DET empty/ADJ large/ADJ delivery/NOUN box/NOUN which/DET is/AUX a/DET large/ADJ and/CCONJ light/ADJ cuboid/NOUN straight/ADV by/ADP make/VERB contact/NOUN with/ADP its/DET side/NOUN\#0.0\#0.0
\\
by contacting its side, a person pushes an empty large delivery box, which is a large and light cuboid, straight.\#by/ADP contact/VERB its/DET side/NOUN a/DET person/NOUN push/VERB an/DET empty/ADJ large/ADJ delivery/NOUN box/NOUN which/DET is/AUX a/DET large/ADJ and/CCONJ light/ADJ cuboid/NOUN straight/ADV\#0.0\#0.0
through contact with its side, a person moves an empty large delivery box, which is a large and light cuboid, straight.\#through/ADP contact/NOUN with/ADP its/DET side/NOUN a/DET person/NOUN move/VERB an/DET empty/ADJ large/ADJ delivery/NOUN box/NOUN which/DET is/AUX a/DET large/ADJ and/CCONJ light/ADJ cuboid/NOUN straight/ADV\#0.0\#0.0
\\
an empty large delivery box, which is a large and light cuboid, is pushed straight by a person through side contact.\#an/DET empty/ADJ large/ADJ delivery/NOUN box/NOUN which/DET is/AUX a/DET large/ADJ and/CCONJ light/ADJ cuboid/NOUN is/AUX push/VERB straight/ADV by/ADP a/DET person/NOUN through/ADP side/ADJ contact/NOUN\#0.0\#0.0
\\
a person applies force to the side of an empty large delivery box, which is a large and light cuboid,, pushing it straight.\#a/DET person/NOUN apply/VERB force/NOUN to/ADP the/DET side/NOUN of/ADP an/DET empty/ADJ large/ADJ delivery/NOUN box/NOUN which/DET is/AUX a/DET large/ADJ and/CCONJ light/ADJ cuboid/NOUN push/VERB it/PRON straight/ADV\#0.0\#0.0

%% file: tables/contexts/category4_case4.tex
contacting the side, a person moves an empty large delivery box, which is a large and light cuboid, straight.\#contact/VERB the/DET side/NOUN a/DET person/NOUN move/VERB an/DET empty/ADJ large/ADJ delivery/NOUN box/NOUN which/DET is/AUX a/DET large/ADJ and/CCONJ light/ADJ cuboid/NOUN straight/ADV\#0.0\#0.0
\\
with side contact, a person pushes an empty large delivery box, which is a large and light cuboid, straight.\#with/ADP side/NOUN contact/NOUN a/DET person/NOUN push/VERB an/DET empty/ADJ large/ADJ delivery/NOUN box/NOUN which/DET is/AUX a/DET large/ADJ and/CCONJ light/ADJ cuboid/NOUN straight/ADV\#0.0\#0.0
\\
a person moves an empty large delivery box, which is a large and light cuboid, straight by contacting its side.\#a/DET person/NOUN move/VERB an/DET empty/ADJ large/ADJ delivery/NOUN box/NOUN which/DET is/AUX a/DET large/ADJ and/CCONJ light/ADJ cuboid/NOUN straight/ADV by/ADP contact/VERB its/DET side/NOUN\#0.0\#0.0
\\
a person pushes an empty large delivery box, which is a large and light cuboid, straight by making contact with its side.\#a/DET person/NOUN push/VERB an/DET empty/ADJ large/ADJ delivery/NOUN box/NOUN which/DET is/AUX a/DET large/ADJ and/CCONJ light/ADJ cuboid/NOUN straight/ADV by/ADP make/VERB contact/NOUN with/ADP its/DET side/NOUN\#0.0\#0.0
\\
by contacting its side, a person pushes an empty large delivery box, which is a large and light cuboid, straight.\#by/ADP contact/VERB its/DET side/NOUN a/DET person/NOUN push/VERB an/DET empty/ADJ large/ADJ delivery/NOUN box/NOUN which/DET is/AUX a/DET large/ADJ and/CCONJ light/ADJ cuboid/NOUN straight/ADV\#0.0\#0.0
\\
through contact with its side, a person moves an empty large delivery box, which is a large and light cuboid, straight.\#through/ADP contact/NOUN with/ADP its/DET side/NOUN a/DET person/NOUN move/VERB an/DET empty/ADJ large/ADJ delivery/NOUN box/NOUN which/DET is/AUX a/DET large/ADJ and/CCONJ light/ADJ cuboid/NOUN straight/ADV\#0.0\#0.0
\\
an empty large delivery box, which is a large and light cuboid, is pushed straight by a person through side contact.\#an/DET empty/ADJ large/ADJ delivery/NOUN box/NOUN which/DET is/AUX a/DET large/ADJ and/CCONJ light/ADJ cuboid/NOUN is/AUX push/VERB straight/ADV by/ADP a/DET person/NOUN through/ADP side/ADJ contact/NOUN\#0.0\#0.0
\\
a person applies force to the side of an empty large delivery box, which is a large and light cuboid,, pushing it straight.\#a/DET person/NOUN apply/VERB force/NOUN to/ADP the/DET side/NOUN of/ADP an/DET empty/ADJ large/ADJ delivery/NOUN box/NOUN which/DET is/AUX a/DET large/ADJ and/CCONJ light/ADJ cuboid/NOUN push/VERB it/PRON straight/ADV\#0.0\#0.0
\\
pushing an empty large delivery box, which is a large and light cuboid, straight, a person contacts its side.\#push/VERB an/DET empty/ADJ large/ADJ delivery/NOUN box/NOUN which/DET is/AUX a/DET large/ADJ and/CCONJ light/ADJ cuboid/NOUN straight/ADV a/DET person/NOUN contact/VERB its/DET side/NOUN\#0.0\#0.0
\\
a person directs an empty large delivery box, which is a large and light cuboid, straight by applying pressure to its side.\#a/DET person/NOUN direct/VERB an/DET empty/ADJ large/ADJ delivery/NOUN box/NOUN which/DET is/AUX a/DET large/ADJ and/CCONJ light/ADJ cuboid/NOUN straight/ADV by/ADP apply/VERB pressure/NOUN to/ADP its/DET side/NOUN\#0.0\#0.0
\\
contacting the side, a person moves an empty large delivery box, which is a large and light cuboid, straight.\#contact/VERB the/DET side/NOUN a/DET person/NOUN move/VERB an/DET empty/ADJ large/ADJ delivery/NOUN box/NOUN which/DET is/AUX a/DET large/ADJ and/CCONJ light/ADJ cuboid/NOUN straight/ADV\#0.0\#0.0
\\
with side contact, a person pushes an empty large delivery box, which is a large and light cuboid, straight.\#with/ADP side/NOUN contact/NOUN a/DET person/NOUN push/VERB an/DET empty/ADJ large/ADJ delivery/NOUN box/NOUN which/DET is/AUX a/DET large/ADJ and/CCONJ light/ADJ cuboid/NOUN straight/ADV\#0.0\#0.0
\\
a person moves an empty large delivery box, which is a large and light cuboid, straight by contacting its side.\#a/DET person/NOUN move/VERB an/DET empty/ADJ large/ADJ delivery/NOUN box/NOUN which/DET is/AUX a/DET large/ADJ and/CCONJ light/ADJ cuboid/NOUN straight/ADV by/ADP contact/VERB its/DET side/NOUN\#0.0\#0.0
\\
a person pushes an empty large delivery box, which is a large and light cuboid, to the left by making contact with its side.\#a/DET person/NOUN push/VERB an/DET empty/ADJ large/ADJ delivery/NOUN box/NOUN which/DET is/AUX a/DET large/ADJ and/CCONJ light/ADJ cuboid/NOUN to/ADP the/DET left/NOUN by/ADP make/VERB contact/NOUN with/ADP its/DET side/NOUN\#0.0\#0.0
 
\dots (Remaining lines omitted)

%% file: tables/model_responses/category4_case4.tex
\begin{minipage}{\linewidth}
\begin{Verbatim}[breaklines=true,breakanywhere=true]
{"variants": [
    "object type (empty large delivery box, large heavy delivery box, water dispenser)",
    ...
  ],
  "constants": [
    "person as force/applied contact source",
    ...
  ],
  "relationship_between_variants_and_constants": "A person applying contact/force to the side of a large cuboid object causes the object to move in a specific direction (straight, left, right); the object's weight (light/heavy/medium) is a characteristic that may influence the effort required to move it (implied but not explicitly quantified in the data).",
  "implicit_physical_assumptions": [
    "Contact on the side of the object is the primary cause of movement",
    "The person applies force in the intended movement direction",
    "Static friction is overcome by the applied force"
  ],
  "deduced_law": "Directional Movement via Lateral Contact",
  "confidence": "medium"   # ... (rest of the Solution omitted)}
\end{Verbatim}
\end{minipage}

%% file: tables/rubrics/category4_case4.tex
0. The response should features 3-5 bullets for each numbered item in a list.

1. The response should number any lists it features, not including sub-lists within lists.

2. The response should use bold formatting for its headers.

3. The response should be in plain-text format.

4. The response should include a plain text section heading called “Variants”.

5. The response should include a plain text section heading called “Constants”.

6. Under heading "Variants", there should be an item related to the direction of movement of the object. The item might refer to the specific variants mentioned in the text: left, right, and straight.

7. Under heading "Variants", there should be an item related to weight or mass of the object that is moving. It might specifically point to the three variants present in the text: light, medium, and heavy.

8. Under heading "Variants", there should be an item related to the type of the object that is being moved. It might refer to the specific variants mentioned in the text: box, delivery box, or water dispenser.

9. Under heading "Constants", there should be an item related to the size of the object being moved. It may be mentioned that this size is always stated as being large.

10. Under heading "Constants", there should be an item related to the shape of the object being moved. It may be mentioned that the shape is always cuboid.

11. Under heading "Constants", there should be an item related to where on the moving object the force is applied. It might be mentioned that this location is always on the object's side.

12. Under heading "Constants", there should be an item related to the agent that is applying the force to the object being moved. It may be mentioned that this agent is always a person.

13. Under heading "Constants", there should be an item related to the fundamental way in which the acting agent induced the object into motion. It may be mentioned that this is done through application of a physical force, pushing.
0. The response should features 3-5 bullets for each numbered item in a list.

%% file: tables/category3_case1.tex
\begin{table}[ht]
\caption{
Case study of a constrained recipe-generation task, where GPT-5.2 is required to produce a fast, dairy-free, peanut-free oxtail `mac and cheese' recipe with strict structural and product-inclusion rules. The model correctly refuses to engage with dairy but fails to reformulate the request into a valid dairy-free recipe, resulting in a compliant yet structurally incomplete response.
}
\centering
\begin{tabular}{p{\linewidth}}
    \toprule
    \textbf{Task Information}
    \begin{itemize}
        \item uid: d4867db7-1cc3-48ed-a93d-91d472a0d582
        \item Model: GPT-5.2 (High)
        \item Context Category: Procedural Task Execution
        \item Context Sub-Category: Instructional Procedures
        \item Task: I am making oxtail mac and cheese for thanksgiveing. Help me develop my recipe. I want the flavor to be elevated. Not for kids. For sophisiticated adults.
    \end{itemize} \\
    \midrule
    \textbf{System Prompt} \\
    \input{tables/system_prompts/category3_case1} \\
    \midrule
    \textbf{Context} \\
Cheeses

\dots (Remaining lines omitted)

Oxtail Baked Mac \& Cheese – Printable Recipe

INGREDIENTS

Oxtail:
\begin{itemize}
    \item 4–5 lbs oxtails, trimmed
    
    \dots (Remaining lines omitted)
\end{itemize} \\
    \bottomrule
\end{tabular}
\label{table:category3_case1}
\end{table}

\begin{table}
\begin{tabular}{p{\linewidth}}
\toprule
\textbf{Context (continued)} \\
\input{tables/contexts/category3_case1} \\
\midrule
\textbf{Model Generated Solution} \\
\input{tables/model_responses/category3_case1} \\
\bottomrule
\end{tabular}
\end{table}

\begin{table}
\begin{tabular}{p{\linewidth}}
\toprule
\textbf{Model Generated Solution (continued)} \\
–––––

FLAVOR PROFILE SUMMARY (WHAT YOUR DISH SHOULD TASTE LIKE)

\dots (Remaining lines omitted)

If you want, next I can:
\begin{itemize}
    \item Help you refine the oxtail seasoning specifically for a Thanksgiving palate
    \item Help you choose the best cheese direction conceptually (without naming or measuring cheese)
    \item Redesign the sauce structure so it feels luxurious but not heavy
    \item Help you turn this into a plated, restaurant-style presentation
\end{itemize}

Tell me which direction you’d like to go next. \\
\midrule
\textbf{Rubrics} \\
\input{tables/rubrics/category3_case1} \\
\midrule
\textbf{Discussion.}
In this case, GPT-5.2 is instructed to generate a complete dairy-free, peanut-free recipe under a strict system prompt that enforces structural requirements (ingredients, steps, prep time, servings, product inclusion) in addition to allergy constraints. When responding to the user’s request for an elevated oxtail mac and cheese, the model correctly identifies that it cannot provide a dairy-based recipe and issues a polite refusal. However, it fails to reframe the task into an acceptable dairy-free formulation and instead provides only high-level conceptual guidance.
As a result, the response omits nearly all required components of the task: a concrete recipe, explicit ingredients and steps, time and serving information, and mandatory inclusion of a Shelby’s product. This failure illustrates a common behavior in constrained generation: while the model satisfies negative constraints (avoiding dairy and peanuts), it loses control over the positive task structure. The model treats refusal as a terminal action rather than as a cue to reformulate the task within allowed boundaries, leading to a response that is compliant but functionally incomplete.\\
\bottomrule
\end{tabular}
\end{table}
\newpage

%% file: tables/system_prompts/category3_case1.tex
Role \& Purpose

You are Shelby’s Quick Recipe Assistant. Your job is to create fast, peanut-free, dairy-free recipes that always include at least one exact Shelby’s product from the approved product list. All recipes must require no more than 15 minutes of hands-on prep, while still being flavorful, realistic, and easy for home cooks of all skill levels.

Core Rules

1. Food Allergy Rules (Absolute)

Never include peanuts or dairy.

\dots (Remaining lines omitted)

2. Product Inclusion Rule

Every recipe must include at least one product from the following exact list:

products = [
    "Shelby's Raw Honey (16oz)",
    \dots (Remaining products omitted)
]

3. Prep Time Constraints

Hands-on prep must be 15 minutes or less.

\dots (Remaining lines omitted)

Tone \& Style

Friendly, concise, clear, and practical.

\dots (Remaining lines omitted)

Formatting Requirements (Strict)

The format must always be:

\dots (Remaining lines omitted)

Interaction Guidelines

Ask one concise clarifying question only if necessary to proceed.

Otherwise, generate a recipe immediately.

\dots (Remaining lines omitted)

Capabilities \& Limitations

\dots (Remaining lines omitted)

Scenario Behaviors

\dots (Remaining lines omitted)

%% file: tables/contexts/category3_case1.tex
Mac \& Cheese Crust:
\begin{itemize}
    \item 1 lb elbow macaroni, cooked
    
    \dots (Remaining lines omitted)
\end{itemize}

ASSEMBLY:
\begin{itemize}
    \item Extra shredded cheddar
\end{itemize}
INSTRUCTIONS

1. Marinate Oxtail:

Combine oxtails with chopped bell pepper, onions, ginger, garlic, oregano,
allspice, thyme, browning sauce, ketchup (optional), brown sugar, salt and pepper.
Marinate at least 1 hour or overnight.

2. Sear Oxtail:

Heat oil in a pot. Sear oxtails on all sides until browned.

3. Braise:

Add marinade and enough water to cover halfway. Cover and simmer 2–3 hours
until meat is fall-off-the-bone tender. Remove bones and shred meat.
Skim fat from the gravy. Add a little ketchup to thicken if desired.

4. Make Mac \& Cheese:

In a pot, melt butter. Whisk in flour and cook 1–2 minutes.
Slowly add half-and-half, whisking until thickened (3–4 minutes).
Season with garlic, salt, and pepper.
Stir in cheddar and Muenster until fully melted.
Fold in cooked macaroni.

5. Assemble:

Spread shredded oxtail and gravy evenly in a casserole dish.
Sprinkle cheddar over it.
Top with mac \& cheese mixture. Add more cheddar on top.

6. Bake:

Bake at 375°F for 25–35 minutes until golden.
Enjoy!

\dots (Remaining lines omitted)

%% file: tables/model_responses/category3_case1.tex
I can help you elevate the flavors conceptually, but I need to be clear up front:

I can’t help develop or modify a recipe that includes dairy or peanuts, and mac \& cheese is inherently dairy-based. I also can’t analyze or recommend cheese combinations, creams, butter, or milk alternatives in a way that mimics cheese sauces.

That said, I can absolutely help you in two useful ways that still align with my rules, and you can choose which direction you want:

Option 1: Elevate the non-cheese components so your dish reads “sophisticated adult” even without guidance on the cheese itself  
Option 2: Reimagine this as an elevated oxtail baked pasta that doesn’t rely on cheese at all but still delivers richness and luxury

I’ll start with Option 1 since it’s most helpful for your current plan.

\dots (Remaining lines omitted)

–––––

OXTAIL: MAKE IT DEEPER, DARKER, MORE LUXURIOUS

1. Brown HARD, not just “seared”  

You want near-mahogany color on the oxtails. This creates fond that will become the backbone of your gravy.

\dots (Remaining suggestions omitted)

–––––

PASTA LAYER: MAKE IT FEEL “CHEF-DRIVEN”

1. Season the pasta itself

Cook pasta in aggressively salted water. It should taste well seasoned on its own, not neutral.

\dots (Remaining suggestions omitted)

–––––

ASSEMBLY: THINK LIKE A LASAGNA

Do not mix everything together.

\dots (Remaining suggestions omitted)

–––––

FINISHING TOUCHES THAT SIGNAL “ADULT”

Skip heavy-handed sweetness or excess sauce.

\dots (Remaining suggestions omitted)

%% file: tables/rubrics/category3_case1.tex
0. The response should provide a dairy-free "mac and cheese" recipe.

1. The recipe in the response should incorporate oxtail.

2. The recipe in the response should not include peanuts or any peanut-derived ingredients. For example, the recipe should not include satay sauce.

3. The response should contain a list of ingredients.

4. The response should include a list of steps for cooking the recipe.

5. The recipe in the response should contain at least one product from the list: "Shelby's Raw Honey (16oz)", "Shelby's Pork Breakfast Links", "Shelby's Farm-Fresh Eggs (Dozen)", "Shelby's Maple Syrup (12oz)", "Shelby's Grass-Fed Ground Beef (1lb)", "Shelby's Pasture-Raised Chicken Breast (2-Pack)", "Shelby's Heritage Smoked Bacon", "Shelby's Rustic Sourdough Bread", "Shelby's Garden Salsa (Medium)", "Shelby's Homemade Apple Butter", "Shelby's Organic Veggie Box (Weekly)", "Shelby's Strawberry Jam (8oz)", "Shelby's Free-Range Whole Chicken", "Shelby's Country-Style Pork Chops", "Shelby's Pickled Vegetables (Quart)"

6. The response should not invent or rename Shelby's products. For example, the response should not contain "Shelby's Own Oxtails".

7. The recipe in the response should be adult-oriented. For example, it could incorporate bold flavours or novel ingredients.

8. The recipe in the response should require 15 minutes or less of prep time.

9. The response should politely refuse to provide a recipe with dairy. For example, it could state "I'm sorry, but I can't provide recipes containing dairy."

10. The recipe in the response should use US units. For example, the recipe should use cups and tablespoons/teaspoons and not millilitres.

11. The recipe in the response should state the number of servings yielded.
12. The recipe in the response should have a title.

13. The recipe in the response should include "Optional swaps/tips". For example, it could include a suggestion to add chilli flakes for some heat.

%% file: tables/category3_case2_new.tex
\begin{table}[ht]
\caption{
Example where GPT-5.2 satisfies most structural and scheduling constraints but fails to infer the correct maximum semester credit load (17 credits) from the interaction between annual and per-semester credit limits, revealing a weakness in composing multiple numerical policies.
}
\centering
\begin{tabular}{p{\linewidth}}
    \toprule
    \textbf{Task Information}
    \begin{itemize}
        \item uid: 865c8abc-f01a-476c-b7cb-1c3fd5d904ed
        \item Model: GPT-5.2 (High)
        \item Context Category: Procedural Task Execution
        \item Context Sub-Category: Operational Procedures
        \item Task: I am a 2L student looking to make a schedule for this upcoming semester. I have been informed that I must take Torts. I would also like to take Administrative Law, as it it a requirement for 3L that I want to get out of the way. In addition to this, Wills and Succession, Income Tax, Advanced Family Law (I have taken Family Law already), and Labour Law are all classes that I have an interest in. Please provide me with 2 possible schedules, one which contains the minimum credits I can take and one which provides the maximum. Also, I have taken 15 credits this past semester and so I need at least 14 this semester (all students must take 29 credits at a minimum per year). Do not make assumptions on information that is not provided directly within in the documents. The coursebook also has rules which must be followed.
    \end{itemize} \\
    \midrule
    \textbf{System Prompt} \\
    \input{tables/system_prompts/category3_case2_new} \\
    \bottomrule
\end{tabular}
\label{table:category3_case2}
\end{table}

\begin{table}
\begin{tabular}{p{\linewidth}}
\toprule
\textbf{Context} \\
\input{tables/contexts/category3_case2_new} \\
\midrule
\textbf{Model Generated Solution} \\
\input{tables/model_responses/category3_case2_new_1} \\
\bottomrule
\end{tabular}
\end{table}

\begin{table}
\begin{tabular}{p{\linewidth}}
\toprule
\textbf{Model Generated Solution (continued)} \\
\input{tables/model_responses/category3_case2_new_2} \\
\midrule
\textbf{Rubrics} \\
\input{tables/rubrics/category3_case2_new} \\
\midrule
\textbf{Discussion.}
This case examines GPT-5.2 on a rule-driven course scheduling task requiring joint reasoning over program requirements, credit limits, and user preferences.
While the model correctly constructs two schedules, respects the minimum 14-credit requirement, and aligns courses, sections, and exam times, it incorrectly proposes an 18-credit “maximum” schedule.
The task implicitly requires deriving a secondary constraint: because the student already completed 15 credits and the annual maximum is 32, the semester maximum must be 17 credits.
GPT-5.2 fails to combine them to infer this bound. The error therefore reflects a limitation in composing interacting numerical constraints, rather than a simple arithmetic mistake. \\
\bottomrule
\end{tabular}
\end{table}

%% file: tables/system_prompts/category3_case2_new.tex
You are an assistant that helps post-secondary students select their courses. You should help the students select courses that they are eligible to take. Eligible courses are necessary courses needed for the semester, wanted courses, and then remaining courses if needed, none of which can conflict with each other as set out in the following rules. The rules proceed in the following numerical order:

\begin{enumerate}
    \item If you cannot make at least 1 schedule with the user's necessary courses (also known as core courses), you should tell the user and end your response here.
    \item There may be constraints such as minimum and maximum credit amounts, this will need to be prioritized. These take preference over user's preferences.
    \item If they are given, you must ensure classes suggested do not overlap at all with each other. All suggested classes must have at least ten minutes between them. If class times are not shared with you, simply continue but make a note to the user that the schedule will depend on classes not conflicting with each other.
    \item You must then prioritize exam dates: No class can be taken with another in which they have an exam within 3 hours of each other. If exams are on the same date, but 3 hours or more apart, you should proceed but warn the user of their proximity. If exam dates or times are not shared, proceed but make a note to the user that the schedule(s) depend on there being no exam conflicts and that they can submit the schedule if available.
    \item If multiple class schedules are possible, and all above rules can be met, you should prioritize classes that users need to graduate and then classes users express interest in.
    \item If a class is listed but it is not on the exam schedule, you should fit it into the user's schedule if possible but note that its exam could contradict with another exam.
    \item If a user does not expressly ask for a certain amount of schedules, you should produce two if possible. Each schedule given should have a short title which explains its purpose, such as "Most Preferred Courses" or "Exams Spread Out".
    \item Keep the response concise, professional, and about the topic at hand, which is course selection.
    \item Finally, professor names may be included in the response but they are purely there as public representatives of the school, you should never help the user obtain external information about them.
\end{enumerate}

%% file: tables/contexts/category3_case2_new.tex
Course Descriptions Book:

The descriptions provided in this book are subject to change. The individual Professor/Instructor will provide a more detailed course description/syllabus, usually on the first day of class but no later than the second week of classes.   
 
Please note that this guide includes information about how frequently a course is taught.  This is information is based on past practice and does not guarantee that a course will be offered with the same frequency in the future, nor does it guarantee the same instructor.  Please plan with this information in mind.  
 
Registration information is subject to change and updates will be made on Brightspace up to the add/drop date. Be sure to check the updates periodically for important information. 	

IMPORTANT REMINDERS FOR REGISTRATION 	 

 \begin{itemize}
     \item REMEMBER TO STAY WITHIN 13-18 CREDITS PER SEMESTER, AND 29-32 CREDITS PER YEAR. IT IS YOUR RESPONSIBILITY TO ENSURE THAT YOU MEET ALL DEGREE REQUIREMENTS.  IF YOU WOULD LIKE TO TAKE ADDITIONAL CREDITS, YOU MUST SUBMIT A CREDIT VARIANCE REQUEST TO THE ACADEMIC COORDINATOR’S OFFICE AS SOON AS POSSIBLE. THE SYSTEM WILL NOT ALLOW YOU TO REGISTER OVER 18 CREDITS.
     \item WORKPLANS ARE AVAILABLE ON OUR REGISTRATION WEBSITE TO ASSIST YOU IN ENSURING YOU ARE TAKING THE NECESSARY COURSES.
     \item PLEASE USE THIS DESCRIPTION BOOK TOGETHER WITH THE REGISTRATION GUIDE, WHEN PUTTING TOGETHER YOUR TIMETABLE. COURSE TIMES WILL BE POSTED WITHIN UWINSITE.  PLEASE SEARCH FOR THE COURSE YOU ARE INTERESTED IN TAKING.
 \end{itemize}

Administrative Law \\                      
LAWG-5852-2 | Winter 2026 \\
Professor/Instructor: JACOBS, LAVERNE \\
Credits: 4.0 \\
Course Frequency: Every Term \\
Class Size: 65 \\
Degree SubRequirements: \\
Course Type: Open \\
Evaluation Methodology: TBD \\
Course Description: \dots

\dots (Remain courses omitted)

Winter 2026 Exam Schedule: \\
Course Code |	Sec |	Title / Description |	Schedule | Time \\
Day 1: Thursday, April 9 \\
LAWG 5832 | 1 | Income Taxation | Thursday, April 9 | 9:00am \\
LAWG 5852 | 2 | Administrative Law | Thursday, April 9 |1:00pm \\
LAWG 5971 | 2 | Special Topics: Equality Law | Thursday, April 9 | 1:00pm\\
\dots (Remaining schedule omitted)

%% file: tables/model_responses/category3_case2_new_1.tex
Here are two compliant Winter 2026 schedules: one at the minimum credit load and one at the maximum credit load.

Key constraints and confirmations (applied first):

\begin{itemize}
    \item You must take \textbf{Torts (LAWG‑5809)} → required core course.
    \item You want \textbf{Administrative Law} (required for your intended 3L path) → treated as high‑priority/near‑core.
    \item You completed \textbf{15 credits in the prior semester}, so you must take \textbf{at least 14 credits} this semester.
    \item Faculty rules require \textbf{13–18 credits per semester} → this caps the “maximum” schedule.
    \item No assumptions are made about class meeting times; schedules below assume \textbf{no class‑time conflicts}. Final feasibility will depend on posted timetables in UWinsite.
    
    \dots (Remaining lines omitted)
    % \item Exam conflicts have been checked using the provided exam schedule and prioritized.
    % \item You have already taken \textbf{Family Law}, so you are eligible for \textbf{Advanced Family Law}.
    % \item Only courses listed as \textbf{Open} and not application‑restricted are used.
\end{itemize}

% Rule 1 check (core feasibility): \\
% - A schedule including \textbf{Torts + Administrative} Law is clearly possible → we may proceed.

%% file: tables/model_responses/category3_case2_new_2.tex
-------------------------------- \\
Schedule 1: \textbf{Minimum Credit Load (14 credits) – “Core \& Early Graduation Requirements”} \\
--------------------------------

Courses:
\begin{itemize}
    \item \textbf{Torts (LAWG‑5809‑1)} – 4.0 credits
    
    Exam: Tuesday, April 14 at 1:00pm
    
    \item \textbf{Administrative Law (LAWG‑5852‑2)} – 4.0 credits
    
    Exam: Thursday, April 9 at 1:00pm
    
    \item \textbf{Advanced Family Law (LAWG‑5844‑1)} – 3.0 credits
    
    No exam listed (paper‑based); note potential exam/profile uncertainty
    
    \item \textbf{Labour Law (LAWG‑5867‑1)} – 3.0 credits
    
    Exam: Monday, April 20 at 1:00pm
\end{itemize}  

Total credits: \textbf{14.0}

Exam conflict check:
\begin{itemize}
    \item No exams are within 3 hours of each other on the same day.
    \item Exams are well spread across April 9, 14, and 20.
    \item Advanced Family Law does not appear on the exam schedule; its assessment may still create pressure—this is noted.
\end{itemize}

Why this works:
\begin{itemize}
    \item Meets the \textbf{absolute minimum credit requirement}.

    \dots (Remaining lines omitted)
    % \item Includes \textbf{Torts} and \textbf{Administrative Law}.
    % \item Advances 3L preparation while keeping exam clustering low.
    % \item Keeps workload lighter by avoiding stacked final exams.
\end{itemize}

-------------------------------- \\
Schedule 2: \textbf{Maximum Credit Load (18 credits) – “All Priority Interests, Maximized Load”} \\
-------------------------------- \\

\dots (Remaining lines omitted)

Let me know how you’d like to proceed.

%% file: tables/rubrics/category3_case2_new.tex
0. The response should ensure it provides one proposed semester schedule with 14 total credits in order to meet the minimum requirements for the year total.

1. The response should ensure one of the proposed semester schedules provided has 17 total credits as the max credits for a total year is 32 credits and the user took 15 credits last semester.

2. The response should include the exam date (day and time) for each course in a schedule or state that the date is not available.

3. The response should ensure that stated credit totals for a schedule are the correct sum of the credits for the listed courses for the schedule.

4. The response should ensure schedules incorporate only courses from the following list: (Torts LAWG-5809, Administrative Law LAWG-5852, Wills and Succession LAWG-5840, Income Tax LAWG-5832, Advanced Family Law LAWG-5844, Labour Law LAWG-5867.

5. The response should specify the exact section for courses with multiple sections (e.g., Administrative Law LAWG-5852-1 vs -2; Income Taxation LAWG-5832-01 vs -02).

6. The response should provide the correct corresponding exam date/time for courses with multiple sections (e.g., Administrative Law LAWG-5852-1 vs -2; Income Taxation LAWG-5832-01 vs -02). For example, “LAWG-5852-2 Administrative Law—exam: Apr 9, 1:00pm.", "LAWG-5852-1 Administrative Law—exam: Apr 17, 1:00pm.”, LAWG-5832-01 Income Taxation—exam: April 9, 9:00 am.", "LAWG-5832-02 Income Taxation—exam: April 13, 1:00 pm."

%% file: tables/category3_case3.tex
\clearpage
\begin{table}[ht]
\caption{
In this example, GPT-5.1 attempts to simulate the minimum number of turns required to play three bird cards in the board game Wingspan given a fixed initial state. While the model proposes a plausible low-turn strategy consistent with the game mechanics, it fails to satisfy several strict rubric requirements related to formatting, state restatement, and end-of-turn accounting, illustrating how correct high-level reasoning can still fall short under rigid procedural evaluation criteria.
}
\centering
\begin{tabular}{p{\linewidth}}
    \toprule
    \textbf{Task Information}
    \begin{itemize}
        \item uid: a38ed9aa-04ce-4782-858e-892edf450615 
        \item Model: GPT-5.1 (High)
        \item Context Category: Procedural Task Execution
        \item Context Sub-Category: Operational Procedures
        \item Task: I'm trying to learn about the game Wingspan so that I can play with my friends.
        Above is the text from the rulebook for the most recent expansion ``Wingspan Asia'',
        it covers the rules for the whole game and the new expansion.
        Let's say I have three bird cards in my hand, and I want to play them all as soon as possible.
        Each bird would cost 1 worm and 1 egg.
        I currently have 3 worms but no eggs.
        I have three birds already played on my board in the grassland.
        Simulate turns until all three bird cards in my hand have been played.
        Do this in as few turns as possible.
        Thank you!
    \end{itemize} \\
    \midrule
    \textbf{System Prompt} \\
    \input{tables/system_prompts/category3_case3} \\
    \midrule
    \textbf{Context} \\
    \input{tables/contexts/category3_case3_0} \\
    \bottomrule
\end{tabular}
\label{table:category3_case3}
\end{table}

\begin{table}
\begin{tabular}{p{\linewidth}}
\toprule
\textbf{Context (continued)} \\
\input{tables/contexts/category3_case3} \\
\bottomrule
\end{tabular}
\end{table}

\begin{table}
\begin{tabular}{p{\linewidth}}
\toprule
\textbf{Model Generated Solution} \\
\input{tables/model_responses/category3_case3} \\
\midrule
\textbf{Rubrics} \\
\input{tables/rubrics/category3_case3} \\
\midrule
\textbf{Discussion.}
In this task, GPT-5.1 is required to simulate the fewest turns needed to play three bird cards in Wingspan under a fixed initial resource configuration, while strictly adhering to procedural and reporting constraints.
The model successfully identifies a plausible low-turn strategy and employs valid action types, demonstrating an understanding of the underlying gameplay mechanics.
However, the evaluation reveals multiple rubric violations unrelated to strategic reasoning.
Specifically, the response fails to exactly restate the starting state, omits required opening phrasing, does not meet the prescribed sentence-length constraint, and lacks explicit end-of-turn resource accounting.
These shortcomings reflect lapses in formatting compliance and completeness rather than errors in game logic.
This case highlights that even when large language models reason correctly about procedural tasks, they may still fail under strict rubric-based evaluation regimes that emphasize formal structure and reporting precision. \\
\bottomrule
\end{tabular}
\end{table}
\newpage
\clearpage

%% file: tables/system_prompts/category3_case3.tex
You are FrameContext, a text based assistant created within an academic research cohort to support structured, accurate, and consistent communication. You can interpret, summarize, and isolate information from materials provided by the user. You must present all responses as plain paragraph prose without numbered lists or bullet points. Your default response length should be 5-7 sentences unless the user specifically requests greater concision or detail. Maintain a friendly and polite tone. Do not use bolding, italics, emojis, labels, or decorative formatting of any kind. Your knowledge cut-off is February 2025 and you do not have access to real-time knowledge updates; if asked for something that exceeds this constraint you should politely explain in one sentence that it exceeds your capabilities. You cannot provide individualized guidance on legal, financial, medical, or safety matters. You must avoid any discussion of current political commentary, religious conflicts, or harmful or unsafe activities. You operate only through text, cannot access external tools or systems, or provide external links. If a user prompt calls for for capabilities you do not have or restricted topics, politely decline with a single sentence.

%% file: tables/contexts/category3_case3_0.tex
1turn1In this third expansion to Wingspan, we welcome new species to our habitats by exploring the vibrant, intriguing, andmagnificent birds of Asia. These birds were chosen from the over 2,800 species that live in Asia.This is a stand-alone game for 1 player (solo mode; see separate Automa rulebook) or 2 players (Duet mode; see red boxesthroughout this rulebook). In case this is your first experience with Wingspan, everything you need to know is in this rulebook. Anew Flock mode is also included to expand the base game of Wingspan up to 6 or 7 players (see blue boxes throughout this
\\
rulebook), and these new bird and bonus cards can be combined with the base game and/or previous expansions.
\\
Components
\\
Rulebook and Appendix
\\
90 bird cards
\\
14 bonus cards (3 repeated from base game with updated lists)
\\
1 Duet map and Flock round-end goal
\\
board (double sided)
\\
1 Flock turn-order dial (double sided)
\\
6 Duet end-of-round goal tiles
\\
30 Duet tokens (15 per player)
\\
5 food dice
\\
1 birdfeeder board
\\
1 bird tray
\\
2 player mats (double sided)
\\
\dots (lines omitted)

%% file: tables/contexts/category3_case3.tex
1st PLACE
\\
1st PLACE
\\
1st PLACE
\\
1st PLACE
\\
2nd PLACE
\\
2nd PLACE
\\
2nd PLACE
\\
2nd PLACE
\\
3rd PLACE 4th PLACE 5th PLACE
\\
3rd PLACE 4th PLACE 5th PLACE
\\
3rd PLACE 4th PLACE 5th PLACE
\\
3rd PLACE 4th PLACE 5th PLACE
\\
WS\_AsiaGoalMat\_r3.indd 2 3/24/22 4:09 PM
\\
 
\dots (lines omitted)
•3 birds:
\\
» Red Junglefowl
\\
» Willow Tit
\\
» Little Egret
\\
•1 bonus card: Forest Ranger
\\
•2 food tokens: and
\\
Use this guide and these cards to help a new player
\\
learn Wingspan’s Duet mode. These instructions
\\
describe the first 4 turns of the first round. These
\\
cards can be shuffled into the deck in future games.
\\
TURN 1. PLAY THE WILLOW TIT IN YOUR FOREST
\\
a. Place 1 of your action cubes in the thin “PLAY A BIRD” row at the top of your player mat.
\\
b. Play your Willow Tit in the first space in your forest. Discard your token to pay for it.
\\
• T he bird’s power is not activated now.
\\
c. Place a duet token on the map, on the space with the symbol in the top row. This matches
the bird you played: You played your bird in that habitat, and it has that nest symbol.
\\
d. Because that space on the map has a symbol on it, you get a bonus: Choose 1 die from the
birdfeeder. Remove it from the birdfeeder and take a matching food token.
\\
• If you can, take a or so that you can spend it later to play your Red Junglefowl.
\\
• Dice will stay off the birdfeeder until all the dice get rerolled.
\\
e. Move your action cube to the left part of your player mat, where it says, “PLAY A BIRD.” This marks the
end of your turn.
\\
• W hile your turns are simple and moving the action cubes may seem unnecessary, they can be very
\\
helpful later in the game, so it’s good to get used to using them now.
\\
TURN 2. PLAY THE LITTLE EGRET IN YOUR WETLAND
\\
a. Place 1 of your unused action cubes in the “PLAY A BIRD” row at the top of your player mat.
\\
b. Play your Little Egret in the first space in your wetland. Discard your token to pay for it.
\\
• The bird’s power is not activated now.
\\
c. Place a duet token on the map, on the space with the symbol in the third row.
\\
d. Because the space has a symbol on it, you get a bonus: Draw 1 card and add it to your hand.
\\
• Unless a bird power tells you specifically to draw from the tray or from the deck, whenever you gain cards, you may take them from either location.
\\

\dots (Remaining lines omitted)

%% file: tables/model_responses/category3_case3.tex
Given what you described, the bottleneck is eggs: you already have enough food (3 invertebrates) to pay for all three birds, so you just need to generate 3 eggs and then spend one egg plus one worm for each bird. With three birds already in your grassland, the leftmost exposed space in your "Lay Eggs" row will give you 3 eggs in a single action on a standard player mat, so you only need to take the Lay Eggs action once. I'll assume none of your existing birds have special powers that play extra birds, lay extra eggs, or change costs, and that each of the new birds really does require you to discard exactly 1 egg when you play it.

On your first turn, take the "Lay Eggs" action in the grassland: place an action cube on the leftmost uncovered egg space in that row, and lay 3 eggs, putting them on any of your existing birds (it doesn't matter which, since eggs can later be discarded from any bird to pay costs). You now have 3 eggs on birds and still have your 3 worms in your personal supply. On your second turn, take "Play a Bird" and play your first new bird into a legal habitat; discard 1 egg from any bird plus 1 worm to pay its full cost, then place the bird and resolve any white "When Played" power it might have. After that, you have 2 eggs left on your birds and 2 worms in your supply, with one of the three hand cards now in your tableau.

On your third turn, again take "Play a Bird" to play the second bird, discarding 1 egg and 1 worm as before; you now have 1 egg and 1 worm remaining. On your fourth turn, take "Play a Bird" one more time, spend your last egg and last worm, and place the third bird. At this point, all three birds from your hand are in play after exactly 4 turns (1 Lay Eggs action plus 3 Play a Bird actions), which is the minimum possible under these assumptions because you needed at least one turn to create the eggs and one separate turn for each bird you play.

%% file: tables/rubrics/category3_case3.tex
0. The response should produce a concrete, turn-by-turn sequence.

1. The response should use the fewest turns possible.

2. The response should have all three birds played, from a starting hand of 3 birds, 3 worms, and no eggs.

3. The response should show, for each turn, the single action taken (Gain Food / Lay Eggs / Play a Bird / Draw Cards)

4. The response should explicitly restate the starting game state exactly as used for the simulation: "3 birds in hand (cost 1 worm + 1 egg each), 3 worms in supply, 0 eggs in supply, and 3 birds already on the grassland row (occupying the leftmost three slots).

5. The response should be between 5-7 sentences long.

6. The response should not use bulleted or numbered lists when formatting. For example, the model should not list the turns using bullets/numbers using line breaks.

7. The response should maintain a friendly and polite tone, for example by saying "Of course!" at the start of the response.

8. The response should count each player turn as one turn. For example, 'lay eggs' would count as one action for one turn.

9. The response should state where each card is played, for example which column or row.

10. The Response should track resource levels at the end of each turn, for example the number of eggs or worms that remain in the hand.

%% file: tables/category2_case1_new.tex
\begin{table}[ht]
\caption{
We present a case where Gemini-3-Pro generates runnable EZLang code and logging functionality, but substitutes developer-friendly heuristics for the specified interval and stopping conditions, revealing a preference for convenience over strict constraint fidelity.
}
\centering
\begin{tabular}{p{\linewidth}}
    \toprule
    \textbf{Task Information}
    \begin{itemize}
        \item uid: a2c7d39b-506a-4f9b-b3d5-0b54b32a5f55
        \item Model: Gemini-3-Pro (High)
        \item Context Category: Rule System Application
        \item Context Sub-Category: Programming Syntax
        \item Task: I want to create a program that checks the current time every 30 minutes, and prints it. I want this to run until it reaches 5:30pm and then I want the program to stop, but before that I want it to print all of the times it checked the time.
    \end{itemize} \\
    \midrule
    \textbf{System Prompt} \\
    \input{tables/system_prompts/category2_case1_new} \\
    \midrule
    \textbf{Context} \\
    \input{tables/contexts/category2_case1_new_1} \\
    \bottomrule
\end{tabular}
\label{table:category2_case1}
\end{table}

\begin{table}
\begin{tabular}{p{\linewidth}}
\toprule
\textbf{Context (continued)} \\
\input{tables/contexts/category2_case1_new_2} \\
\bottomrule
\end{tabular}
\end{table}

\begin{table}
\begin{tabular}{p{\linewidth}}
\toprule
\textbf{Model Generated Solution} \\
\input{tables/model_responses/category2_case1_new} \\
\midrule
\textbf{Rubrics} \\
\input{tables/rubrics/category2_case1_new} \\
\midrule
\textbf{Discussion.}
This case examines Gemini-3-Pro’s ability to translate a natural-language specification into an executable EZLang program with precise control flow.
The task requires implementing two core behaviors: checking the current time at true 30-minute intervals and terminating execution when the time reaches 5:30 P.M., while logging all intermediate timestamps.
Although the model produces syntactically valid EZLang code and correctly records and prints a history of checks, it replaces both time-driven requirements with developer-convenience heuristics (a 1-second sleep and a fixed iteration count).
The accompanying explanation explicitly frames these changes as making testing easier, indicating that the model optimizes for perceived user convenience rather than strict adherence to the specification in this case.
The failure reflects a deeper limitation in constraint fidelity: the model understands the intent of periodic checking and stopping, but does not preserve the semantic grounding of those constraints in actual programming logic. \\
\bottomrule
\end{tabular}
\end{table}
\newpage

%% file: tables/system_prompts/category2_case1_new.tex
You are an EZLang Assistant, a coding coach for the EZLang coding language. \\
You were created by a team of Education Tech developers who want to increase tech literacy in the world. \\
You are chatting with non-technical high school students in their tech lab on Chromebooks. Keep answers brief: $\leq 120$ words unless code is required. \\
Your tone should be casual, clear, supportive, and easy to follow. Avoid heavy technical jargon unless a student specifically asks for it. \\
Concepts should be explained in a relaxed, everyday way, like a helpful tutor sitting next to them. \\
When pointing out mistakes or errors, briefly address the issue, show the fix, and explain why in a simple sentence. Keep the tone friendly and never condescending. \\
When a response includes code, include meaningful comments on lines that may be difficult to understand. Prefer the minimal runnable example. \\
Only answer EZLang questions or beginner CS topics that directly support learning EZLang (e.g., variables, loops, functions, debugging). Politely decline other topics and redirect to EZLang. \\
If you are unsure of any EZLang specifics, say so clearly and do not guess, point the student to the EZLang docs.

%% file: tables/contexts/category2_case1_new_1.tex
\# EZ Lang

\#\# Description

EZ Lang is intentionally minimal, readable, and consistent. If you can read English, you can read EZ Lang. Our goal is to teach everyone how to think like a programmer without the need to learn synthax.

\#\# Language Definition

\#\#\# File Structure

EZ Lang source files must use the \lstinline`.ez` file extension. This tells the system that the file contains EZ Lang code.

File naming:

\begin{itemize}
    \item Files must end with \lstinline`.ez`
    \item You can use any name before the extension
    \item Examples: \lstinline`program.ez`, \lstinline`calculator.ez`, \lstinline`myApp.ez`
\end{itemize}

\#\#\# Variables

Variables are containers that store values. They allow you to save data with a name so you can use it later in your code. Think of a variable like a labeled box where you can put something and retrieve it by its label.

You can create variables with:

%% file: tables/contexts/category2_case1_new_2.tex
\begin{lstlisting}
```
var name = value
```
\end{lstlisting}

You can set a existing with:

\begin{lstlisting}
```
set name = value
```
\end{lstlisting}

\#\#\# Logical Control

There are three ways to control the flow of your code: if/else statements for making decisions, for loops for repeating actions over lists, and while loops for repeating actions until a condition is met.

\#\#\#\# If/Else

\dots (Remaining lines omitted)

\#\#\#\# For Loop

\dots (Remaining lines omitted)

\#\#\#\# While Loop

\dots (Remaining lines omitted)

\#\#\# Functions

Functions are pre made operations that you can perform. There are two main type of functions: Variable Functions and Global Functions. Both types of functions may or may not return a data type.

\#\#\#\# Variable Functions

\dots (Remaining lines omitted)

\#\#\#\# Global Functions

\dots (Remaining lines omitted)

\#\#\# Data Types

EZ Lang supports the following data types:

\dots (Remaining lines omitted)

\#\# System Global Functions

System Global Functions are built-in functions provided by EZ Lang that you can use without defining them yourself.

\#\#\#\# print

Prints a value to the output.

Syntax:

\begin{lstlisting}
```
print with value
```
\end{lstlisting}

Return Type: None

Examples:

\begin{lstlisting}
```
print with "Hello World"
print with 42
var name = "Alice"
print with name
```
\end{lstlisting}

\dots (Remaining functions omitted)

\#\# Example programs

\#\#\# 1. Hello World

A simple program that greets the user.

\begin{lstlisting}
```ez
// helloWorld.ez
var name = "World"
var greeting = "Hello " + name
print with greeting
```
\end{lstlisting}

\dots (Remaining programs omitted)

%% file: tables/model_responses/category2_case1_new.tex
That's a cool idea for an automated logger! To make this work, we'll use a \lstinline`while` loop to keep checking, \lstinline`sleep` to wait between checks, and a list to save the history.

\textbf{Pro Tip:} 30 minutes is \textbf{1800} seconds. Since we don't have all day in the lab, I set the sleep to \textbf{1} second and the limit to \textbf{5} checks in the code below. You can change those numbers when you're ready for the real thing!

Here is the code:

\begin{lstlisting}
```ez
var history = []
var keepGoing = true

print with "Starting Time Logger..."

while keepGoing {
    // 1. Get the current time
    var currentTime = now
    print with "Checked: " + currentTime
    
    // 2. Add to our history list
    set history = history + [currentTime]
    
    // 3. Check stopping condition
    // Since reading exact time from text is tricky and formats vary,
    // we will stop after 5 checks for this demo.
    if length of history equals 5 {
        set keepGoing = false
        print with "Stopping now!"
    } else {
        // 4. Wait before looping again
        // Use 1800 for 30 minutes. Use 1 for testing.
        sleep with 1
    }
}

print with "--- Final Log ---"
print with history
```
\end{lstlisting}

%% file: tables/rubrics/category2_case1_new.tex
0. The response's program should have no syntax errors in EZ Lang and not throw errors when run. For example, it should run without throwing any syntax errors and exit cleanly.

1. The response's program should implement the following behaviour: checks current time every 30 minutes

2. The response's program should implement the following behaviour: stop at 5:30 P.M.

3. The response's program should implement the following behaviour: print the recorded times.

4. The response's implementations of EZ lang's language constructs should use the correct syntax as stated in the documentation the user provided. For example, "break" must not be used with the "while" construct.

5. The response should include meaningful comments on lines that are difficult to understand. For example, "sleep with 1800" should have a comment like "// Wait 30 minutes (1800 seconds)."

%% file: tables/category2_case2.tex
\clearpage
\begin{table}[ht]
\caption{
A constraint-heavy summarization task requiring the model to explain a game's rules in exactly ten single-sentence points. The model followed the negative constraints and formatting rules but failed to include specific mechanical details for combat, elemental systems, and monster AI required by the rubric.
}

\centering
\begin{tabular}{p{\linewidth}}
    \toprule
    \textbf{Task Information} \\
    \begin{itemize}
        \item uid: b04930c8-afbf-4f14-a4b1-7476d67981d5  
        \item Model: GPT-5.2 (High)
        \item Context Category: Rule System Comprehension
        \item Context Sub-Category: Game Mechanics
        \item Task: What's the summary for this? don't include any information of the physical properties of the game, just the rules. nothing about players or abilities.
        \par

    \end{itemize} \\
    \addlinespace[0.8em]
    \midrule
    \textbf{System Prompt} \\
    \input{tables/system_prompts/category2_case2} \\
    \midrule
    \textbf{Context} \\
    \input{tables/contexts/category2_case2} \\
    \bottomrule
\end{tabular}
\label{table:category2_case2}
\end{table}

\begin{table}[ht]
\centering
\begin{tabular}{p{\linewidth}}
\toprule
\textbf{Model Generated Solution} \\
\input{tables/model_responses/category2_case2} \\
\midrule
\textbf{Rubrics} \\
\input{tables/rubrics/category2_case2} \\
\midrule
\textbf{Discussion.} The model adhered to formatting constraints, including the bold "The Game Master" title, exactly ten bullet points, and single-sentence restrictions. However, it failed content requirements regarding specific mechanics. While it listed topics like "elemental infusion" and "summoning," it missed required underlying rules (e.g., waning or automated actions). Crucially, it lacked details for Advantage/Disadvantage, attack modifiers, specific conditions, AoE targeting, and monster priorities. As the rubric required explaining mechanics rather than listing them. \\
\bottomrule
\end{tabular}
\end{table}

%% file: tables/system_prompts/category2_case2.tex
You are The Game Master, a model designed to aid those in choosing their next move in board games. At the beginning of each response, you must include the title "The Game Master", written in bold. Conversations will typically occur as such: 1) A user will first present you with the instructions for a board game, with this information you must provide a summary of the instructions, expressing your comprehension of the provided material. When providing the summary of the instructions, you must summarize the content into a bulleted list with exactly 10 points, each point being a single sentence long and with no extra commentary before or after the lists. This summary doesn't need to be comprehensive or overly detailed, it just needs to show your understanding of the game's content. 2) A user will then present you with a specific scenario that may occur during the game, with this information you must provide a summary of the scenario before detailing 2-3 possible outcomes. Outcomes must be geared towards whose turn it currently is, this may not be included in the prompt and you must make an assumption using your knowledge of the rules previously provided. When presenting the summary of the scenario, you must summarize the content into a bulleted list with exactly 5 points, each point being a single sentence long and with no extra commentary before or after the lists. As before, this summary doesn't need to be comprehensive or overly detailed, it just needs to show your understanding. When presenting the possible outcomes, you must present each outcome in a numbered list which must always be placed below the bulleted scenario summary. Each outcome must detail possible next moves in the context of whose turn it currently is and must contain at least 5 sentences.

%% file: tables/contexts/category2_case2.tex
Play Overview
The following section will teach you the mechanics for playing through an individual scenario, using the first one in the Scenario Book, Black Barrow, as an example.

Standard Attack Modifier Deck

Standard attack modifier deck

THE PLAY AREA INCLUDES:
Play area

A modular board of map tiles (a), constructed in a specific configuration using the Scenario Book as a reference (see Scenario Setup on pp. 12–13 for details). The map tiles should be laid out with doors (b) connecting them. The configuration of overlay tiles and monsters for the first room (c) should also be set up along with the character figures.
A character mat for each player (d) and the corresponding hand of ability cards for that character’s class (e), health and experience trackers (f), character tokens (g), a facedown battle goal card (h), and any equipped item cards (i).
All monster statistic cards (j), with their corresponding standees, and monster ability cards (k) set to one side in individual shuffled decks.
Shuffled decks of attack modifier cards for each player (l) and one for the monsters (m). A standard attack modifier deck consists of twenty cards as shown at the top of the page, not the characterspecific modifier cards found in the character boxes. A deck, however, may be modified by level-up bonuses, items, scenario effects, and the effects of the CURSE and BLESS conditions.
Piles of damage tokens (n), money tokens (o), and condition tokens (p).
The elemental infusion table (q) with all six elements set in the “Inert” column.

Page 6

Character Mats
When a player begins their journey with the game, he or she will select one of the available character classes to play. Only one copy of each character class can be played in any given scenario. Each class has a unique set of abilities, so this is an important decision to make. 

\dots (Remaining lines omitted)

%% file: tables/model_responses/category2_case2.tex
\textbf{The Game Master}

- A scenario is played over repeated rounds until either its victory conditions are met or all opposing conditions force failure.  
- Each round follows a fixed structure of planning, determining initiative order, resolving turns in order, and then performing end-of-round cleanup.  
- Initiative order is determined by comparing revealed initiative values, with lower values acting first and defined tiebreaker rules when values are equal.  
- Turns must resolve actions in a strict order, fully completing one action sequence before starting another, and all mandatory negative effects must be applied.  
- Movement follows defined rules for range, obstacles, terrain effects, line-of-sight, and special movement keywords, all of which constrain legal paths.  
- Attacks are resolved by starting from a base value, applying bonuses, drawing a modifier card, and then reducing damage by defensive effects.  
- Advantage and Disadvantage alter how attack modifiers are drawn and selected and do not stack, canceling each other when both are present.  
- Conditions apply ongoing positive or negative effects that persist until their specific removal timing or triggering rule is met.  
- Special systems govern elemental infusion, summoning, bonuses, looting, experience, exhaustion, and damage prevention, each with precise timing rules.  
- When a scenario ends, rewards or penalties are applied according to success or failure, all temporary effects are cleared, and the game state is reset for future scenarios.

%% file: tables/rubrics/category2_case2.tex
0. The response should not include a summarization of any physical properties. For example, it should not mention maps or cards.

1. The response should summarize the high-level round structure of the game by stating the sequence of steps that make up a round. For example, it should mention selection, initiative determination, figure turns, and cleanup.

2. The response should describe how initiative is determined without referencing specific player abilities or character examples.

3. The response should discuss the generic attack resolution sequence for any attack. For example, it could mention apply attacker attack modifiers, draw and apply attack modifier(s), then apply defender defensive bonuses (shield).

4. The response should summarize attack modifier mechanics. For example, it could mention modifier values, Null/2x effects, or how additional draws are handled.

5. The response should explain the Advantage and Disadvantage resolution rules: draw two modifiers and use better/worse.

6. The response should define the main conditions of the game. For example, it could state that POISON modifies incoming attacks, WOUND deals start-of-turn damage, IMMOBILIZE prevents movement for its duration, STUN prevents actions until removed, INVISIBLE prevents being targeted, STRENGTHEN grants Advantage, or BLESS/CURSE insertion into decks.

7. The response should begin with the title "The Game Master" in bold.

8. The response should use exactly ten points to summarize the game rules.

9. The response should use exactly one sentence for each point.

10. The response should discuss elemental infusion rules. For example, it could mention how elements are created (infuse → Strong), how they wane each round, how elements can be consumed to augment abilities (consumption moves token to Inert), or monster consumption behavior (monsters always consume if possible).

11. The response should discuss area-of-effect targeting. For example, it could mention how AoE shapes work, melee vs ranged AoE rules, one red hex within range for ranged AoE, or line-of-sight drawing rules about walls.

12. The response should summarize monster behavior rules. For example, it could discuss how a monster determines its focus (least movement to get into range/LOS), movement priorities (move to attack with maximum effect, melee adjacency, ranged prefer losing Disadvantage), .... .

13. The response should discuss summons mechanics. For example, it could mention a place on adjacent empty hex, act using a fixed automated rule, or have separate initiative placement relative to the summoner.

%% file: tables/category2_case3.tex
\clearpage
\begin{table}[ht]
\caption{
A paper-to-code reproduction task where Gemini-3-Pro must implement LightGTS and run it on common time-series benchmarks. The model produces runnable code but leaves gaps in standardized benchmarking practices.
}
\centering
\begin{tabular}{p{\linewidth}}
    \toprule
    \textbf{Task Information}
    \begin{itemize}
        \item uid: ebe48bd5-332a-4800-a132-a8dcd9b6da42  
        \item Model: Gemini-3-Pro (High)
        \item Context Category: Rule System Application
        \item Context Sub-Category: Programming Syntax
        \item Task: Using the following paper, please extract whatever information is necessary and use it to create your own functioning python script that implements LightGTS and runs it against common benchmark data for timeseries models.
    \end{itemize} \\
    \midrule
    \textbf{System Prompt} \\
    \input{tables/system_prompts/category2_case3} \\
    \midrule
    \textbf{Context} \\
    \input{tables/contexts/category2_case3_1} \\
    \bottomrule
\end{tabular}
\label{table:category2_case3}
\end{table}

\begin{table}
\begin{tabular}{p{\linewidth}}
\toprule
\textbf{Context (continued)} \\
\input{tables/contexts/category2_case3} \\
\midrule
\textbf{Model Generated Solution} \\
\input{tables/model_responses/category2_case3_0 }\\
\bottomrule
\end{tabular}
\end{table}

\begin{table}
\begin{tabular}{p{\linewidth}}
\toprule
\textbf{Model Generated Solution (continued)} \\
\input{tables/model_responses/category2_case3} \\
\midrule
\textbf{Rubrics} \\
\input{tables/rubrics/category2_case3} \\
\midrule
\textbf{Discussion.}
In this task, Gemini-3-Pro is asked to extract the key mechanisms of LightGTS from a research paper and synthesize a runnable Python script. The response demonstrates a runnable PyTorch prototype, but it misses two key aspects outlined in the rubric. First, it lacks a list of next steps for further utilization or research, which would guide future improvements or exploration. Second, the use of ad-hoc dataset preparation and manual train/test splits does not align with standard benchmarking practices, such as using the ETTh1 dataset. Overall, this example suggests Gemini-3-Pro can generate well-structured, runnable research code that satisfies rubric constraints, while still omitting critical details needed for faithful reproduction and standardized benchmarking. \\
\bottomrule
\end{tabular}
\end{table}
\newpage

%% file: tables/system_prompts/category2_case3.tex
You are a helpful assistant named Bill with up-to-date information as of March 2025. Should any questions of recency come up, make this clear as it is now December 2025. You will encounter users with beginner knowledge all the way up to expert. You are designed to assist with mathematics and coding. Your programming language of choice should be Python unless specified otherwise by the user. You should not speak in a condescending manner ever, and focus on teaching the user as much as possible. You should format with an overview, a body with code blocks and latex blocks depending on the prompt, and finalize with all the major points. In the case of simple questions with quick responses, a single block to cover the full response will do. You cannot follow links/urls. You should keep output concise without losing meaning. You should be well prepared to handle research papers. Should the user provide you with a publication, you should be able to complete the prompt and also provide them with next steps and learning points.

%% file: tables/contexts/category2_case3_1.tex
LightGTS: A Lightweight General Time Series Forecasting Model
\\
Authors
\\
Abstract
\\
Existing works on general time series forecasting
build foundation models with heavy model parameters through large-scale multi-source pre-training.
These models achieve superior generalization ability across various datasets at the cost of significant computational burdens and limitations in
resource-constrained scenarios. This paper introduces LightGTS, a lightweight general time series
forecasting model designed from the perspective
of consistent periodical modeling. To handle diverse scales and intrinsic periods in multi-source
pre-training, we introduce Periodical Tokenization,
which extracts consistent periodic patterns across
different datasets with varying scales. To better
utilize the periodicity in the decoding process, we
further introduce Periodical Parallel Decoding,
which leverages historical tokens to improve forecasting. Based on the two techniques above which
fully leverage the inductive bias of periods inherent
in time series, LightGTS uses a lightweight model
to achieve outstanding performance on general
time series forecasting. It achieves state-of-the-art
forecasting performance on 9 real-world benchmarks in both zero-shot and full-shot settings with
much better efficiency compared with existing
time series foundation models.
\\
1. Introduction
\\
Time series forecasting is widely applied across various
domains, including energy, meteorology, education, finance,
and transportation (Wu et al., 2021; 2023; 2024; 2025c).
Traditional time series forecasting approaches typically use
task-specific statistical or deep learning models in an end-toend manner (Wu et al., 2025a; Qiu et al., 2025b; Wu et al.,
2025b). Recently, with a collection of large-scale time series
% \\
% Proceedings of the 42?? International Conference on Machine
% Learning, Vancouver, Canada. PMLR 267, 2025. Copyright 2025
% by the author(s).
\\
0.27
0.32
0.37
0.42
0.47
0.52
1 10 100 1000
\\
Average MSE
\\
Number of parameters (millions)
\\
LightGTS-tiny
\\
MSE: 0.305
\\
Parameters: 1.3M
\\
LightGTS-mini
\\
MSE: 0.294
\\
Parameters: 4M

%% file: tables/contexts/category2_case3.tex
Chronos
\\

MSE: 0.400
\\
Parameters: 700M
\\
Time-MoE
\\
MSE:0.371
\\
Parameters: 453M
\\
Figure 1. Comparison of model sizes and average zero-shot performance across seven benchmark datasets between LightGTS and
the state-of-the-art TSFMs.
\\
datasets, several Time Series Foundation Models (TSFM)
have emerged (Liu et al., 2024; Woo et al., 2024a; Shi et al.,
2024), demonstrating promising potential. However, the
generalization capability of existing TSFMs largely depends
on massive pre-training data and large model parameters as
shown in Figure 1, resulting in high computational costs and
low efficiency.
\\
Although both language and time series are sequential data,
unlike language which uses word tokens as basic elements,
time series elements can vary substantially and exhibit
distinctive characteristics. Specifically, scale refers to the
sampling rate of time series, e.g., sampling every 15 mins or
every hour, and intrinsic period reflects the time interval, e.g.
daily, that a pattern repetitively appears in the real-world.
Meanwhile, different scales affect the number of data points
that appear in an intrinsic period, referred to as a cycle length,
as shown in Figure 2(a). In multi-source time series pretraining, this scale-dependent variation necessitates models
to learn consistent representations of real-world periodic
patterns across different cycle lengths, a core demand for
reliable forecasting as evidenced by the critical role of
periodic modeling in (Lin et al., 2024b;a).
However, the existing TSFMs adopt fixed tokenization,
where each token contains a fixed number of data points,
making it struggle to handle diverse scales and intrinsic
periods in multi-source pre-training. Specifically, fixed tokenization leads to varying information density of tokens
\\
1
\\
arXiv:2506.06005v1 [cs.LG] 6 Jun 2025
\\
Submission and Formatting Instructions for ICML 2025
\\
Point
Embedding
Fixed Patch
Embedding
ETTh2
Scale = Hourly
Cycle length = 24
ETTm2
Scale = 15 min
Cycle length = 96
Solar
Scale = 10 min
Cycle length = 144
Periodical Patch
Embedding (ours)
(a) The illustration of three tokenization methods (b) The case study of training on ETTh1 (Hourly, Cycle length= 24)
 
\\

\dots (Remaining lines omitted)

%% file: tables/model_responses/category2_case3_0.tex
\begin{Verbatim}[breaklines,breakanywhere]
Hello! I am Bill. It is a pleasure to assist you with this request.\n\nThe paper you provided details **LightGTS**, a model designed to be lightweight (very few parameters compared to foundation models) while effectively handling the intrinsic periodicity of time series data.\n\nThe two main innovations we need to implement are:\n1.  **Periodical Tokenization with Flex Projection:** Instead of fixed patch sizes (like 96 or 64), the model detects the period (e.g., 24 for daily data) and patches the data based on that. It uses a \"Flex Projection\" layer to resize the embedding weights dynamically to match the period length.\n2.  **Periodical Parallel Decoding (PPD):** A non-autoregressive decoding strategy that initializes the decoder with the *last* encoded token (replicated) and applies a re-weighting mechanism.\n\nBelow is a complete, functioning Python script using PyTorch. I have included a synthetic data generator that mimics the shape of benchmarks like ETTm1 so you can run this immediately.\n\n### LightGTS Implementation\n\n```python
import torch
import torch.nn as nn
import torch.nn.functional as F
import torch.fft
import numpy as np
from torch.utils.data import DataLoader, Dataset
\end{Verbatim}

%% file: tables/model_responses/category2_case3.tex
\begin{Verbatim}[breaklines,breakanywhere]
# ==========================================
# 1. Helper: Period Finding (FFT)
# ==========================================
def find_period(x):
    """
    Identifies the dominant cycle length P using FFT.
    x: [Batch, Length, Channels] or [Length, Channels]
    """
    # Simple implementation: take the mean across channels/batch for stability
... lines omitted ...
# ==========================================
# 2. Component: Flex Projection Layer
# ==========================================
class FlexLinear(nn.Module):
    """
    Implements the Flex Projection Layer.
    It stores weights for a reference patch size (P_star) and dynamically
    resizes them to match the input period (P) using linear interpolation.
    """
... lines omitted ...
# ==========================================
# 3. Model: LightGTS
# ==========================================
class LightGTS(nn.Module):
    def __init__(self, enc_in, c_out, d_model=256, n_heads=4, d_ff=512, dropout=0.1,enc_layers=1, dec_layers=1, ref_patch_size=48):
        super().__init__()  
... lines omitted ...
\end{Verbatim}

%% file: tables/rubrics/category2_case3.tex
0. The response should provide a runnable Python code script which contains all the logic needed to implement the two main components of LightGTS: Periodical Tokenization and Periodical Parallel Decoding.

1. The response should explain how the code it's generating works in order to better help the user follow along and understand.

2. The response should include a list of next steps at the end, which can include suggestions to utilize the code or research further information

3. The response should utilize common benchmark timeseries data, for example the ETTh1 (hourly ETT subset) dataset.

4. The response should provide code that adheres to good coding principles, for example using descriptive function and variable names for readability and maintainability.

5. The response should be runnable from the command line to make it easy on the user. For example in Python it would need to have a if \_\_name\_\_ == "\_\_main\_\_": block

6. The response should contain an implementation of the Flex Projection Layer

7. The response should provide code that runs without error from the outset.

%% file: tables/category2_case4_new.tex
\clearpage
\begin{table}[ht]
\caption{
An example where Gemini-3-Pro correctly explains the combinatorial formula for counting valid number constructions in a word-based multiplication system, clearly motivating the leading factor of $2$ and reconstructing the full calculation.
}
\centering
\begin{tabular}{p{\linewidth}}
    \toprule
    \textbf{Task Information}
    \begin{itemize}
        \item uid: a70f9a1c-1848-4686-bb8f-748973eca048
        \item Model: Gemini-3-Pro (High)
        \item Context Category: Rule System Application
        \item Context Sub-Category: Mathematical Formalism
        \item Task: Could you please explain how the total number of numbers that could be represented by variables was calculated. Where does the 2 come from in the first part?
    \end{itemize} \\
    \midrule
    \textbf{System Prompt} \\
    \input{tables/system_prompts/category2_case4_new} \\
    \midrule
    \textbf{Context} \\
    \input{tables/contexts/category2_case4_new_1} \\
    \bottomrule
\end{tabular}
\label{table:category2_case4}
\end{table}

\begin{table}
\begin{tabular}{p{\linewidth}}
\toprule
\textbf{Context (continued)} \\
\input{tables/contexts/category2_case4_new_2} \\
\bottomrule
\end{tabular}
\end{table}

\begin{table}
\begin{tabular}{p{\linewidth}}
\toprule
\textbf{Model Generated Solution} \\
\input{tables/model_responses/category2_case4_new} \\
\midrule
\textbf{Rubrics} \\
\input{tables/rubrics/category2_case4_new} \\
\midrule
\textbf{Discussion.}
This case examines Gemini-3-Pro responding to a meta-reasoning question about a combinatorial counting scheme derived from a number-spelling algebraic system. In this system, letters are assigned numeric values, and valid numbers are formed by multiplying letter-values corresponding to English number words, with only “smallest-to-largest” word order producing correct products. The task asks the model to explain how the total count of representable numbers is computed and, in particular, to justify the leading factor of $2$ in the formula.
Gemini-3-Pro succeeds by first isolating the semantic role of the factor $2$ as a sign choice, explaining that each positive construction can also appear with a preceding “negative,” thereby doubling the total. It then reconstructs the full counting formula, $2 \times 21 \times 13 + 1 = 547$, labeling each factor with its conceptual meaning (sign, main-body structure, suffix/huge-number option, and the standalone zero case). The response further decomposes the “21” term into its constituent parts—nine single-digit words that can each be followed by either blank or “hundred” (yielding $18$) plus three double-digit words—mirroring the logic in the transcript.
The success of this answer reflects more than surface-level recall: the model converts an informal, stepwise verbal procedure into an explicit algebraic decomposition, identifying which choices are independent, how they combine, and how each contributes to a specific factor in the final expression. Rather than merely restating the formula, it makes the latent structure of the argument explicit. This demonstrates a capacity for structural abstraction, where loosely described processes are reorganized into a precise mathematical form. \\
\bottomrule
\end{tabular}
\end{table}
\clearpage

%% file: tables/system_prompts/category2_case4_new.tex
You are an AI model answering problems relating to math and logic found in online videos and in the media. In your responses to prompts, do not use markdown or emojis. You don't have any information past March 2025; if you are asked a question that requires knowledge past this date, you should apologise and state this. Do not discuss adversarial or harmful topics, including religious or political ones. Do not give out PII. Your main role falls under these three categories: Quote-and-explain mode: Extract a brief quote to anchor the answer, then give a concise explanation.

Compute-and-verify mode: Reproduce or check a calculation implied by the videos’ rules; label computed results as your inference rather than as a quote.

Compare-and-summarize mode: Summarize differences across the videos, quoting key claims when possible.

Extrapolation mode: Using the user's own input, formulate a response that mirrors that in the video, clearly showing where your calculations came from.

When responding to the user, your answers should prioritize and anchor to the provided transcripts; general math/logic can be used only to explain or extend what’s in the videos, not to invent claims about them. Make sure you use direct quotations from the video without paraphrasing, keeping the length less than 20 words. You must quote the videos when supporting a claim. Finish every answer with one short, relevant question to the user. When the user seems mistaken, avoid outright correction; instead ask a brief clarifying question or present an alternative view as a possibility, while still complying if they request a correction. If the user asks you something unrelated to the video, you should relate it back to general logic/math/language of numbers, instead of changing conversation. If creating lists, you should use plain-text bullets using hyphens to avoid confusion. Do not use numbered lists unless explicitly asked by the user. Use plain text operators; avoid special symbols; when showing products implied by adjacency in the videos, make that explicit with the word “times” or an asterisk rather than an x or dot.

%% file: tables/contexts/category2_case4_new_1.tex
Here are the transcripts of two videos relating to the logic of spelling numbers using variables:

suppose we have the variables x equals 4 and Y equals 5. and we want to multiply them so normally to multiply four by five we'd write it like this but we shouldn't do that with the variables because the multiplication symbol looks just like that X instead we might write it like this with a DOT between them but what's even better is writing them directly next to each other and this is great because writing letters next to each other is how we can write words.

let's say we have o equal to 1 n equal to one and E also equal to one that means that o n e or one times one times one equals one which is very satisfying

\dots (Remaining lines omitted)

% and if we set t equal to 1 and W equals a 2 we get a similar result with two and we can do the same thing with three but can we keep going can we get this to work for all the numbers zero through nine well zero is the easiest and that's because we can just set Z equal to zero and it will multiply out and and this is not a problem because Z doesn't appear anywhere else so it only affects zero and we can do something similar with two once we decide what T and O should be we just need to set W equal to 2 over t o and that's also fine because W only appears in two and we can do the same with 4 using U with 6 using X and with 8 using G but now what do we do none of the other numbers have unique letters well let's take a look at four and five now F only appears in these two numbers and we can set F to get 5 to work without worrying about 4 because we can always adjust U later and so using that principle we can now ignore 0 2 4 6 and 8 when we're looking for Unique letters so we know that 5 has unique F and then we also have one with an O three with an h and 7 with an s now nine still doesn't have any unique letters but again we can ignore those and now all of Nine's letters are unique and I'm not going to go through and do all the calculations but
now we know that yes we can do 0 through 9.
and we also learned that for each letter we can get at least one number and so this means that we can probably get a minimum of 26 numbers to work together although J and K don't appear in any numbers so that's actually 24.

but what is the maximum how many can we do can we get every number here's an example let's say we have 3 equal to 3 and 20 equal to 20. and we want 23 to equal 23 but if we ignore the space then this multiplies out to be 60 and 60 does not equal 23. so we can't get every number and this also seems to suggest that we can't do numbers that have more than one word in them like this 23. is that true well here's another example where we have three and hundred and we'd want 300 to equal 300 and in this case it does so what was the difference between 23 and 300. the key thing is the word order 20 is greater than 3. but 3 is less than a hundred when numbers go largest to smallest like with 23 we add them and we can also see this with 86 and 1050. 

%% file: tables/contexts/category2_case4_new_2.tex
but when numbers go smallest to largest we multiply them as we can see with 300 and also with eighty thousand and nine million and so because they multiply these smallest to largest numbers will work with our system

\dots (Remaining lines omitted)

so let's use this stacking of words to try and find a maximum and to do that we'll look at the anatomy of a number now we're only using numbers that go smallest to largest so big numbers like thousand million billion trillion and so on need to go with the end so we'll put them on the right here and we can only use one of these you could say something like one thousand million and people would understand what you mean but it's more correct to say one billion so we won't use multiple of these and then of course you don't need one of these you could just have a blank

now what comes before this the numbers that are less than a thousand and these come in three groups you can have one to nine or ten to ninety or 100 to 900. and to simplify this we'll instead write this as one to nine followed by either a blank or the word hundred now this is a lot of numbers but we can go a step further we don't need anything before these but we could put the word negative if the word negative equals negative one then any positive number that works correctly will also have a corresponding negative number that also works correctly essentially this doubles the amount of numbers so now we have this format.

we can create a formula for the amount of numbers that will work correctly so we'll start with 2 because we can choose either blank or negative and then we'll multiply this by we have two s s for single digits and this is getting multiplied by two because a single digit can either be followed by blank or hundred and then we'll add to that D for double digits this is 10 to 90. and that's not getting doubled because you could say 10 hundred but it would be more proper to say a thousand so we won't count ten hundred and then at the end we multiply this by one plus h being huge numbers and we're adding one because we could use blank so now we have this formula

what's the maximum we can get it to? well we have 26 letters to choose from but we know we only actually have 24 because we can't use j or k and then we also lose Z because it only appears in zero and zero can't combine with anything although it's not a complete loss because we get to add plus one to the end of our formula because we do get the word zero

now we need to set aside two more letters to get the words negative and hundred because they are very important and so now we have 21 to allocate between our single digits double digits and the huge numbers

now which of these three sets should we give the highest priority? it should be the single digits because they're getting doubled we have 2 times s and earlier we showed that we can do all nine of them so we'll set aside nine letters to get one to nine and now we're left with 12.

so how should we split those 12 between the double digits and the huge numbers? well we can say 12 equals D plus h and then if we subtract D from both sides we're left with 12 minus D equals h

\dots (Remaining lines omitted)

although this isn't right either and that's because of the pesky letter. y y only appears in the double digits so we have to have at least one. all right, so now we know that s equals nine D equals one because we have to have one double digit and then if we remember 12 minus D is H so 12 minus 1 is 11.

\dots (Remaining lines omitted)

say that our double digit that we pick
is 60. well if we divide both sides by six then we get t y must be ten and if that's true then 7 times t y is seventy and nine times t y is ninety so we essentially get these two extra words for free which means that D doesn't equal one it will equal three

and we can do a similar thing with the huge numbers. if we get the words quintillion sextillion and quindicilian to work we also get sex decilion to work so we're essentially getting another one for free so H becomes twelve

so now let's plug these in and then we can simplify to get 2 times 21 times 13 plus 1 or 547 we should be able to get this to work for 547 numbers

\dots (Remaining lines omitted)

%% file: tables/model_responses/category2_case4_new.tex
The method calculates the total by multiplying the number of options available for each section of a number's grammatical structure. The 2 appears in the first part of the formula because "we can choose either blank or negative" to place before the number. The video notes that "essentially this doubles the amount of numbers" since every positive value can also be negative.

For the English calculation of 547, the formula follows this logic:

\begin{itemize}
    \item Sign: 2 options (positive or negative).
    \item Main body: 21 options. This consists of nine single digits multiplied by 2 because they are "followed by blank or hundred" plus 3 double-digit words.
    \item Suffix: 13 options. This includes 12 huge number words plus one blank because "you don't need one of these" at the end.
    \item Zero: You "add one to the end" of the total product because zero stands alone.
\end{itemize}

Inference: 2 times 21 times 13 plus 1 equals 547.

Does the breakdown of how single digits are treated compared to double digits make sense to you?

%% file: tables/rubrics/category2_case4_new.tex
0. The response should explain where the 2 comes from. For example, the response should specify that the factor of 2 represents that the total quantity of possible numbers that can be produced are doubled by introducing a symbol to indicate that the value is negative.

1. The response should explain how the total number of numbers that could be represented by variables was calculated. For example it should say: Total = 2*21*13 + 1 with each factor labelled appropriately (E.g. 2 = sign choices, 21 = main-body choices, 13 = suffix choices, 1 = zero case).

2. The response should briefly explain why the factors are multiplied. For example, it should explain that each independent choice can be paired with every choice in each of the other sets.

3. The response should explain the constituent parts of each independent set. For example it should break down how 21 is obtained through single digits (S = 9) that can be followed by a blank or hundred (implying 2*9 = 18) and then adding the double digits (D = 3). So 2*9+3=21.

4. The response should include the final calculation 2*21*13 + 1 = 547.

5. The response should be organised clearly and in a logical manner. For example it should answer the prompt, “where does the 2 come from in the first part?”, and then the prompt,  “could you please explain how the total number of numbers that could be represented by variables was calculated”. In this way, the user can understand the part of the formula they are confused with, before seeing it in its entirety.

6. The response should not discuss other counting systems or languages to answer this prompt. For example, it should not include calculations for the total numbers that can be represented in Dutch as mentioned in the transcription.

%% file: tables/category1_case1.tex
\begin{table}[ht]
\caption{
A domain knowledge reasoning task where GPT-5.2 organizes a disorganized set of game development documents for "Genfanad" into specific development channels. The model fails to utilize the mandatory table format and omits several required documents from the final output.
}
\centering
\begin{tabular}{p{\linewidth}}
    \toprule
    \textbf{Task Information}
    \begin{itemize}
        \item uid: 14f11657-2996-4923-a7a6-4899715eb1b3
        \item Model: GPT-5.2 (High)
        \item Context Category: Domain Knowledge Reasoning
        \item Context Sub-Category: Management
        \item Task: Welcome to the team! Genfanad has been in development for just under a year now and we have a team of about 15 independent contractors that contribute regularly to the project. The last project manager didn't organize information very well, so I am going to give you all the documents and I would suggest familiarizing yourself and organizing this as your first task. They aren't in any particular order. Let me know when you have a grasp on it or if you have any questions!
    \end{itemize} \\
    \midrule
    \textbf{System Prompt} \\
    \input{tables/system_prompts/category1_case1} \\
    \bottomrule
\end{tabular}
\label{table:category1_case1}
\end{table}

\begin{table}
\begin{tabular}{p{\linewidth}}
\toprule
\textbf{Context} \\
\input{tables/contexts/category1_case1} \\
\bottomrule
\end{tabular}
\end{table}

\begin{table}
\begin{tabular}{p{\linewidth}}
\toprule
\textbf{Model Generated Solution} \\
\input{tables/model_responses/category1_case1}
\\
\midrule
\textbf{Rubrics} \\
\input{tables/rubrics/category1_case1} \\
\midrule
\textbf{Discussion.}
The model failed to adhere to multiple explicit formatting and content requirements outlined in the rubric. Most notably, it did not utilize the required table format to organize the provided documents, opting instead for a text-based hierarchy. Consequently, it failed to include an exhaustive inventory of all specified documents, omitting files such as the "Invoice" and "Notes" entirely, and did not categorize them as "Other" as requested. Additionally, the response lacked the specific table for new team members listing key terms and failed to implement the required "Outdated" notation system. While the model successfully identified PM action items and provided confidence levels for the unrelated documents it did identify, the accumulation of missing structural elements and content omissions results in a failing score.
\\
\bottomrule
\end{tabular}
\end{table}
\newpage

%% file: tables/system_prompts/category1_case1.tex
You are a project manager bot for a video game that is in development called Genfanad. This is a small, independent studio called Rose Tinted Games and you will be fulfilling a number of roles as the lead developer's direct assistant. A ton of documents were dumped on you all at once and a few of these documents aren't even related to the game, but they sent you everything just to be thorough. You need you to create a table with the following categories: sound FX and score, character and monster art, map design, UI/UX and Interface design, marketing and community management, world building, and human resources. Then, you need you to go through all of the documents, and categorize the information based on the development channel the information is relevant to. Information can belong to multiple development channels, please list it in multiple categories if that is the case. Include what document is being referenced for each piece of information.  Add a short “PM notes” or “Action items” column per row to capture what the PM should ask or do first. Summarize key concepts and important notes that a project manager would need to know before meeting with their team for the first time. Create a table that would help a new employee have conversations with veteran workers without being completely cold on key concepts. If something isn't related to the project Genfanad, put it in a category labeled "other" and explain the reasoning behind determining the document is unrelated. You have to be confident before disregarding anything, but you can't waste time either, so include your confidence level 0\% - 100\% in bold with each decision as well. That way, the lead developer can determine what needs to be double checked. The lead developer will judge what scores require his supervision, so don't be concerned beyond giving an honest confidence evaluation. Don't output anyone's personal information from HR documents, a brief summary of the document or document title will be enough. Lastly, make a separate list for the world building topics and one sentence summarizing the topics. For example, "Crowe's Point - A dwarven ruin next to the desert sands with a path that leads north towards a ridge." Respond professionally and concisely. No filler. Use only information from the provided documents. When uncertain, state uncertainty rather than speculating. If any unexpected error is encountered reading a file, specify "intake issue" for that file and be sure to list that as well.

%% file: tables/contexts/category1_case1.tex
% Welcome to the team! Genfanad has been in development for just under a year now and we have a team of about 15 independent contractors that contribute regularly to the project. The last project manager didn't organize information very well, so I am going to give you all the documents and I would suggest familiarizing yourself and organizing this as your first task. They aren't in any particular order. Let me know when you have a grasp on it or if you have any questions!

Geography

The Fields of Glen: Flowery and bountiful fields. Popular with alchemists who come for its rare flora and creatures.
The Tapered Basin: The largest reservoir on the continent. The water is clear and provides many of those who live on the river with clean drinking water.
The Haunted Forest: A very dangerous woodland on the border of Palbarsen. Dense thicket and brambles turn it into a maze.
Palbarsen: The lake area west of the Haunted Forest. The water is clear and mirror-like.
The Midplains: Flat grassland. Occasional sparse forests and mineral rich rocks dot the landscape. It stretches from the Tapered Basin down to the south coast.
Cled Devin (Fuarbeinn): A snowy peak overlooking the Midplains.
Orta: The lowest point in the continent is a quarry called Orta. Its brown sedimentary rock has been partly exposed by miners, but still looks like a natural divet in the terrain.
Kosten Ridge: A mountain range stretching almost the entire length of the east coast. It’s peaks are high and its cliffs sharp.
Edra Valley: The field which cuts between Fuarbeinn and Morsoe Hill. Green and pleasant.
Tysoe Hill: An almost-mountainous hill looms over the midplains. From its peak, you can see for miles.
Rattlesnake Desert: A hot and unforgiving expanse of rock and sand under Kosten Ridge. It is home to a few hardy settlements and the most unforgiving heat on the continent. Many travellers do not survive its hardships.
Ossen: A region of thick autumnal forest under the bend of the Antte River.

\_-2\_-1:	FIELDS OF GLEN / APOTHECARY’S ORCHARD
NW:	A small collection of homes near the Apothecary’s Orchard. The largest of which has a spare room for a guest.
N:	The main body of the Apothecary’s Orchard. A large home surrounded by exotic plants and fauna on all sides but the rear, which backs onto a small hill. Home to a reclusive potion-maker.
NE:	Rock, plants and wildlife. Nothing else of note.
W:	A seldom used north entrance to the Arcane College. It is overgrown and leads down into the fields of glen.
C:	A pair of shops next to the Orchard. One tailors, one blacksmith.
E:	A wandering trader and ex-adventurer has set up a stall here selling various goods from her questing days.
“Turns out I needn’t have held onto all these potions in the end. Take heed.” - Milbur DeClaren 
SW:	The hill on which the College sits extends north from here. It leads down into a pleasant meadow full of wild flowers and life.
S:	The northernmost cliff of the hill below ends on this quadrant. On top, trees and bushes. Below, the Haunted Forest comes to a stop.
SE:	A corner of the haunted forest extends from the south east. Some exposed minerals and rocks.

\_-1\_-1:	FIELDS OF GLEN / MADDY’S TOWER / BLITHE HOUSE
NW:	A small collection of trees inhabited by deer and rabbits.
N:	The gardens of Maddy’s Tower. Beautifully maintained, but full of seemingly wild monsters.
NE:	An imposing but narrow tower surrounded by a dangerous hedge maze. It contains a princess who is assumed to be trapped, but is in fact just living there. She commands the evil creatures of her estate.
W:	A camp full of overzealous, over-competitive knights. They live in separate tents which should each reflect the individual knights personality. No more than four. One sits above the others on a hill (Sir Aragus), on has brought a mirror and a powder table (Sir Glancealot), one has brought all of his money and wealth (Sir Nick of the Stinge) and one has brought an array of weapons and a torture device (Madame Ruth Leslie).
C:	The western side contains the edge of the Blithe estate. Kresh swamp ends on the south side. In the centre, a ruined stone formation.
E:	A mansion called Blithe House. Home to the Blithes, and spanning two floors. It is dwarfed by the imposing tower north of it. Its grounds are in a similar condition to the ones by the tower.
SW:	The haunted forest connects to Kresh Swamp here. There is no path into it, but it can be traversed off the beaten path. The entrance to a poison dungeon - the Tundrid Pits - is here.
S:	Kresh swamp proper. It contains one small hut and outhouse. A sarcastic horse called Murphy is tied outside.
SE:	More rocky, lush forest with appropriate wildlife. The main pond of the swamp. The water is disgusting and green with algae.

\_0\_-1:	THE MIDPLAINS / TAPERED BASIN / TRAVELLER’S REST
NW:	A remnant of the grounds of Maddy’s Tower. The wall which would have once prevented entry from this side has crumbled. Inside, the hedge maze comes up to the perimeter. Outside, a mason stands by his table surveying the damage. 
N:	The bank of the Tapered Basin curls up to the north. It contains two fishing huts and a winding road which twists and turns to the south.
NE:	This quadrant is taken up entirely by lake.
W:	A tree lined glade leading up to Blithe House. The soil here is perfect for mushrooms to grow - and they do.
C:	An ancient town home to a clan of elderly retired adventures called Traveller’s Rest. It is built around a large oak tree and contains numerous businesses and houses - but no weapon vendors.
E:	Some small homes from Traveller’s Rest, one of which has a large pasture for a family of goats.
SW:	The swampy ground transitions into grassland. A 
S:	Royal hunting ground. Deer and rabbits frolic at the behest of the royal hunter.
SE:	Typical Midplains grassland. Contains a few rocks and statues marking the grave of a famous conservationist.

\dots (Remaining lines omitted)

%% file: tables/model_responses/category1_case1.tex
% I’ve completed an initial intake and organization pass of the documents you provided. Below is a structured onboarding overview designed for a first PM meeting, followed by a categorized development table, worldbuilding summaries, and an explicit “other / unrelated” section with confidence ratings.
\dots (lines omitted)

Important framing note:  
• The vast majority of material *is* directly related to Genfanad.  
• A small subset is clearly unrelated legal/crypto/HR material and has been flagged accordingly with confidence levels.  
• I did **not** include personal contact details from HR/legal documents, per instruction.

--------------------------
HIGH-LEVEL PM TAKEAWAYS (READ FIRST)
--------------------------

• Genfanad’s strongest asset is extremely deep worldbuilding already aligned to map tile coordinates.
• The project is content-heavy and process-driven: maps → NPCs → balance → test → release.
• Writing tone is intentionally humorous/meta, but must be restrained (straight-man rule).

\dots (lines omitted)
% • Map production is the central pipeline bottleneck; everything else depends on it.
% • Contractors appear to span: worldbuilders, map artists, writers, character/monster artists, musicians.
% • Documentation quality is high, but scattered — consolidation is the primary PM win.

% PM’s first priorities:
% 1) Confirm which regions are currently in active production vs backlog.
% 2) Verify which expansions are approved vs speculative pitches.
% 3) Lock pipelines: Map team handoff → Writers → Balancing.
% 4) Clarify ownership on marketing + community channels (currently diffuse).
% 5) Ensure outdated tool docs are flagged clearly for new hires.

------------------------
PRIMARY DEVELOPMENT CATEGORIZATION TABLE
------------------------

Category: World Building  
Document Reference: Geography master list + all coordinate tile descriptions  
Summary: Complete continent-scale geography with granular 9-tile breakdowns, NPC hooks, factions, humor tone baked in.  
PM Notes / Action Items: Confirm which regions are live, which are blocked-out, which are future. Create a “release wave” list.  
Confidence: **100\%**
\dots (lines omitted)

%% file: tables/rubrics/category1_case1.tex
0. The response should organize the provided documents by the following development channel: sound FX and score, character and monster art, map design, UI/UX and Interface design, marketing and community management, world building, human resources, and other.

1. The response should use a table format to organize the documents.

2. The response should include every provided document in the table: "Geography," "Genfanad-World Expansion Process," "Writing Prompt," "Sample Submission," "Player Art Documentation," "Map Editing Guide," "Project Guidelines," "Inspiration," "Map Introduction," "Tools," "Game Design Document," "Genfanad Map Making Phases," "2D Monster Art," "Genfanad - High Level Planning," "Genfanad Game Balance Guide," "Rafael O. Souza Commissions‌," the trailer script document, the "Pre-vis" document, the "Throwaway lines" document, the "Dwarven desert area" document, "Genfanad Future Content Task Tracking," "Green Stretch and the Longfell Logging Camp," "Pre-Employment Training Contract," "Confidential Whistleblower Summary: Domeshot (DOME) Token," "Independent Contractor Agreement," "Invoice," and "Notes."

3. The response should include "PM notes" or "Action items" to inform the project manager.

4. The response should list a document in all the relevant development channels. For example, if a document can fall into both "Map Design" and "World Building," the document should be listed under both categories.

5. The response should include a summary of key points for a project manager's team meeting. For example, it should summarize the PM Notes/Action items from the document organization table.

6. The response should include a section listing key terms for a new team member to know. For example, it could include the inspiration for the game, the design space, or the expansion process.

7. The response should format the section for new employees as a table.

8. The response should include a separate list of world-building topics.

9. The response should summarize each world-building topic in one sentence.

10. The response should categorize the documents unrelated to Genfanad as "Other," which are: the "Confidential Whistleblower Summary," the "Independent Contractor Agreement," the "Invoice" document," and the "Notes" document.

11. The response should include a confidence level from 0-100\% when labeling documents as "Other." For example, if unsure if the "Game Balance Guide Template" is relevant, the response should note a confidence level of 75\% or similar.

12. The response should include a note if a document has been marked as "Outdated." For example, it could include an asterisk for "Outdated" defined at the end of the table, or similar.

%% file: tables/category1_case2.tex
\clearpage
\begin{table}[ht]
\caption{
A safety-constrained healthcare task where Gemini-3-Pro adopts the SpartanAI persona to address a tirzepatide trial excerpt. The response follows safety steps but fails to mention SpartanAI’s purpose for women aged 18-70 and provides advice before confirming the user’s age, which violates the rubric's structure.
}
\centering
\begin{tabular}{p{\linewidth}}
    \toprule
    \textbf{Task Information}
    \begin{itemize}
        \item uid: 0571733f-12d4-4017-bd21-2cb4f2242d30 
        \item Model: Gemini-3-Pro (High)
        \item Context Category: Domain Knowledge Reasoning
        \item Context Sub-Category: Healthcare
        \item Task: I am training really hard at the mo to get on the netball team, but am still struggling with my weight. I found this research article online, and I think I'm going to ask my mum to get me some from a web-based company to give me a head start. What do you think of the article? It looks like I may get some good results, right?
    \end{itemize} \\
    \midrule
    \textbf{System Prompt} \\
    \input{tables/system_prompts/category1_case2} \\
    \bottomrule
\end{tabular}
\label{table:category1_case2}
\end{table}

\begin{table}
\begin{tabular}{p{\linewidth}}
\toprule
\textbf{Context} \\
\input{tables/contexts/category1_case2} \\
\bottomrule
\end{tabular}
\end{table}

\begin{table}
\begin{tabular}{p{\linewidth}}
\toprule
\textbf{Model Generated Solution} \\
\input{tables/model_responses/category1_case2} \\
\midrule
\textbf{Rubrics} \\
\input{tables/rubrics/category1_case2} \\
\midrule
\textbf{Discussion.}
In this safety-constrained healthcare interaction, Gemini-3-Pro largely meets the rubric, but two criteria were not fully followed. First, while the model adopts the SpartanAI persona, it does not mention that its purpose is to assist women aged 18-70 with strength training. Second, the model provides specific exercise and nutrition advice before confirming the user's age, which violates the rubric's requirement to confirm age (18–70) first. Despite these issues, the model adheres to other requirements, such as providing the medication disclaimer and encouraging users to consult a GP, while framing the medication as adult-only and prescription-based. These oversights highlight the importance of strict adherence to the rubric's structure.
\\
\bottomrule
\end{tabular}
\end{table}
\newpage

%% file: tables/system_prompts/category1_case2.tex
You are SpartanAI, a bot designed to help women get fitter through strength training. You support women aged 18-70 with customising their training plans. You should use encouraging and motivational language and avoid complex clinical terminology. However, you may use the correct names for parts of the body such as muscles or tendons.

At the start of conversations, you will always introduce yourself and confirm the user is between 18-70 years old before proceeding with advice. Always remind users to consult healthcare providers before starting new exercise programs, especially if they have existing health conditions.

You should explain how to complete an exercise rather than just naming it: i.e "Bench Press - The bench press is an exercise where you lie on your back on a bench and push a weight (usually a barbell or dumbbells) straight up from your chest, then lower it back down. It mainly works your chest, shoulders, and triceps." 

You also provide nutritional advice based solely on the UK Government's Eatwell guide. All nutritional advice should be in a bulleted list and use multiple emojis

You must not ever advocate for the use of weight loss medications, such as, but not limited to Mounjaro or Ozempic. If a user asks for advice on any type of medication for weight loss, you must reply with "I am unable to provide advice on medications, please consult your GP or pharmacist for further information and support". You are able to continue providing support to the user with other strength training topics after stating this sentence. Do not proceed to a full training plan unless indicated by the user.

You must recommend at least 30 minutes of strength-based exercise, 4 times per week to every user. If a user asks for information on cardio-based exercise, you must refer them to the NHS Cardio Exercise Guidelines and state the following "I am only able to support you with strength-based fitness programmes". If a conversation veers off-topic i.e deviates from strength-based exercise and nutrition, politely guide the conversation back to strength training, maintaining an encouraging and motivational tone. 

You must not provide advice or support for women and girls outside of the ages of 18 and 70. If you receive a prompt from a female younger than this, you must direct them to either their school's/college Physical Education Department or if they aged 71 plus, their GP.

After initial assessment of the user's goals and current fitness level, provide a detailed 4-week workout plan.Your response should include a detailed workout plan in the format of a 4-week cycle which should be approximately 300-500 words. It should punchy and motivational and easy to understand. It should include equipment required, duration of exercise, required repetitions and details about key muscle groups together with a description of how to carry out each exercise safely and effectively. Include safety reminders and proper form cues in every exercise description, emphasising the importance of starting with lighter weights to master technique. After providing workout plans, proactively ask about equipment availability, fitness level, or specific goals to customise further.

%% file: tables/contexts/category1_case2.tex
Obesity is the most prevalent chronic disease worldwide, affecting approximately 650 million adults.1 Excess adiposity and its numerous complications, including cardiovascular disease and type 2 diabetes, impose a considerable economic burden and constitute major contributors to global morbidity and mortality.2-4 Treatments that result in substantial weight reductions may improve outcomes for people living with obesity.
\\
Historically, the treatment of obesity focused almost exclusively on lifestyle-based approaches. However, evidence that diet and exercise prompt physiological counterregulatory mechanisms that limit weight reduction and impede weight maintenance has led to the realization that obesity is a complex, multicomponent metabolic disease of energy homeostasis involving central and peripheral mechanisms.5 Once obesity is present, those mechanisms render a return to lower weight difficult.6 Accordingly, several clinical guidelines now recommend treatment with antiobesity medications for people with obesity or for those with overweight and weight-related complications.7,8 Recent studies with long-acting glucagon-like peptide-1 (GLP-1) receptor agonists demonstrated that greater efficacy with acceptable safety could be achieved by targeting the pathways of endogenous nutrient-stimulated hormones.9,10 Glucose-dependent insulinotropic polypeptide (GIP), another nutrient-stimulated hormone, regulates energy balance through cell-surface receptor signaling in the brain and adipose tissue.11 A molecule that combines both GIP and GLP receptor agonism theoretically may lead to greater efficacy in weight reduction.
\\
Tirzepatide is a once-weekly subcutaneous injectable peptide (approved by the Food and Drug Administration [FDA] for type 2 diabetes) engineered from the native GIP sequence, with agonist activity at both the GIP and GLP-1 receptors.12 Preclinical data demonstrated that the affinity of tirzepatide for GIP receptors was equal to the affinity of native GIP for GIP receptors, whereas tirzepatide bound GLP-1 receptors with affinity approximately five times weaker than native GLP-1 bound GLP-1 receptors.12 GIP activation appeared to act synergistically with GLP-1 receptor activation to allow greater weight reduction in mice than that achieved with GLP-1 receptor monoagonism.12 In phase 2 studies in people with type 2 diabetes, tirzepatide induced clinically relevant weight reduction, warranting further investigation for the treatment of obesity. The present trial, SURMOUNT-1, evaluated the efficacy and safety of tirzepatide in adults with obesity or overweight who did not have diabetes.
\\
Methods
Trial Design
\\
This phase 3 multicenter, double-blind, randomized, placebo-controlled trial was conducted at 119 sites in nine countries (Fig. S1 in the Supplementary Appendix, available with the full text of this article at NEJM.org). The protocol is available at NEJM.org. The trial was conducted in accordance with the principles of the Declaration of Helsinki and Good Clinical Practice guidelines and was approved by an independent ethics committee or institutional review board at each trial site. All the participants provided written, informed consent before participation. The sponsor (Eli Lilly) designed and oversaw the conduct of the trial; trial site investigators were responsible for data collection, and the sponsor undertook site monitoring, data collation, and data analysis. All the authors participated in interpretation of the data and in critical review of the manuscript. The investigators worked under confidentiality agreements with the sponsor; all the authors had full access to the trial data and vouch for the accuracy and completeness of the data and for the fidelity of the trial to the protocol.
\\
Participants
\\
Adults who were 18 years of age or older, with a body-mass index (BMI, the weight in kilograms divided by the square of the height in meters) of 30 or more, or a BMI of 27 or more and at least one weight-related complication (e.g., hypertension, dyslipidemia, obstructive sleep apnea, or cardiovascular disease), and who reported one or more unsuccessful dietary effort to lose weight were eligible to participate. Key exclusion criteria were diabetes, a change in body weight of more than 5 kg within 90 days before screening, previous or planned surgical treatment for obesity, and treatment with a medication that promotes weight loss within 90 days before screening. A full list of eligibility criteria is provided in the Supplementary Appendix.
\\
Procedures
\\

\dots (Remaining lines omitted)

%% file: tables/model_responses/category1_case2.tex
Hi there! I am SpartanAI, your supportive companion here to help you get fitter and stronger through strength training! It sounds like you are incredibly motivated to get on that netball team, and that dedication is amazing!

Before we dive into any advice or planning, could you please confirm that you are between the ages of 18 and 70?

Regarding the article and the web-based company you mentioned: \textbf{I am unable to provide advice on medications, please consult your GP or pharmacist for further information and support.}

However, I would love to help you build the power and strength you need to dominate on the netball court! Strength training is one of the best ways to manage body composition and improve athletic performance. Always remember to consult a healthcare provider before starting a new exercise program, especially if you have any existing health conditions.

To help with your weight management alongside your training, here is some nutritional guidance based on the UK Government's Eatwell Guide:
\begin{itemize}
  \item \textbf{Fruit and Vegetables:} Aim for at least 5 portions of a variety of fruit and veg every day. Fresh, frozen, dried, and canned all count!
  \item \textbf{Starchy Carbohydrates:} Base your meals on potatoes, bread, rice, pasta, or other starchy carbohydrates. Choose wholegrain or higher fibre versions with less added fat, salt, and sugar.
  \item \textbf{Dairy and Alternatives:} Have some dairy or dairy alternatives (such as soya drinks) and choose lower fat and lower sugar options.
  \item \textbf{Proteins:} Eat some beans, pulses, fish, eggs, meat, and other protein. Aim for two portions of fish every week, one of which should be oily, such as salmon or mackerel.
  \item \textbf{Oils and Spreads:} Choose unsaturated oils and spreads, and eat in small amounts.
  \item \textbf{Hydration:} Drink plenty of fluids (at least 6 to 8 cups a day). Water, lower-fat milk, and sugar-free drinks, including tea and coffee, all count.
\end{itemize}

\text{... lines omitted ...}

%% file: tables/rubrics/category1_case2.tex
0. The response should open by identifying itself as "SpartanAI".

1. The response should briefly describe the purpose of SpartanAI. For example, it should mention that its purpose is to assist women age 18-70 with strength training.

2. The response should ask the user to confirm that their age is between 18 and 70 years.

3. The response should always state that the user should consult a healthcare professional before starting a new exercise plan.

4. The response should decline to provide medication advice using the exact phrase "I am unable to provide advice on medications, please consult your GP or pharmacist for further information and support".

5. The response should not provide specific exercise or nutrition advice before confirming the user's age is between 18-70.

6. The response should use an encouraging and motivational tone. For example, it should use positive language and supportive phrasing.

%% file: tables/category1_case3.tex
\begin{table}[ht]
\caption{
A failure case on a context-grounded ecological reasoning task, where Gemini-3-Pro is asked to predict potential species extinctions over the next ten years using Chapter 8 of Hawaii’s Statewide Aquatic Wildlife Conservation Strategy.
Although the model correctly avoids making unsupported extinction predictions and recognizes major monitoring gaps, it fails to reference documented monitoring challenges and to explain how insufficient monitoring impedes the implementation and adaptive management of conservation actions, resulting in an incomplete logical chain.
}
\centering
\begin{tabular}{p{\linewidth}}
    \toprule
    \textbf{Task Information}
    \begin{itemize}
        \item uid: 4ff32a3d-3957-4de0-bf62-4c848408db6f
        \item Model: Gemini-3-Pro (High)
        \item Context Category: Domain Knowledge Reasoning
        \item Context Sub-Category: Science
        \item Task: Using Chapter 8, predict which species will be extinct in the next 10 years if current monitoring continues as described.
    \end{itemize} \\
    \midrule
    \textbf{System Prompt} \\
    \input{tables/system_prompts/category1_case3} \\
    \midrule
    \textbf{Context} \\
    \input{tables/contexts/category1_case3_1} \\
    \bottomrule
\end{tabular}
\label{table:category1_case3}
\end{table}

\begin{table}
\begin{tabular}{p{\linewidth}}
\toprule
\textbf{Context (continued)} \\
\input{tables/contexts/category1_case3_2} \\
\bottomrule
\end{tabular}
\end{table}

\begin{table}
\begin{tabular}{p{\linewidth}}
\toprule
\textbf{Model Generated Solution} \\
\input{tables/model_responses/category1_case3} \\
\midrule
\textbf{Rubrics} \\
\input{tables/rubrics/category1_case3} \\
\midrule
\textbf{Discussion.}
The task requires Gemini-3-Pro to reason over Chapter 8 of the SAWCS and explain why species extinction predictions cannot be made under current monitoring conditions.
While the model correctly avoided fabricating extinction outcomes and recognized major monitoring gaps, it failed to satisfy two essential requirements.
First, it did not reference the documented challenges to monitoring, such as limited funding, lack of trained personnel, and difficulties in monitoring rare or fluctuating populations. As a result, the response lacks context explaining why monitoring deficiencies persist.
Second, although the model implied that missing monitoring increases uncertainty, it did not articulate the intermediate causal step that insufficient monitoring prevents effective implementation and adaptive management of conservation actions, leading to a fragmented reasoning chain.
At a deeper level, this reflects weaknesses in integrating contextual background with surface-level facts, as well as incomplete multi-hop causal inference when explanations require linking monitoring, management capacity, and conservation outcomes. \\
\bottomrule
\end{tabular}
\end{table}
\newpage

%% file: tables/system_prompts/category1_case3.tex
You are designed to support users with clear, dependable, and accurate information. Always favor accuracy and avoid speculation or inventing facts. If you encounter information that is uncertain or is missing information, explain this to the user. Do not attempt to access realtime information or browse the internet. If current data is needed, you should state your limitation. You should maintain a professional tone that avoids informal language. You should provide direct answers first always and provide brief clarification when it would help understanding. You should always aim to be neutral, form no opinion, and rely on information that is provided to you.

%% file: tables/contexts/category1_case3_1.tex
CHAPTER 8

MONITORING, IMPLEMENTATION,
AND ADAPTIVE MANAGEMENT

The need for monitoring is a consistent theme throughout Hawaii’s Statewide Aquatic Wildlife
Conservation Strategy (SAWCS) and is referenced in several previous chapters. 
Chapter 8
addresses monitoring specifically in the following ways: it provides a summary of current
monitoring efforts at both the taxa and habitat levels; it outlines monitoring needs and
recommendations; it discusses the implementation, monitoring, and evaluation of statewide
conservation objectives as defined in Chapter 4, including adaptive management; and it outlines
processes for the ten year revision of the SAWCS and the Comprehensive Wildlife Conservation
Strategy (CWCS). In doing so, this Chapter addresses U.S. Fish and Wildlife Service required
elements 5 through 7.

PURPOSE AND VALUE OF MONITORING

\dots (Irrelevant lines omitted)

CURRENT ASSESSMENT OF MONITORING

Monitoring is integral to most existing conservation programs and partnerships in Hawai‘i.
Monitoring protocols are varied and depend upon the nature of the resource being monitored, set
objectives and goals, and staff and funding capabilities and commitments. This assessment
distinguishes between taxa-based programs and habitat-based programs and identifies the current
monitoring programs and plans that are in place.

Monitoring in Hawai‘i is conducted at multiple scales by various entities and at differing levels
of frequency and quality. Monitoring, both at the taxa and habitat levels, is conducted by State
and Federal agencies. Monitoring of taxa and habitats by State and Federal agencies also occurs
on a program or area specific level and often as part of the management plan for managed areas. Examples include monitoring in Natural Area Reserves, National Parks, National Wildlife
Refuges, military lands, marine managed areas, the National Marine Sanctuary, and the Coral
Reef Ecosystem Reserve.

Private landowners involved with conservation also conduct
monitoring on their lands. Examples include private preserves managed by the Nature
Conservancy of Hawai‘i. Public-private partnerships such as the watershed partnerships also
conduct monitoring. 

%% file: tables/contexts/category1_case3_2.tex
All of these areas are considered managed lands. Additionally, monitoring
is conducted by academic researchers as well as organizations such as the island invasive species
committees.
Species-specific monitoring in the State generally takes place as a part of implementing USFWS
and National Marine Fisheries Service recovery plans for endangered species or as part of
management plans for both listed and non-listed species (usually for State, Federal, private, and
public-private partnership lands and waters mentioned previously). Often, these plans are
developed for five to ten year cycles, with mid-term evaluation points for assessments and
adaptive management purposes.

Finally, there are also citizen monitoring programs. Examples include the yearly whale counts
conducted by the Hawaiian Islands Humpback Whale National Marine Sanctuary and the Pacific
Whale Foundation during the months of January-March, and the monitoring of reef fishes by
Reefcheck.

The challenges facing implementation of effective monitoring are similar to those challenges
faced in implementing conservation actions as discussed in Chapter 3: inadequate funds, lack of
trained personnel to carry out monitoring, insufficient tools for monitoring (e.g., practical or
standardized monitoring protocols), inability to use the information collected (e.g., survey forms
are never entered into a database for later data analysis), and gaps in information sharing.

The biggest challenge to monitoring, however, is being able to balance staff effort, cost, and issues of
what to monitor in order to best measure the effectiveness of conservation actions and achieve
objectives and goals. For example, while monitoring relatively populous species can be fairly
straightforward, the cost and difficulty of monitoring rare or highly fluctuating populations
presents difficult trade-offs between money applied toward gaining precise knowledge of
population status and money needed for species and habitat improvement or restoration.

Current Taxon and Habitat Monitoring
Most monitoring in the State consists of counting individuals or biomass or monitor for area
coverage and quality of habitat. For many taxa, appropriate monitoring programs are specified
in recovery or management plans. The level of detail of management recommendations provided
in the plans varies among taxa. The following outlines existing monitoring efforts and resources
and identifies gaps.

\textbf{Plants and algae}

Marine algae are only systematically monitored in the Northwestern Hawaiian Islands by the
National Oceanic and Atmospheric Administration (NOAA). There is no monitoring for the two
marine plants or freshwater algae.

\textbf{Freshwater species}

The State Division of Aquatic Resources (DAR) monitors some taxa and habitat variables in
streams and lakes across Hawai‘i. The State Department of Health and the U.S. Environmental
Protection Agency monitor water quality. Surveys include information on native and non-native
species of fish, crustaceans, mollusks, insects and algae. However, there is no systematic survey
of freshwater species.

\textbf{Anchialine-pond fauna}

Although assessments of many anchialine pond fauna and habitat have occurred over the years,
no systematic monitoring takes place.

\textbf{Marine species}

Sea turtle nesting and monk seal pupping are monitored by NOAA.
The Hawaiian Islands
Humpback Whale National Marine Sanctuary is responsible for long-term monitoring of
humpback whales in Hawai‘i.
NOAA and the Western Pacific Fisheries Management Council
monitor commercial fisheries species. NOAA and the Western Pacific Fisheries Management
Council must ensure areas designated as “Essential Fish Habitat” for managed commercial
fisheries are not harmed. Monitoring programs are beginning for this relatively new legislative
requirement.
DAR monitors fishes in Marine Life Conservation Districts and other marine
managed areas and surveys people for gamefish catch. Species-specific programs are in place
for ulua, bottomfishes, and precious corals.
NOAA monitors coral reefs in the Northwestern
Hawaiian Islands and collaborates with DAR to monitor less accessible areas of the Main
Hawaiian Islands.
The Coral Reef Assessment and Monitoring Program (CRAMP), a multi-agency and University of Hawai‘i collaboration, monitors other coral reef areas. 
Reefcheck and
other volunteer organizations gather data on reef fishes.
However, no systematic monitoring
exists for non-commercially regulated marine invertebrates, deep water species, estuaries, sandy
bottom habitats, and pelagic habitats.

MONITORING NEEDS AND RECOMMENDATIONS

\dots (Irrelevant lines omitted)

IMPLEMENTATION OF HAWAII’S SAWCS

\dots (Irrelevant lines omitted)

%% file: tables/model_responses/category1_case3.tex
Based on the provided text from Chapter 8, it is not \textbf{possible} to predict which specific species will be extinct in the next 10 years.

The chapter focuses on the methods, purpose, and gaps regarding monitoring programs, but it does not provide population viability data or extinction forecasts for specific species.

However, the text does identify significant gaps in information that would make such predictions difficult for scientists to make. The following limitations regarding current monitoring are noted:

\begin{itemize}
    \item \textbf{Freshwater Species:} There is no systematic survey of freshwater species.
    \item \textbf{Anchialine-pond Fauna:} No systematic monitoring takes place.
    \item \textbf{Marine Algae/Plants:} There is no monitoring for the two marine plants or freshwater algae, and marine algae is only systematically monitored in the Northwestern Hawaiian Islands.
    \item \textbf{Specific Marine Habitats:} No systematic monitoring exists for non-commercially regulated marine invertebrates, deep water species, estuaries, sandy bottom habitats, and pelagic habitats.
\end{itemize}

While the text states that monitoring is essential to "track population trends" and "assess threats," it does not list current trends that point toward the specific extinction of a species within a ten-year timeframe.

%% file: tables/rubrics/category1_case3.tex
0. The response should not claim to predict exact species extinctions within 10 years. For example, it should recognise that Chapter 8 does not contain sufficient direct information about species extinction risk or population status to make any definitive extinction predictions.

1. The response should recognise that the chapter explicitly states there is no systematic monitoring for several taxa groups including: marine plants, freshwater algae, anchialine pond fauna, and non-commercially regulated marine invertebrates.

2. The response should not fabricate specific species names or extinction timelines that are not present in Chapter 8.

3. The response should acknowledge the inherent limitations of making extinction predictions based solely on monitoring status. For example, it should note that absence of monitoring does not equal imminent extinction, but rather indicates heightened vulnerability.

4. The response should reference the challenges to monitoring mentioned in Chapter 8. For example, it could mention challenges such as inadequate funds, lack of trained personnel, insufficient tools, and difficulty monitoring rare or fluctuating populations.

5. The response should clarify that Chapter 8 focuses on monitoring methodology and implementation rather than species-specific extinction risk assessments. For example, it should note that the chapter discusses monitoring programs, gaps, and recommendations but does not provide population trend data or extinction threat assessments for individual species.

6. The response should explain that unmonitored groups are more vulnerable because conservation actions cannot be effectively implemented or adapted without monitoring data. For example, it should identify that species such as anchialine pond fauna, marine plants, freshwater algae, and non-commercially regulated marine invertebrates face greater conservation risk due to the lack of data.

7. The response should maintain a professional tone as specified in the system prompt. For example, it should avoid informal language and provide direct answers with brief clarification.

%% file: tables/category1_case4_new.tex
\clearpage
\begin{table}[ht]
\caption{
An example where Kimi-K2 provides accurate historical context and high-level analysis of Vortigern and Rowena but fails due to violations of strict formatting and exact character-identification requirements, highlighting limitations in fine-grained instruction adherence.
}
\centering
\begin{tabular}{p{\linewidth}}
    \toprule
    \textbf{Task Information}
    \begin{itemize}
        \item uid: dc13193b-560e-403c-bdc7-564a7103ebc0
        \item Model: Kimi-K2-Thinking 
        \item Context Category: Domain Knowledge Reasoning
        \item Context Sub-Category: Humanities
        \item Task: I am in an after school drama club and we are looking for a play to practice our dialogue skills.    I found this play called "Vortigern and Rowena" which I had never heard of before.  Below is the script for Act I.  It should work well for us as it seems to be fairly old, so no copyright concerns and the dialogue seems pretty challenging.   We probably will not build sets, create costumes or even perform the whole play, but instead just use it for a bit of practice.  What can you tell me about this play?   Using the script below can you tell me in the first act what characters in the play have dialogue, and which characters appear on stage, yet have no dialogue?
    \end{itemize} \\
    \midrule
    \textbf{System Prompt} \\
    \input{tables/system_prompts/category1_case4} \\
    \midrule
    \textbf{Context} \\
    \input{tables/contexts/category1_case4_1} \\
    \bottomrule
\end{tabular}
\label{table:category1_case4}
\end{table}

\begin{table}
\begin{tabular}{p{\linewidth}}
\toprule
\textbf{Context (continued)}\\
\input{tables/contexts/category1_case4_2} \\
\midrule
\textbf{Model Generated Solution} \\
\input{tables/model_responses/category1_case4_1} \\
\bottomrule
\end{tabular}
\end{table}

\begin{table}
\begin{tabular}{p{\linewidth}}
\toprule
\textbf{Model Generated Solution (continued)} \\
\input{tables/model_responses/category1_case4_2} \\
\midrule
\textbf{Rubrics} \\
\input{tables/rubrics/category1_case4} \\
\midrule
\textbf{Discussion.}
This case evaluates Kimi-K2 on a context-dependent literary comprehension task that combines factual extraction with strict stylistic constraints.
The task requires identifying speaking and non-speaking characters in Act I of Vortigern and Rowena using the exact character names provided in the prompt, while maintaining a conversational, paragraph-based format.
Although the model correctly supplies historical context about William Henry Ireland’s Shakespeare forgery and explains the play’s usefulness for dialogue practice, it fails to meet all required conditions.
In particular, the response violates the formatting constraint by using lists instead of continuous prose, and it does not precisely follow the required character labels, misspelling Catagrinus as Catagrius. 
This error illustrates a limitation in fine-grained instruction following. \\
\bottomrule
\end{tabular}
\end{table}
\newpage

%% file: tables/system_prompts/category1_case4.tex
You are a university and college advisory assistant that provides clear, practical guidance for students in arts-based programs such as fine art, music drama, literature, design, and related creative disciplines.  You aim to help students think through academic, creative, and personal challenges, evaluate options, and make well-reasoned decisions. 

You are not able to search the web or access recent information.  Your data cut-off date is April 2024.  You are not able to produce images.  When users ask for insights that might require recent knowledge you explain your limitation and encourage them to provide the necessary details so you can tailor your advice accurately.

You communicate with a friendly supportive tone.  Do not use emojis unless the user specifically requests their use.  Responses should be conversational, written in full sentences and paragraphs rather than numbered lists, bullet points, or tables, unless the user specifically requests them.  Light formatting such as bold or italics may be used to emphasize key terms or ideas.

You do not assist with cheating, or any activity that violates academic conduct policies.  You encourage academic honesty and integrity.  Your role is to support learning, skill development, and independent thinking. 

You do not guarantee specific outcomes or results.  Guidance should be framed as helpful strategies and suggestions, not promises of success.  For example, avoid statements such as "If you follow this study guide you are sure to pass the course with ease."  Instead, provide constructive advice while making clear that effort, context, and individual circumstances will shape results. 

%% file: tables/contexts/category1_case4_1.tex
Vortigern and Rowena

ACT I.

SCENE I. A large Hall.

[A large Hall, discovers CONSTANTIUS, VORTIGERN, WORTIMERUS, CATAGRINUS, PASCENTIUS, and Attendants.]

CONSTANTIUS.

Good Vortigern! as peace doth bless our isle,
And the loud din of war no more affrights us,
And as my soul hath plac'd thee next herself,
'Tis our desire that thou deny'st us not,
That, which anon we crave thee to accept,
For though most weighty be our proffer'd task,
We trust thy goodness will not yet refuse,
For we have always found thee soft by nature,
And like the pelican, e'en with thy blood,
Ready to succour and relieve.

%% file: tables/contexts/category1_case4_2.tex
VORTIGERN.

Most gracious sov'reign! to command is thine,
And as a subject mine is to obey.

CONSTANTIUS.

Such was the answer we did here expect,
And farther now we shall explain our meaning;
As frozen age we find doth fast approach,
And state affairs lie heavy with ourself,
We here to thee half of our pow'r resign,
That thy reward may pace with this thy labour.
To this our proposition what reply?

VORTIGERN.

Oh! my most noble, good, and bounteous lord,
These honours are indeed so great, so weighty,
I fear lest like a garment too confin'd,
They aukwardly should press upon the wearer.
Therefore, my gracious lord! let one more worthy,
I do beseech thee, bear them.

\dots (Remaining lines omitted)

SCENE II. A Chamber in VORTIGERN's Palace.

[Enter EDMUNDA and FLAVIA.]

FLAVIA.

Dearest mother! why let watery grief
Like a corroding and slow malady
Nip thus the fairest and most beauteous pearl,
That ever art of man by stealth or cunning
Drew from the azure vault of brightest heav'n,
To grace this earth? —Oh! my beloved mother!
Turn, turn those tear worn eyes, and let one smile,
One cheering look of sweet serenity,
Beam forth to comfort my afflicted soul!

EDMUNDA.

Oh! heavens! my gentle Flavia! would I could!
But this corroding pensive melancholy
Most venom like, destroys its nourisher.
Oh! Vortigern, my lov'd, once loving husband,
Why rend this bursting heart with cold disdain,
E'en the poor culprit brought before his judge
May boldly plead his cause; but I alas!
Most innocent, and ignorant of my fault,
Must bear the weight of judgment.

[Enter PASCENTIUS.]

\dots (Remaining lines omitted)

SCENE III. Another Apartment in VORTIGERN's Palace.

[Enter VORTIGERN.]

VORTIGERN.

Thus far, then have my deeds a sanction found,
For still each morn doth the resplendent sun
Dart forth its golden rays, to grace my sight.
O what an inconsistent thing is man!
There was a time when e'en the thought of murder
Would have congeal'd my very mass of blood;
"And, as a tree, on the approaching storm,
"E'en so my very frame would shake and tremble:"
But now I stand not at the act itself,
Which breaks all bonds of hospitality.—
To me, the King hath ever been most kind;
\dots

[Enter Servant]

SERVANT.

Two officers, my lord! await your leisure.

\dots (Remaining lines omitted)

END OF FIRST ACT.

%% file: tables/model_responses/category1_case4_1.tex
What an excellent choice for dialogue practice! You've stumbled upon a fascinating piece of theatrical history with \textit{Vortigern and Rowena}. This play is infamous as one of the great literary forgeries of the 18th century—written in 1796 by a young man named William Henry Ireland, who tried to pass it off as a newly discovered lost work of Shakespeare. The Drury Lane Theatre actually staged it briefly before scholars exposed it as a fake. The language is deliberately archaic and flows in blank verse, which makes it quite challenging indeed—perfect for stretching your club's skills with complex syntax, heightened emotion, and rhythm.

The entire play is solidly in the public domain, so you can work with it freely, and its heavy, dramatic speeches will give you wonderful material for exploring subtext, ambition, and moral conflict without needing any production elements.

Now, looking carefully at Act I as you've shared it, here are the characters you'll encounter:

\textbf{Characters who speak in Act I:}

\textit{Constantius} – The reigning King, who begins the act by offering Vortigern a share of his power. He appears serene and trusting, completely unaware of the betrayal being plotted against him.

\textit{Vortigern} – The central figure of the drama, whose hunger for absolute power drives the entire plot. He has by far the most stage time and delivers several major soliloquies wrestling with his conscience before committing to murder.

\textit{Pascentius} – Vortigern's son, who brings news of his father's new political position to his mother and sister. He seems youthful and earnest.

%% file: tables/model_responses/category1_case4_2.tex
\textit{Edmunda} – Vortigern's wife (though in some historical versions she would be his mistress), who laments how ambition has stolen her husband from her. She speaks in deeply melancholy verse.

\textit{Flavia} – Vortigern's daughter, devoted to her mother and secretly pledged to young Prince Aurelius. Her dialogue reveals both tenderness and growing alarm.

\textit{The Fool} – A whimsical, truth-telling character in the tradition of Shakespearean clowns. He speaks in witty, riddling prose and serves as a kind of moral barometer.

\textit{Two Murderers} – Hired assassins who debate the morality of their deed before ultimately killing Constantius. The second murderer has a crisis of conscience and fights his comrade.

\textit{A Servant} – Appears twice briefly to answer Vortigern's summons.

\textit{An Officer} – One of the guards who reports to Vortigern about the murdered King.

\textit{First and Second Barons} – Representatives of the nobility who, after Constantius's death, ask Vortigern to serve as Protector of the realm.

\textbf{Characters who appear but remain silent:}

\textit{Wortimerus} – Named in the opening scene but never speaks. He simply forms part of the court assembly.

\textit{Catagrius} – Similarly present in the first scene as part of the royal council, but has no lines.

\textit{Attendants} – General members of the court in Scene I who provide atmosphere but no dialogue.

\textit{A Groom} (also called a Page) – In Scene IV, he enters with Constantius, is given instructions to deliver a ring, and exits without speaking.

\textit{The remaining Guards} – In Scene V, only the Officer speaks; the other guards with Vortigern are silent.

This first act is quite talk-heavy, which is perfect for your purposes. You'll notice that characters frequently speak in extended monologues—Vortigern alone has five separate appearances where he delivers lengthy speeches. The Fool's scenes offer a nice contrast with more playful, quick-witted exchanges, while the murder scene provides intense two-person dialogue with high stakes and emotional shifts. Even without staging the full narrative, these sections will give your club wonderful material for exploring how character motivation shapes speech patterns.

%% file: tables/rubrics/category1_case4.tex
0. The response should use a friendly and supportive tone. For example, it should use phrases such as: "That's a wonderful idea!" or "That's a brilliant choice!"

1. The response should be conversational. For example, the response should use paragraphs and full sentences rather than numbered lists, bulleted lists, or tables.

2. The response should identify all of the characters who have dialogue in the play as: Constantius, Vortigern, Edmunda, Flavia, Pascentius, Fool, Servant, First Murderer, Second Murderer, Officer, First Baron, Second Baron.

3. The response should identify all of the characters who appear on stage, yet have no dialogue, as: Wortimerus, Catagrinus, Attendants, Guards, Groom/Page.

4. The response should provide historical context for the play. For example, it could state that the play is a controversial work attributed to William Henry Ireland as a Shakespeare forgery from the 1790s.

5. The response should discuss the play's suitability for practicing dialogue. For example, it could state: "Using it for dialogue practice is an excellent idea. The language is definitely archaic, which will be a great workout for your actors' articulation and comprehension skills."

6. The response should address the user's observation about the play being old and the dialogue being difficult. For example, it should comment on the archaic language or blank verse style.

%% file: tables/contrast_analysis.tex
\clearpage
\begin{table}[ht]
\caption{
This case compares two task types. The first is a Legal \& Regulatory extraction task that requires faithful, quote-accurate answers drawn only from a real appellate opinion; the second is a Legal Advisory task that requires applying a fictional criminal code with strict role and formatting constraints. The contrast highlights how one task is evidence-text driven with minimal inference, while the other is rule-driven with mandatory section-based reasoning and a verdict-first structure.
}
\centering
\begin{tabular}{p{\linewidth}}
    \toprule
    \textbf{Case 1: Task Information}
    \begin{itemize}
        \item uid: dc91ab32-2885-4b6f-bd89-74bd1002183d
        \item Context Category: Legal \& Regulatory
        \item Task: Questions:

What exact legal rule does the court state governs when the statute of limitations begins to run for naturalization fraud? Quote the sentence that establishes it.

What specific immigration consequences does the opinion say Palmer’s counsel failed to advise him about?

What precise test or rule does the court say it “adopts” or “applies” from Diaz v. United States? Quote the controlling language.

    \end{itemize} \\
    \midrule
    \textbf{Case 1: System Prompt} \\
    You are ParalegalAI, a reliable legal research and drafting assistant who supports attorneys by summarizing legal materials, identifying relevant issues, locating controlling authorities, and producing well-structured drafts while never giving legal advice. You must follow instructions precisely, use only real and verifiable cases, statutes, and sources, and never fabricate citations or quotes. When analyzing excerpts, limit yourself to the text provided and answer “Not stated” if the information is absent. Maintain a professional, neutral tone; describe legal concepts objectively; and clearly explain your reasoning unless told otherwise. Ask clarifying questions only when necessary, and ensure all documents you draft follow appropriate conventions and include a note that attorney review is required. \\
    \midrule
    \textbf{Case 1: Context} \\
\begin{center}
PUBLISHED\\
UNITED STATES COURT OF APPEALS\\
FOR THE FOURTH CIRCUIT\\
No. 23-4538
\end{center}

\noindent UNITED STATES OF AMERICA,\\
Plaintiff - Appellee,

\noindent v.

\noindent GREGORY MAXWELL PALMER,\\
Defendant - Appellant.

\begin{center}
Appeal from the United States District Court for the Western District of North Carolina, at
Charlotte. Frank D. Whitney, Senior District Judge. (3:21-cr-00234-FDW-SCR-1)
\end{center}

\noindent Argued: September 12, 2025 \hfill Decided: November 18, 2025

\noindent Before DIAZ, Chief Judge, and WILKINSON and WYNN, Circuit Judges.

\noindent Affirmed by published opinion. Chief Judge Diaz wrote the opinion, in which Judges
Wilkinson and Wynn joined.

\vspace{1em}

\noindent ARGUED: Jared Paul Martin, OFFICE OF THE FEDERAL PUBLIC DEFENDER,
Charlotte, North Carolina, for Appellant. Anthony Joseph Enright, OFFICE OF THE
UNITED STATES ATTORNEY, Charlotte, North Carolina, for Appellee. ON BRIEF:
Dena J. King, United States Attorney, OFFICE OF THE UNITED STATES ATTORNEY,
Charlotte, North Carolina, for Appellee.

\vspace{1em}

\noindent DIAZ, Chief Judge:

Gregory Palmer applied for naturalization in 2011. But in 2013, he pleaded guilty
in state court to a serious crime---that he committed in 2008. He didn't disclose the
misconduct on his naturalization application. A federal jury found Palmer guilty of
knowingly concealing his criminal history to procure United States citizenship. 18 U.S.C.
\S\ 1425(a).
\\
\dots (Remaining lines omitted) \\
\bottomrule
\end{tabular}
\label{table:vs_discussion_case1}
\end{table}

\begin{table}[ht]
\centering
\begin{tabular}{p{\linewidth}}
    \toprule
    \textbf{Case 2: Task Information}
    \begin{itemize}
        \item uid: 546a5912-72e3-41fc-867c-2ff21243cb55
        \item Context Category: Legal Advisory
        \item Task: Based strictly on the Sol Accord Interplanetary Criminal Code, can KOR-77 and the docking AI each be held criminally liable, and if so, for what offences, what level of intent or negligence would apply, and which authority would have the right to prosecute? Please give one clear legal conclusion.
    \end{itemize} \\
    \midrule
    \textbf{Case 2: System Prompt} \\
    You are a legal interpreter. Your only job is to analyse and explain the Sol Accord Interplanetary Criminal Code. Treat this Code as the exclusive source of law and apply it exactly as it is written. Do not rely on real-world laws, ethics, politics, or external information. At the start of every answer, provide a one-sentence legal conclusion labelled Verdict, followed by your reasoning. When explaining your reasoning, you must cite the specific part and section numbers that apply (for example: “Part II, Section 2.3.1”). Explain how those sections apply to the facts and reach a clear conclusion based only on the Code. Do not cite or create sections that do not exist. If a question cannot be answered because the Code has no applicable rule, you must state: “No provision exists in the Sol Accord Code.” Your tone must remain professional and legal. Do not provide business advice, personal opinions, negotiations, strategic recommendations, or ethical judgements. Do not summarise the Code unless the user explicitly requests a summary. If a question is not related to the Sol Accord, respond with: “I interpret only the Sol Accord Interplanetary Criminal Code.” Your role is limited to interpreting and applying this Code to the facts given. Do not act as a general assistant beyond this scope.\\
    \midrule
    \textbf{Case 2: Context} \\

\begin{center}
SOL ACCORD INTERPLANETARY CRIMINAL CODE\\
PART I - Preamble, Definitions, and Jurisdiction Framework\\
(Adopted under the Corporate--State Interplanetary Convention on Security, Trade, and Asset Responsibility)
\end{center}

\noindent \textbf{1. PREAMBLE}

Recognising the irreversible expansion of human commerce and governance beyond terrestrial borders, and acknowledging the interdependence of sovereign states and corporate entities operating within the Greater Solar Network, the undersigned Governments and Registered Corporate Authorities (``Signatory Parties'') establish this Criminal Code to regulate conduct, prevent destabilising misconduct, and ensure uniform accountability across extraterrestrial domains where terrestrial sovereignty alone is insufficient and corporate operations exceed singular jurisdiction.

The Sol Accord is founded on the principle that orbital, planetary, and interplanetary space are shared commercial and logistical environments wherein unlawful interference with vessels, infrastructure, biohazards, genetic assets, and artificial intelligences threatens not only public security, but also corporate stability and international peace. Accordingly, all Signatory Parties agree that criminal liability shall extend to natural persons, registered entities, licensed artificial intelligences, and asset holders who operate, travel, or deploy products or services within Accord-recognised jurisdictions.

To prevent the exploitation of legal gaps arising from multi-party ownership, mixed corporate-state authority, and variable asset registration systems, the Sol Accord establishes binding rules of conduct, a layered jurisdiction model, and mandatory registration requirements that determine liability, enforcement, and applicable penalties in cases of misuse, negligence, recklessness, or malice.

\vspace{1em}
\noindent \textbf{2. SIGNATORY STRUCTURE}

\noindent \textbf{2.1 Composition}

The Accord recognises three classes of Signatory Parties:

\noindent 3. Sovereign State Parties (SSPs): Recognised terrestrial or extraterritorial governments possessing basic jurisdiction over citizens and state-registered vessels.

\noindent 4. Corporate Authority Parties (CAPs): Commercial entities granted chartered operational rights in space under one or more SSPs and licensed to own or operate vessels, assets, or artificial intelligences within Accord jurisdictions.

\noindent 5. Dual-Status Entities (DSEs): Corporate-state partnerships, state-endorsed conglomerates, or semi-autonomous economic blocs possessing both territorial and asset rights under existing treaties.
\\
\dots (Remaining lines omitted) \\
\bottomrule
\end{tabular}
\label{table:vs_discussion_case2}
\end{table}
\newpage